\documentclass{article} % For LaTeX2e
\usepackage{iclr2027_conference,times}

\usepackage{amsmath,amsfonts,bm}

\def\eqref#1{equation~\ref{#1}}
\def\1{\bm{1}}

\DeclareMathAlphabet{\mathsfit}{\encodingdefault}{\sfdefault}{m}{sl}
\SetMathAlphabet{\mathsfit}{bold}{\encodingdefault}{\sfdefault}{bx}{n}

\DeclareMathOperator*{\argmax}{arg\,max}

\usepackage{hyperref}
\usepackage{url}
\usepackage{multirow}
\usepackage{booktabs}
\usepackage{wrapfig}
\usepackage{cleveref}
\usepackage[table]{xcolor}
\usepackage{graphicx}
\usepackage{subfig}
\usepackage{pifont}
\usepackage{algorithm,algpseudocode}

\crefname{equation}{Eq.}{Eqs.}
\crefname{section}{Sec.}{Secs.}
\crefname{table}{Tab.}{Tabs.}
\crefname{figure}{Fig.}{Figs.}
\crefname{algorithm}{Alg.}{Algs.}
\crefname{appendix}{App.}{Apps.}

\title{Compositional Diffusion Planning with\\Chunk Selection and Reachability Measure}

\author{Chanyoung Park\quad Youngjoong Kim\quad Jaesik Park\\
Seoul National University\\
\texttt{\{vitacpark, noah.kim, jaesik.park\}@snu.ac.kr} \\
}

\iclrfinalcopy % Uncomment for camera-ready version, but NOT for submission.
\begin{document}

\maketitle
\lhead{Preprint}

\vspace{-1em}
\begin{abstract}
Compositional diffusion planners have demonstrated remarkable success on long-horizon diffusion planning by sampling short trajectories, namely chunks, and composing them into a long trajectory. However, averaging overlapping chunks can induce the mode-averaging problem, resulting in discontinuous trajectories. Several works aim to address this problem, whereas a non-negligible portion of plans remains infeasible. In this work, we propose \textit{Selective Compositional Diffuser (SCD)}, handling this problem by measuring reachability. To this end, we present a self-supervised learning method for the reachability classifier. This is trained on offline data and generated chunks jointly to address potential distributional mismatch, while reducing the bias inherited from the generated data. Based on this, we construct a graph whose edge costs represent whether consecutive chunks are reachable, and traverse the graph to find effective yet efficient routes. Lastly, we generate bridging trajectories and refine the composed trajectory for smooth transitions. We evaluate our method on OGBench stitching tasks. It practically reduces incompatible mode compositions and the mode-averaging problem compared to the studied baselines. Across tasks, SCD improves success rates, reaching state-of-the-art results on HumanoidMaze-Large and Giant tasks.
\end{abstract}

\section{Introduction}
Diffusion models~\citep{NEURIPS2019_3001ef25, NEURIPS2020_4c5bcfec} have demonstrated remarkable results in robotic planning and decision-making~\citep{ajay2023is, Chi-RSS-23, pmlr-v162-janner22a}. These models abstract planning as a network-based generative process and are known to handle multimodality, flexible conditioning, and guidance-style control signals~\citep{NEURIPS2021_49ad23d1, ho2021classifierfree}. However, since diffusion models are inherently constrained by the training data, diffusion-based planners may require a combinatorial-scale dataset to model all possible plans.

To handle this problem, recent works~\citep{luo2025generative, lee2026refiningcompositionaldiffusionreliable, sun2026energybased, ICLR2026_6fbe71b5} try to stitch or compose short trajectories into a full trajectory for long-horizon tasks. To this end, they learn short-horizon trajectories, namely chunks, conditioned on boundary states and compose them at inference time. Since short trajectories are more practical to collect in the real world than full ones, compositional diffusion planners represent a promising and practical direction.

\begin{wrapfigure}{R}{0.5\linewidth}
\centering
\begin{minipage}{0.05\linewidth}\centering\rotatebox{90}{\small{CompDiffuser}}\end{minipage}
\begin{minipage}{0.30\linewidth}\centering\includegraphics[width=\linewidth]{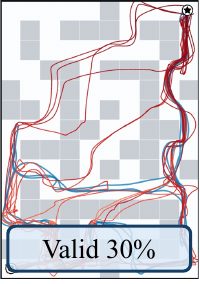}\end{minipage}
\begin{minipage}{0.30\linewidth}\centering\includegraphics[width=\linewidth]{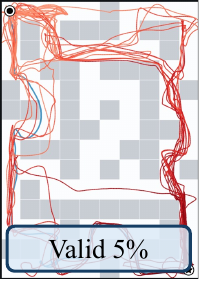}\end{minipage}
\begin{minipage}{0.30\linewidth}\centering\includegraphics[width=\linewidth]{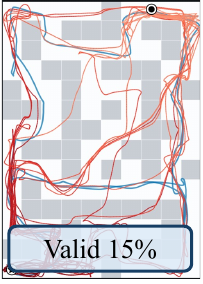}\end{minipage}

\begin{minipage}{0.05\linewidth}\centering\rotatebox{90}{\small{ECD}}\end{minipage}
\begin{minipage}{0.30\linewidth}\centering\includegraphics[width=\linewidth]{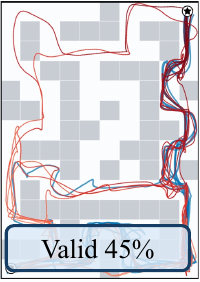}\end{minipage}
\begin{minipage}{0.30\linewidth}\centering\includegraphics[width=\linewidth]{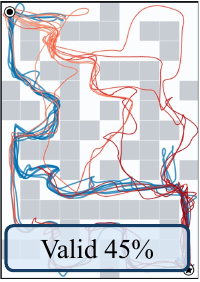}\end{minipage}
\begin{minipage}{0.30\linewidth}\centering\includegraphics[width=\linewidth]{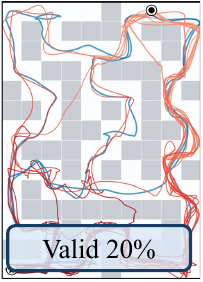}\end{minipage}

\begin{minipage}{0.05\linewidth}\centering\rotatebox{90}{\small\textbf{SCD (Ours)}}\end{minipage}
\begin{minipage}{0.30\linewidth}\centering\includegraphics[width=\linewidth]{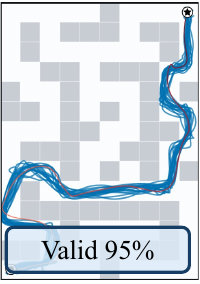}\end{minipage}
\begin{minipage}{0.30\linewidth}\centering\includegraphics[width=\linewidth]{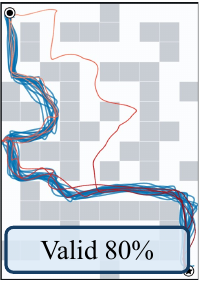}\end{minipage}
\begin{minipage}{0.30\linewidth}\centering\includegraphics[width=\linewidth]{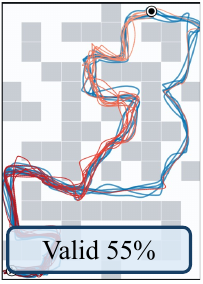}\end{minipage}

\begin{minipage}{0.05\linewidth}\ \end{minipage}
\begin{minipage}{0.30\linewidth}\centering\small{(a) Task 01}\end{minipage}
\begin{minipage}{0.30\linewidth}\centering\small{(b) Task 02}\end{minipage}
\begin{minipage}{0.30\linewidth}\centering\small{(c) Task 03}\end{minipage}

\caption{\textbf{Failure cases in HumanoidMaze-Giant.} The red color denotes infeasible trajectories and the blue color denotes feasible ones. \textit{Valid} refers to the percentage of feasible trajectories. More samples are provided in \cref{app:qualitative-results}, \cref{fig:more-failure-cases}.}
\label{fig:failure-cases}
\vspace{-4em}
\end{wrapfigure}

Previously, several works~\citep{pmlr-v229-mishra23a, luo2025generative} generate multiple chunks and compose overlapping regions by averaging. In particular, when the chunk distribution is multimodal, it is known to suffer from the mode-averaging problem, and it can result in discontinuous trajectories. Recently, \citet{ICLR2026_6fbe71b5,sun2026energybased,lee2026refiningcompositionaldiffusionreliable} have addressed this problem with iterative resampling~\citep{ICLR2026_6fbe71b5}, conservative score correction~\citep{sun2026energybased}, and overlap consistency guidance~\citep{lee2026refiningcompositionaldiffusionreliable}. However, as shown in \cref{fig:failure-cases}, \textit{we observe that a non-negligible portion of plans remains infeasible, failing to arrange feasibly composable chunks.}

\begin{figure}[t]
\centering
\includegraphics[width=0.95\linewidth]{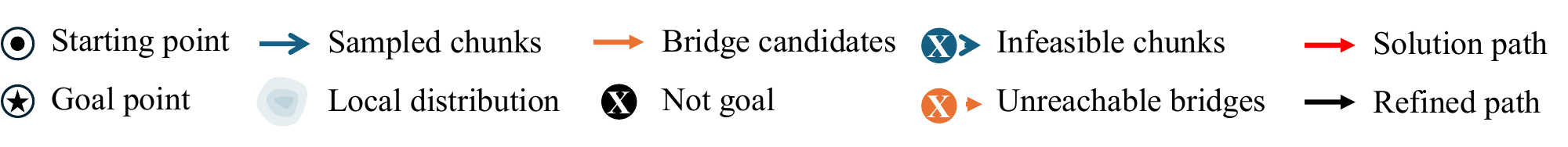} \\
\vspace*{-3mm}
\subfloat[Generated chunks]{
\includegraphics[width=0.23\linewidth]{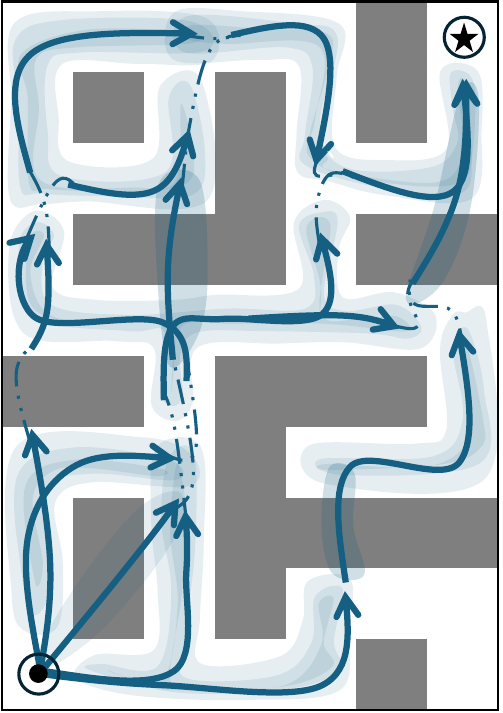}
\label{fig:overview-a}
}
\subfloat[Measuring reachability]{
\includegraphics[width=0.23\linewidth]{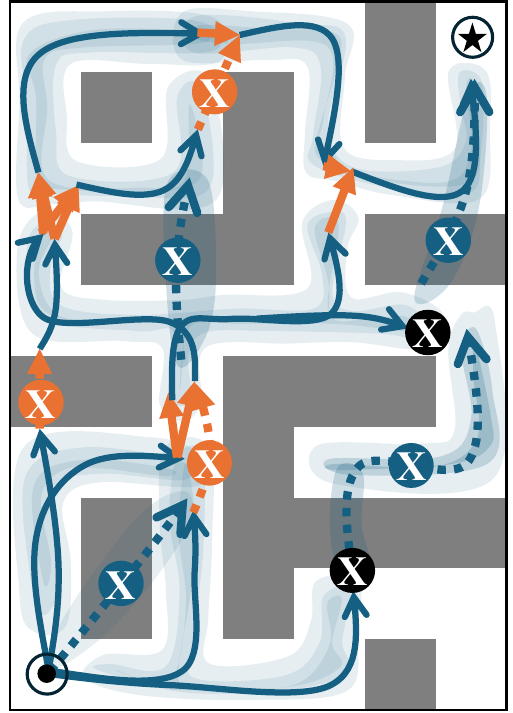}
\label{fig:overview-b}
}
\subfloat[Graph-based search]{
\includegraphics[width=0.23\linewidth]{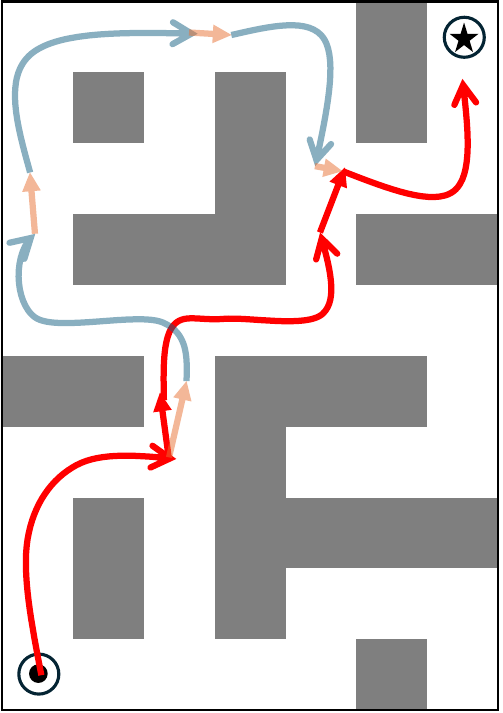}
\label{fig:overview-c}
}
\subfloat[Refining the solution]{
\includegraphics[width=0.23\linewidth]{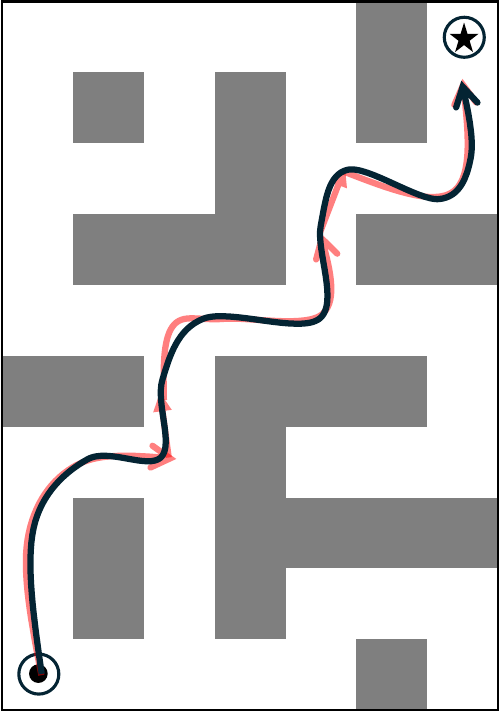}
\label{fig:overview-d}
}
\caption{\textbf{Overview of the SCD pipeline.} Double-dotted lines in \cref{fig:overview-a} represent the initially generated plans. To find the executable plans, (i) we generate plans as in \cref{fig:overview-a}, (ii) evaluate each state connection using our reachability classifier as in \cref{fig:overview-b}, (iii) select a minimum-cost state sequence as in \cref{fig:overview-c}, and (iv) build the bridge and refine the generated trajectory as in \cref{fig:overview-d}.}
\label{fig:overview}
\end{figure}

To handle this issue, \textit{we propose Selective Compositional Diffuser, called SCD}. As shown in \cref{fig:overview}, we reformulate composition as a composable chunk selection problem over the candidate plans. To find a composable chunk sequence, we construct a graph using chunks as vertices. Edges are constructed between chunks and weighted by a score representing whether the next chunk could be \textit{reachable} within a fixed time budget. By traversing the graph, we can find the minimum-cost path, which selects the sequence of chunks that are most likely to be reachable, thereby composable. We connect them with a fixed-length trajectory, often called a \textit{bridge}.

To measure reachability, we present a self-supervised method to train a lightweight MLP network, classifying whether one state can reach the other within a given time lag. Once we find the minimum-cost path with this learned reachability score, we generate bridges and finally stitch the chunks.

We evaluate our method on long-horizon planning tasks from OGBench~\citep{park2025ogbench}. We observe that our method reduces infeasible transitions between composed chunks. Our results outperform baseline diffusion planners and goal-conditioned reinforcement learning methods, particularly on the giant-sized HumanoidMaze task, improving the success rate from the previous state-of-the-art result of 67\% to 83\% in the studied settings.

\section{Related Work}

\textbf{Planning with Diffusion Models.} Diffusion models~\citep{NEURIPS2019_3001ef25, NEURIPS2020_4c5bcfec} have been adopted for decision-making~\citep{pmlr-v162-janner22a, ajay2023is, Chi-RSS-23}. In particular, diffusion planners~\citep{pmlr-v162-janner22a} model complete state-action sequences, enabling inference-time constraints through conditioning and guidance. However, monolithic diffusion planners have been observed to degrade when generating trajectories longer than those learned during training~\citep{luo2025generative}. This is relevant when the available data are primarily short and local trajectories.

\textbf{Compositional Diffusion Planning.} To address this limitation, compositional diffusion planners~\citep{pmlr-v229-mishra23a, luo2025generative} represent a long trajectory through multiple local trajectories. Generative Skill Chaining~\citep{pmlr-v229-mishra23a} composes skill-specific diffusion models through shared boundary variables, while CompDiffuser~\citep{luo2025generative} generates overlapping trajectory segments with bidirectional neighbor conditioning. However, as local trajectory distributions are multimodal, neighboring segments may converge to incompatible modes. Combining these predictions can therefore produce a low-density transition that is supported by neither segment.

Recent methods aim to address this mode-averaging failure. RCD~\citep{lee2026refiningcompositionaldiffusionreliable} guides sampling using reconstruction-based density estimation with overlap consistency, ECD~\citep{sun2026energybased} derives a global energy correction, and CDGS~\citep{ICLR2026_6fbe71b5} searches over combinations of local modes through iterative resampling and pruning. However, we observe a non-negligible portion of infeasible plans in \cref{fig:failure-cases}, failing to find feasibly composable chunk sequences.

\textbf{Graph Search for Long-Horizon Planning.} Graph search has been used to convert locally reliable transitions into long-horizon plans. SoRB~\citep{NEURIPS2019_5c48ff18} constructs a graph over states from the replay buffer and uses a goal-conditioned value function as a cost. In offline settings, GAS~\citep{pmlr-v267-baek25a} and TTGS~\citep{opryshko2026testtime} construct subgoal graphs from offline datasets, and locally plan the searched subgoal sequence using a goal-conditioned policy.

Recently, XDiffuser~\citep{hassidof2026planfirstdiffuselater} constructs a temporal-distance graph from offline states and searches waypoints to guide a compositional diffusion model. Using offline data for graph construction may become incompatible with the current planning problem, and thus our method aims to construct graphs with currently generated trajectories to reduce the offline-online gap (\cref{sec:graph construction}).

\textbf{Reachability and Temporal-Distance Estimation.} Goal-conditioned reinforcement learning often measures how states are connected. C-Learning~\citep{eysenbach2021clearning} measures the state relation by distinguishing future states from independently sampled states with the recursive classifier. Contrastive RL~\citep{eysenbach2022contrastive} also learns this relational metric by contrastive learning, while QRL~\citep{pmlr-v202-wang23al} models the goal-reaching cost as a quasimetric value function. Primarily, they do not aim to directly measure whether two states are reachable in a given time budget.

More directly, \citet{savinov2018episodic} predicts whether two observations occur within a fixed number of environment steps, and \citet{mezghani2022learning} adopts it for offline RL. They use observed trajectories for training datasets, and may suffer from the distributional mismatch over generated samples. This motivates us to use generated samples jointly for our reachability classifier in \cref{sec:discriminator}.

\section{Method}
\label{sec:method}
In this section, we introduce our method for composing the generated chunks by leveraging the learned reachability score. We reformulate composition as a chunk selection problem over candidates, as described in \cref{sec:failure}. We propose the self-supervised method for a lightweight reachability network in \cref{sec:discriminator}, measuring the feasibility of a chunk connection. Then we present the graph-based composition method for reachable and feasible plans in \cref{sec:graph construction,sec:bridge-refinement}.

\subsection{Selection of Composition Mode}
\label{sec:failure}
Given the start state $q_s$ and the goal state $q_g$, the compositional diffuser generates $N$ candidate plans $\mathcal C_{1:N}$, each consisting of $K$ overlapping chunks $\tau$:
\begin{align}
    \mathcal C_n = \tau_{1:K}^{(n)} = (\tau_{1}^{(n)},\ldots,\tau_{K}^{(n)}), \quad \tau_{k}^{(n)} = s_{1:H}^{(n,k)} = (s_{1}^{(n,k)}, \ldots, s_{H}^{(n,k)}),
\end{align}
where $s_{i}^{(n,k)}$ is a state vector. Typically, adjacent chunks are merged into a single trajectory. This operation is based on the assumption that the adjacent chunks on the same route are continuous and composable. However, when they commit to different route modes, merging overlapping regions by averaging cannot achieve a globally feasible plan. As shown in \cref{fig:failure-cases}, it either interpolates the infeasible region or leaves the mode mismatch to the boundary chunks.

The observed failure arises from chunk arrangement, rather than their local generation. To handle this problem, we retain the chunks as a pool of candidates $\tau_i\in \{\tau_{1}^{(1)},\ldots,\tau_{K}^{(N)}\}$ for $\tau_i = s^{(i)}_{1:H}$, and reformulate the composition as a composable chunk selection problem, $\{\tau_i\}_{i=1}^{T}$. Let an assembly $A_i$ be a subpart of the chunk:
\begin{align}\label{eq:assembly}
A_{i}=\tau_i[\ell_i\mathord{:}r_i]=s_{\ell_i:r_i}^{(i)}.
\end{align}
We aim to select assemblies which are most likely to be executable when connected, while penalizing an unnecessarily long horizon:
\begin{align}\label{eq:assembly-chain}
A^* = \underset{A \in \mathcal A(\mathcal C;q_s,q_g)} {\argmax} p(E|A) \gamma^{T(A)}, \quad 0<\gamma \leq 1,
\end{align}
where $\mathcal A(\mathcal C; q_s, q_g)$ is a set of assembly chains, $A=(A_1, \ldots, A_T)$ for $s_{\ell_1}^{(1)}=q_s$ and $s_{r_T}^{(T)} = q_g$, $T(A)$ is the length of the plan generated by the chain $A$, and $p(E|A)$ denotes the executability of the given chain $A$. We use a learned reachability score to estimate local executability as described in \cref{sec:discriminator}, and propose the optimal selection strategy in \cref{sec:graph construction}.

\subsection{Measuring Reachability}
\label{sec:discriminator}
To be an executable trajectory, (i) we suppose that adjacent boundary states of consecutive chunks should be reachable, and (ii) we generate bridging trajectories for composition rather than overlap averaging. Previous work~\citep{mezghani2022learning} trains a classifier to represent reachability by using temporally close states as positive pairs and independently sampled or temporally distant states as negative pairs. Both pairs are sampled from the offline dataset; however, our classifier is required to evaluate generated endpoints, which may lie in sparse or unseen regions of offline trajectory distributions. Predictions on these endpoints can therefore be overconfident, or confuse an infeasible shortcut with a reachable path, as shown in \cref{fig:heatmap-comparison,tab:classifier-ablation}.

To address this distribution mismatch, we construct a negative dataset as a union of the offline data and the generated data, given by the same generator used in the planning. In this way, the classifier is fit where it will be queried. However, this approach can lead to a second issue: the generated states are not uniformly distributed, and the reachability score can be biased by the generative density (Eq.~\ref{eq:biased-logit}). Therefore, we approximately estimate this generative density to reweight the negative term in \cref{eq:loss}, leaving the score depending on the offline data distribution while reducing the bias effect.

For a randomly sampled start state $x_{\rm obs}=s_t$ and lag $L$, we sample the reached state $y_{\rm obs}\sim \mathcal U\{s_{t+l}\}_{l=1}^L$ and sample negative states $x_{\rm neg},y_{\rm neg}$ independently from a negative density $p_\mathrm{neg}$. Let $p_{\rm obs}(x, y|L)$ denote the joint density of the reachable state pair under the behavior policy for $L$ steps. We train a classifier network $f_\phi$ to distinguish observed $L$-step transitions from negative ones:
\begin{align}\label{eq:network}
D_\phi(x,y,L) = \sigma(f_\phi(x,y,L)), \quad f_\phi(x, y, L) = Q_\psi(x, \mathrm{PE}(L))^\top\frac{K_\theta(y)}{\|K_\theta(y)\|_2},\quad \phi = \{\psi, \theta\},
\end{align}
where $\sigma$ is a sigmoid function and $\mathrm{PE}$ is a sinusoidal positional encoding. In practice, we use a 3-layer MLP network for the query encoder $Q_\psi$ and key encoder $K_\theta$.

We consider a weighted binary cross-entropy objective,
\begin{align}
\label{eq:loss}
&\mathcal L((x_{\rm obs} ,y_{\rm obs}),(x_{\rm neg},y_{\rm neg}),L) \nonumber \\
&= - \log D_\phi(x_{\rm obs},y_{\rm obs},L) - w(x_{\rm neg})w(y_{\rm neg})\log (1-D_\phi(x_{\rm neg},y_{\rm neg},L)),
\end{align}
for a scalar weight $w(x)>0$. Minimizing this loss gives the optimal classifier:
\begin{align}
D^*(x,y,L) &= \frac{p_{\rm obs}(x,y|L)}{p_{\rm obs}(x,y|L) + w(x)w(y)p_{\rm neg}(x)p_{\rm neg}(y)}, \\
f^*(x,y,L) &= \log p_{\rm obs}(x,y|L) - \log (w(x)w(y)p_{\rm neg}(x)p_{\rm neg}(y)) \label{eq:biased-logit}.
\end{align}
Without reweighting, the logit contains the term including $p_{\rm neg}(x)$ and $p_{\rm neg}(y)$, biasing the model toward the distribution of the negative dataset. To reduce this bias, we estimate the density of the negative dataset with a $k$-NN density estimator:
\begin{align}
\hat p_{\rm neg}(x) = \frac k {N_{\rm neg}V_dr_k(x)^d} \propto \frac{1}{r_k(x)^d},
\end{align}
where $N_{\rm neg}$ is the dataset size, $V_d$ is the volume of the unit ball in $d$-dimensional state space, and $r_k(x)$ is the distance from $x$ to its $k$-th nearest neighbor in the negative state dataset.

Then, we use this estimated density for the weight $w(x)$:
\begin{align}
w(x) = \operatorname{clip}\left(r_k(x)^d / \overline{r_k^d},\ w_{\min},\ w_{\max}\right),
\qquad \overline{r_k^d} = \mathbb E_{x \sim p_{\rm neg}} \left[r_k(x)^d\right],
\end{align}
where the clipping bounds the weight on unreliably estimated densities. Then $w(x) \propto  1/\hat p_{\rm neg}(x)$ for reliably estimated supports and the weighted density is approximately constant in this space. Thus,
\begin{align}
f^*(x,y,L) &\approx \log p_{\rm obs}(x) + \log p_{\rm obs}(y|x,L) - \log C,
\end{align}
where $C$ is a constant. This pushes the classifier to predict reachability only with offline distribution, rather than with generative distribution. The score can jointly measure two densities, whether the start state is observed in the offline distribution, $\log p_{\rm obs}(x)$, and whether the transition is reliable, $\log p_{\rm obs}(y|x, L)$. We use it as the reachability score when constructing the graph in \cref{sec:graph construction}.

We can also interpret this score as a lower bound on local executability. Let $E$ be the event that $(x,y)$ is executable within $L$, and $X$ be the event that the pair is an observed $L$-or-fewer-step pair. A pair from the offline data is obviously executable, hence $X\subseteq E$. Since $D^*$ is the posterior of $X$, the reachability score now can be viewed as a lower bound on executability $E$:
\begin{align}
D^*(x,y,L)= p(X|x,y,L) \le p(E|x,y,L).
\end{align}

\subsection{State Chunk Graph Construction}
\label{sec:graph construction}
To find an optimal assembly chain $A^*$ in \cref{eq:assembly-chain}, we construct a directed graph to select composable chunks. Candidate boundaries $\ell_i$ and $r_i$ in \cref{eq:assembly} include chunk endpoints $\ell_i = 1$ and $r_i = H$ following \citet{yoon2025compositional}, and we extend them to interior points $\ell_i\ge 1$ and $r_i\le H$ for more flexible chunk selection. Each selected segment $\tau_i[\ell_i\mathord{:}r_i]$ is preserved as generated, while the connections between the segments, namely bridges, will be generated after the graph search.

\begin{figure}[t]
\centering
\includegraphics[width=\linewidth]{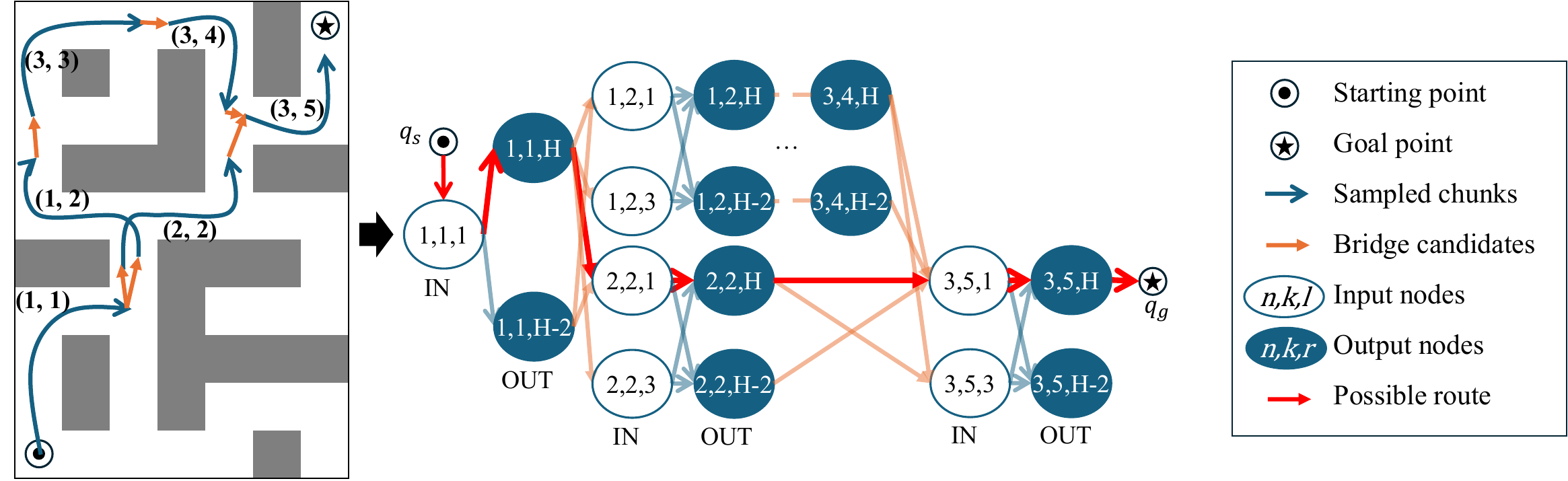}
\caption{An example of graph construction. This example follows the setting of \cref{fig:overview-b}.}
\label{fig:graph-construction}
\end{figure}

As shown in \cref{fig:graph-construction}, we set the candidate boundaries at a fixed stride, such as $\ell\in \{1, 3, 5, \ldots\}$. We represent boundary states of each segment $\tau^{(n)}_k[\ell\mathord{:}r]$ with two vertices, $\mathrm{IN}(n, k, \ell)$ and $\mathrm{OUT}(n, k,r)$. For generated chunks, an edge between these vertices represents the segment:
\begin{align}\label{eq:edge-segment}
\tau^{(n)}_k[\ell\mathord{:}r]: {\rm IN}(n,k,\ell) \longrightarrow {\rm OUT}(n,k,r), \quad r-\ell\geq M,
\end{align}
where $M$ is the chunk length excluding overlapping regions on both sides. We add additional candidate edges for bridges, connecting:
\begin{align}\label{eq:edge-bridge}
{\rm OUT}(n,k,r) \longrightarrow {\rm IN}(n',k',\ell')
\end{align}
for $(n,k)\neq (n',k')$. We can evaluate the edge between two consecutive vertices $v$ and $v'$ by computing the reachability score of adjacent boundary states, regularized by the length penalty:
\begin{align}
    c(v, v') = -\log(D_\phi(v, v',h_o)) - h_o \log \gamma,
\end{align}
where $\gamma \in (0, 1]$ is a discount factor and $h_o$ is a time budget between two vertices. We use $h_o=r-\ell$ for the segment and a preset constant for the bridge. The reachability score of the segment in \cref{eq:edge-segment} is $D_\phi(s^{(n, k)}_{\ell}, s^{(n, k)}_r, r - \ell)$, and the score of the bridge in \cref{eq:edge-bridge} is $D_\phi(s^{(n,k)}_r,s^{(n',k')}_{\ell'}, h_o)$.

Then we reorganize an assembly chain as a sequence of vertices $A = (v_0 = q_s, v_1, \ldots, v_{J-1}, v_J = q_g)$ that form a directed path from $q_s$ to $q_g$. We can interpret the executability of $A$ as a product of the executability on each element:
\begin{align}
p(E|A) = \prod_{j=0}^{J-1}p(E|v_j, v_{j+1}, h_o(v_j, v_{j+1})) \ge \prod^{J-1}_{j=0} D_\phi(v_j,v_{j+1},h_o(v_j, v_{j+1})),
\end{align}
where $p(E|v_j, v_{j+1}, h_o)$ is a likelihood of the executability between two states $v_j$ and $v_{j+1}$ within a time budget $h_o$ and $h_o(v_j, v_{j+1})$ is a time budget between two vertices. If we assume that the planner can generate a feasible local plan between reachable state pairs within a given time budget, the executability of the full assembly chain $A$ can be decomposed, thereby inducing the lower bound.

As we prefer short paths, the length-regularized executability becomes:
\begin{align}
\gamma^{T(A)} \cdot p(E|A) \geq \prod^{J-1}_{j=0} \gamma^{h_o(v_j, v_{j+1})}D_\phi(v_j, v_{j+1}, h_o(v_j, v_{j+1})),
\end{align}
which coincides with the objective in \cref{eq:assembly-chain}. Moreover, this is consistent with our cost function in log scale. Therefore, we maximize this decomposed lower bound by minimizing the cost:
\begin{align}\label{eq:solution}
    A^* = \underset{A}{\arg \min} \sum_{j=0}^{J-1} c(v_j,v_{j+1}).
\end{align}
Since the cost is non-negative, we can use Dijkstra's algorithm to find the minimum-cost path even if the graph contains cycles. We report the wall-clock times of the studied baselines in \cref{tab:time-comparison}.

\subsection{Bridge Generation and Refinement}
\label{sec:bridge-refinement}
The solution path from \cref{eq:solution} specifies which segments to retain and to connect. For each bridge $\beta_i$, we provide the adjacent segments as the left and right references including $s^{(i)}_{r_i}$ and $s^{(i+1)}_{\ell_{i+1}}$ for the diffusion planner, then generate the trajectory with length $h_o$. We concatenate the bridges with the segments to form a full trajectory $\tilde \tau$.

Although graph selection identifies reachable connections, it does not guarantee stable motion across the segments. Therefore, we apply a refinement to jointly adjust the segments and bridges. We perturb the full trajectory $\tilde \tau$ with small noise and re-denoise the remaining steps, while inpainting the start and goal states. The trajectory is covered with partially overlapping windows, and we denoise each window in parallel, merging the final outputs into a single trajectory. Within the overlapping band, neighboring predictions are combined by a linear interpolation:
\begin{align}\label{eq:interpolation}
\hat s_i = \left(1-\frac{i}{h_o-1}\right)\hat s_i^{\rm left}+ \left( \frac i {h_o-1}\right)\hat s_i^{\rm right}, \quad i=0,\ldots , h_o-1
\end{align}
where $\hat s_i^{\rm left} $ and $\hat s_i^{\rm right}$ are the refined windows for the same overlap position. For stable generation, we follow the sizes of the window and the overlapping band in \citet{luo2025generative}.

\section{Experiments}

In this section, we evaluate our method on long-horizon composition planning tasks from OGBench~\citep{park2025ogbench}. Our experiments focus on three aspects: (1) whether our method reduces infeasible transitions between composed trajectory segments, (2) whether improved plan feasibility leads to higher task success rate as the planning horizon and dynamic complexity increase, and (3) how each component accounts for the observed improvements.

\textbf{Datasets and Environments.}
We consider the stitch datasets of PointMaze, AntMaze, and HumanoidMaze, each evaluated on the Medium, Large, and Giant maze layouts. The stitch datasets consist of short trajectory segments that do not individually connect the start and goal states of evaluation settings. Therefore, solving these tasks requires composing locally observed behaviors into much longer trajectories.

\textbf{Baselines.} We compare against offline goal-conditioned RL (GCRL) and diffusion-based compositional planning methods. The offline GCRL baselines include goal-conditioned behavioral cloning~\citep[GCBC,][]{lynch2020learning}, goal-conditioned implicit V-learning and Q-learning~\citep[GCIVL, GCIQL,][]{kostrikov2022offline}, Quasimetric RL~\citep[QRL,][]{pmlr-v202-wang23al}, Contrastive RL~\citep[CRL,][]{eysenbach2022contrastive}, and Hierarchical Implicit Q-Learning~\citep[HIQL,][]{park2023hiql}. The compositional planning baselines consist of Generative Skill Chaining~\citep[GSC,][]{pmlr-v229-mishra23a}, CompDiffuser~\citep[CD,][]{luo2025generative}, CDGS~\citep{ICLR2026_6fbe71b5}, ECD~\citep{sun2026energybased}, and RCD~\citep{lee2026refiningcompositionaldiffusionreliable}. We use the same pretrained compositional diffusion model and evaluate SCD on OGBench, following the same evaluation protocol for all diffusion-based methods.

\subsection{Evaluating Plan Feasibility}

\begin{table}[t]
\caption{\textbf{Success rates (\%) on OGBench Stitch maze tasks.} Each environment provides 5 goal-conditioned tasks and we evaluate 20 episodes per task for three random seeds.}
\label{tab:ogbench-stitch}
\begin{center}
\resizebox{1.0\linewidth}{!}{
\begin{tabular}{ll|ccccccccc|c}
\toprule
\multirow{2}{*}[-0.53ex]{\textbf{Family}}
& \multirow{2}{*}[-0.53ex]{\textbf{Method}}
& \multicolumn{3}{c}{\textbf{PointMaze}}
& \multicolumn{3}{c}{\textbf{AntMaze}}
& \multicolumn{3}{c|}{\textbf{HumanoidMaze}}
& \multirow{2}{*}[-0.53ex]{\textbf{Avg.}}

\\
\cmidrule{3-11}
&
& \textbf{Medium} & \textbf{Large} & \textbf{Giant}
& \textbf{Medium} & \textbf{Large} & \textbf{Giant}
& \textbf{Medium} & \textbf{Large} & \textbf{Giant} 

\\
\midrule

\multirow{6}{*}{\shortstack[l]{Offline\\GCRL}}
& GCBC
& $23{\pm}18$ & $7{\pm}5$ & $0{\pm}0$
& $45{\pm}11$ & $3{\pm}3$ & $0{\pm}0$
& $29{\pm}5$ & $6{\pm}3$ & $0{\pm}0$ 
& $12.56$ \\

& GCIVL
& $70{\pm}14$ & $12{\pm}6$ & $0{\pm}0$
& $44{\pm}6$ & $18{\pm}2$ & $0{\pm}0$
& $12{\pm}2$ & $1{\pm}1$ & $0{\pm}0$ 
& $17.44$ \\

& GCIQL
& $21{\pm}9$ & $31{\pm}2$ & $0{\pm}0$
& $29{\pm}6$ & $7{\pm}2$ & $0{\pm}0$
& $12{\pm}3$ & $0{\pm}0$ & $0{\pm}0$
& $11.11$ \\

& QRL
& $80{\pm}12$ & $84{\pm}15$ & $50{\pm}8$
& $59{\pm}7$ & $18{\pm}2$ & $0{\pm}0$
& $18{\pm}2$ & $3{\pm}1$ & $0{\pm}0$
& $34.67$ \\

& CRL
& $0{\pm}1$ & $0{\pm}0$ & $0{\pm}0$
& $53{\pm}6$ & $11{\pm}2$ & $0{\pm}0$
& $71{\pm}3$ & $6{\pm}1$ & $0{\pm}0$
& $15.67$ \\

& HIQL
& $74{\pm}6$ & $13{\pm}6$ & $0{\pm}0$
& $94{\pm}1$ & $67{\pm}5$ & $21{\pm}2$
& $\mathbf{96{\pm}4}$ & $31{\pm}3$ & $12{\pm}2$
& $45.33$ \\

\midrule

\multirow{6}{*}{\shortstack[l]{Diffusion\\planners}}
& GSC
& $\mathbf{100{\pm}0}$ & $\mathbf{100{\pm}0}$ & $29{\pm}3$
& $\mathbf{97{\pm}2}$ & $66{\pm}2$ & $20{\pm}1$
& ${92{\pm}1}$ & $70{\pm}3$ & $5{\pm}1$
& $64.33$ \\

& CD
& $\mathbf{100{\pm}0}$ & $\mathbf{100{\pm}0}$ & $68{\pm}3$
& $96{\pm}2$ & ${86{\pm}2}$ & $65{\pm}3$
& ${91{\pm}1}$ & ${72{\pm}3}$ & $67{\pm}4$
& $82.78$ \\

& CDGS
& $\mathbf{100{\pm}0}$ & $\mathbf{100{\pm}0}$ & $87{\pm}3$
& $\mathbf{97{\pm}1}$ & $88{\pm}2$ & ${85{\pm}3}$
& $93{\pm}1$ & $74{\pm}2$ & $55{\pm}3$
& $86.56$ \\

& ECD
& $\mathbf{100{\pm}0}$ & $\mathbf{100{\pm}0}$ & ${84{\pm}2}$
& $\mathbf{97{\pm}1}$ & $82{\pm}1$ & $82{\pm}6$
& ${92{\pm}2}$ & $64{\pm}4$ & ${49{\pm}1}$
& $83.33$ \\

& RCD
& $\mathbf{100{\pm}0}$ & $\mathbf{100{\pm}0}$ & $\mathbf{100{\pm}0}$
& $\mathbf{97{\pm}2}$ & $91{\pm}2$ & $\mathbf{89{\pm}2}$
& ${93{\pm}2}$ & $79{\pm}3$ & ${62{\pm}3}$
& $90.11$ \\

& \cellcolor[HTML]{efefef}\textbf{Ours}
&  \cellcolor[HTML]{efefef}${\mathbf{100{\pm}0}}$
&  \cellcolor[HTML]{efefef}$\mathbf{100{\pm}0}$
&  \cellcolor[HTML]{efefef}$91{\pm}1$
&  \cellcolor[HTML]{efefef}${95{\pm}1}$
&  \cellcolor[HTML]{efefef}$\mathbf{92{\pm}1}$
&  \cellcolor[HTML]{efefef}$85{\pm}2$
&  \cellcolor[HTML]{efefef}$95{\pm}1$
&  \cellcolor[HTML]{efefef}$\mathbf{89{\pm}2}$
&  \cellcolor[HTML]{efefef}$\mathbf{83{\pm}3}$
&  \cellcolor[HTML]{efefef}$\mathbf{92.22}$ \\

\bottomrule
\end{tabular}
}
\end{center}
\end{table}
\begin{figure}[t]
\centering
\vspace*{-1em}
\subfloat[CompDiffuser]{%
    \includegraphics[width=0.185\linewidth]{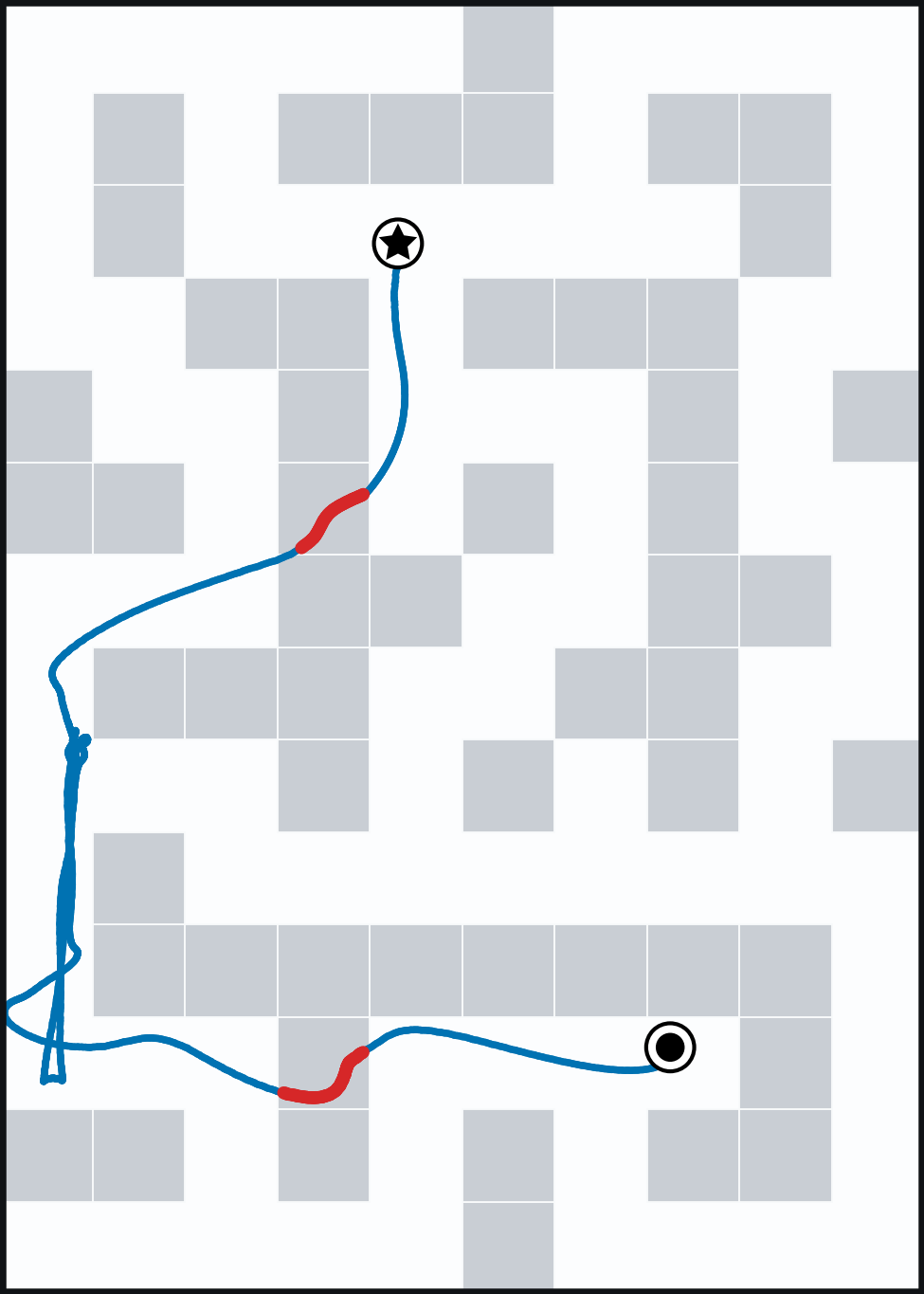}
}
\subfloat[CDGS]{%
    \includegraphics[width=0.185\linewidth]{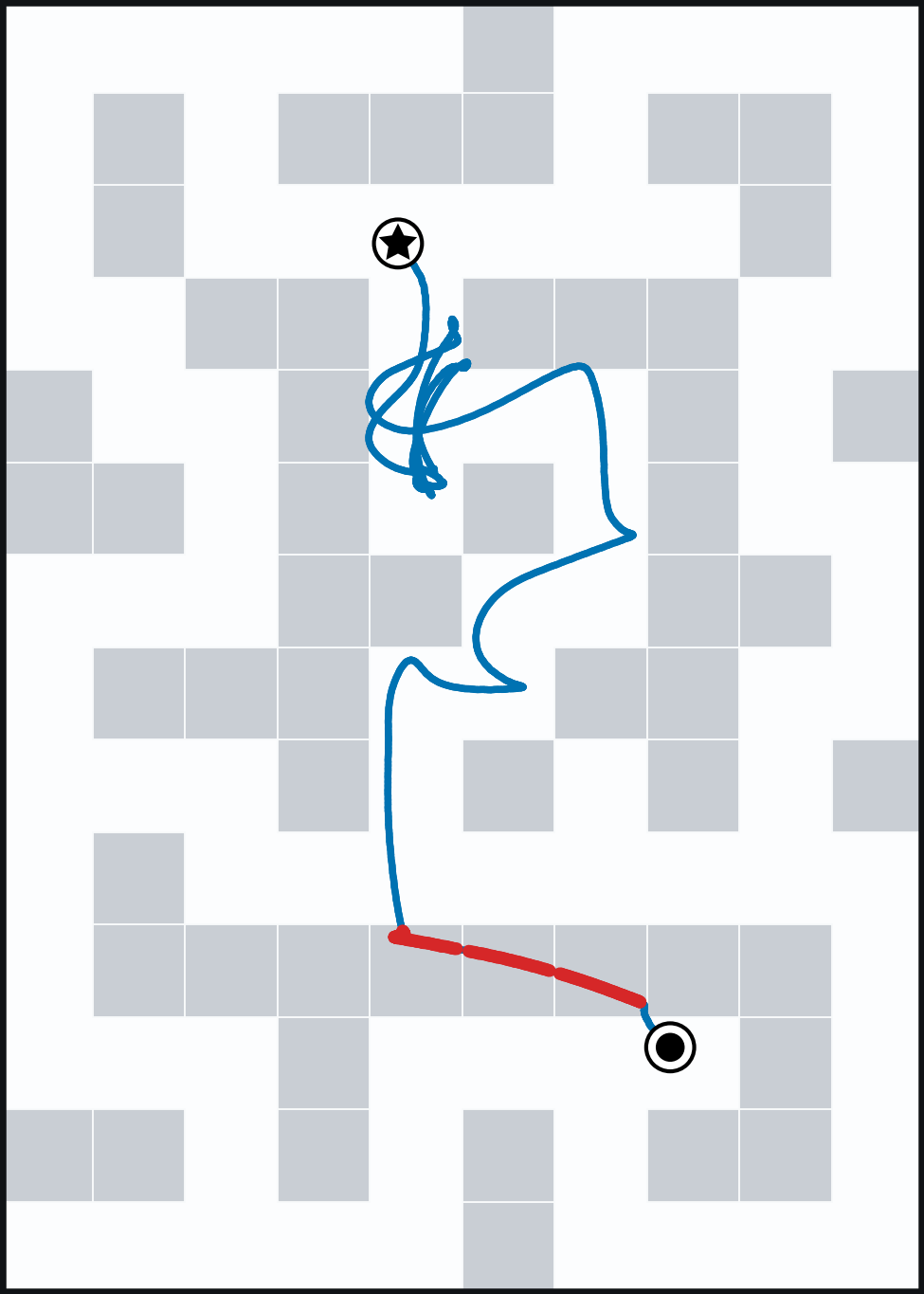}
}
\subfloat[ECD]{%
    \includegraphics[width=0.185\linewidth]{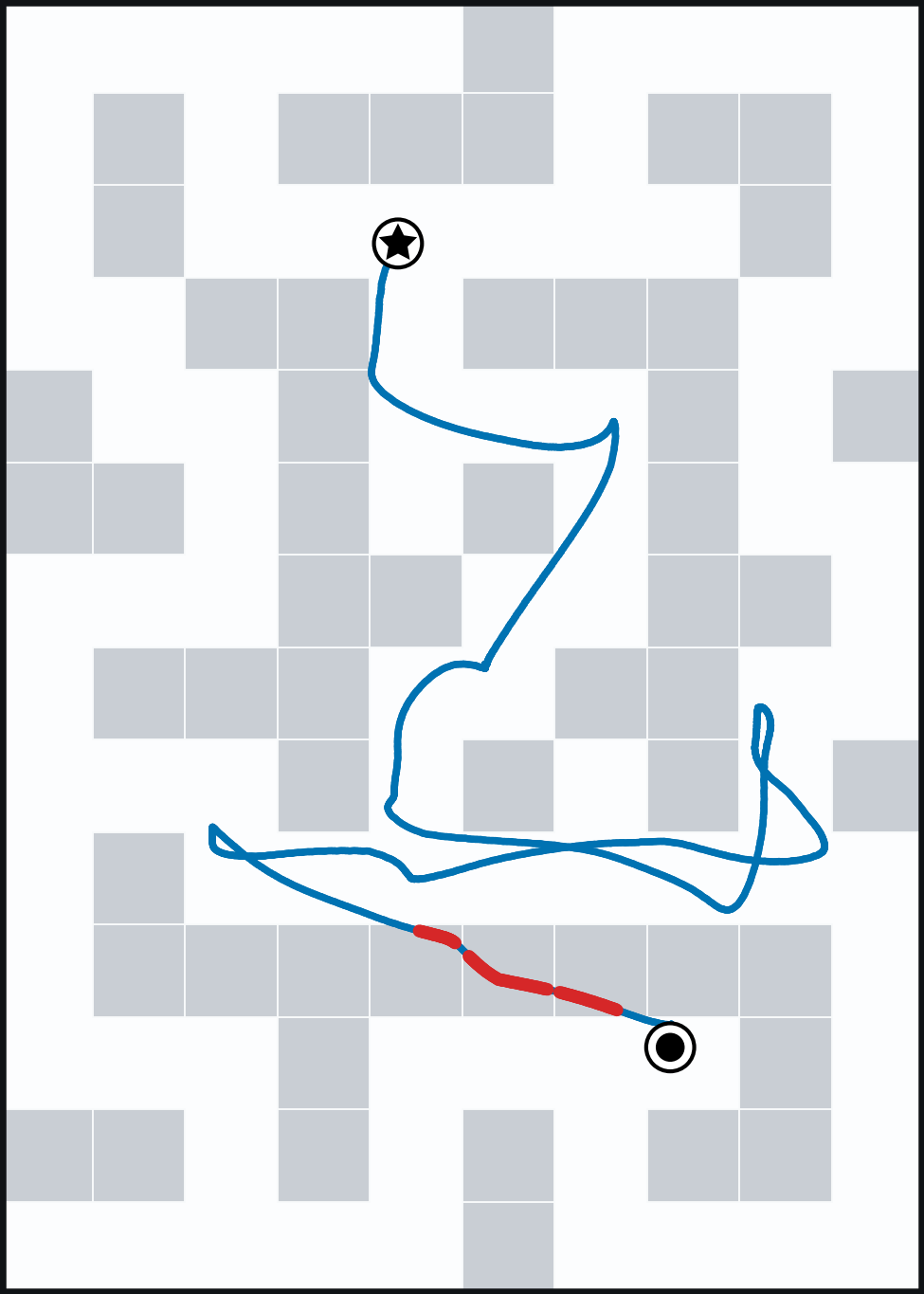}
}
\subfloat[RCD]{%
    \includegraphics[width=0.185\linewidth]{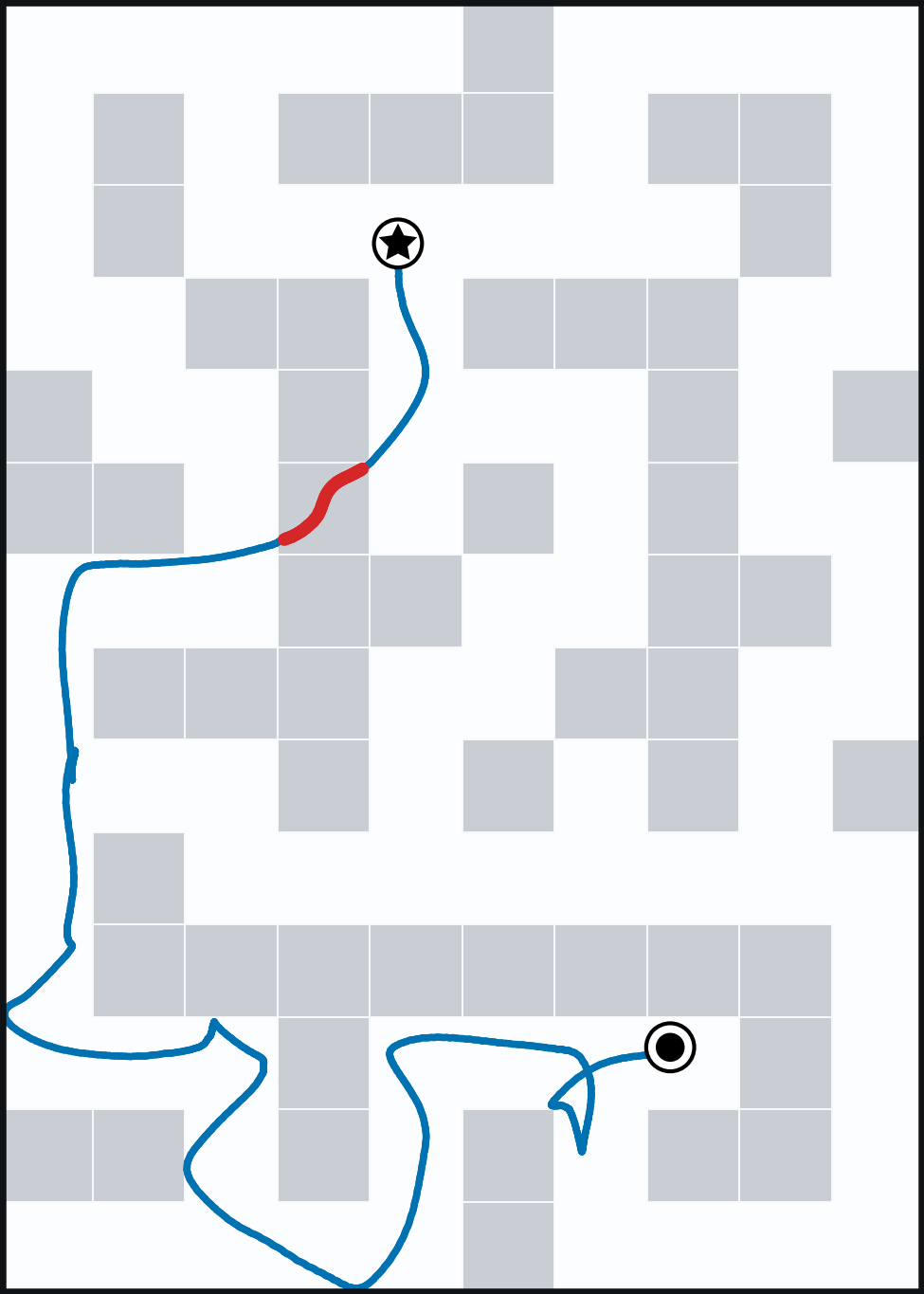}
}
\subfloat[\textbf{SCD (Ours)}]{%
    \includegraphics[width=0.185\linewidth]{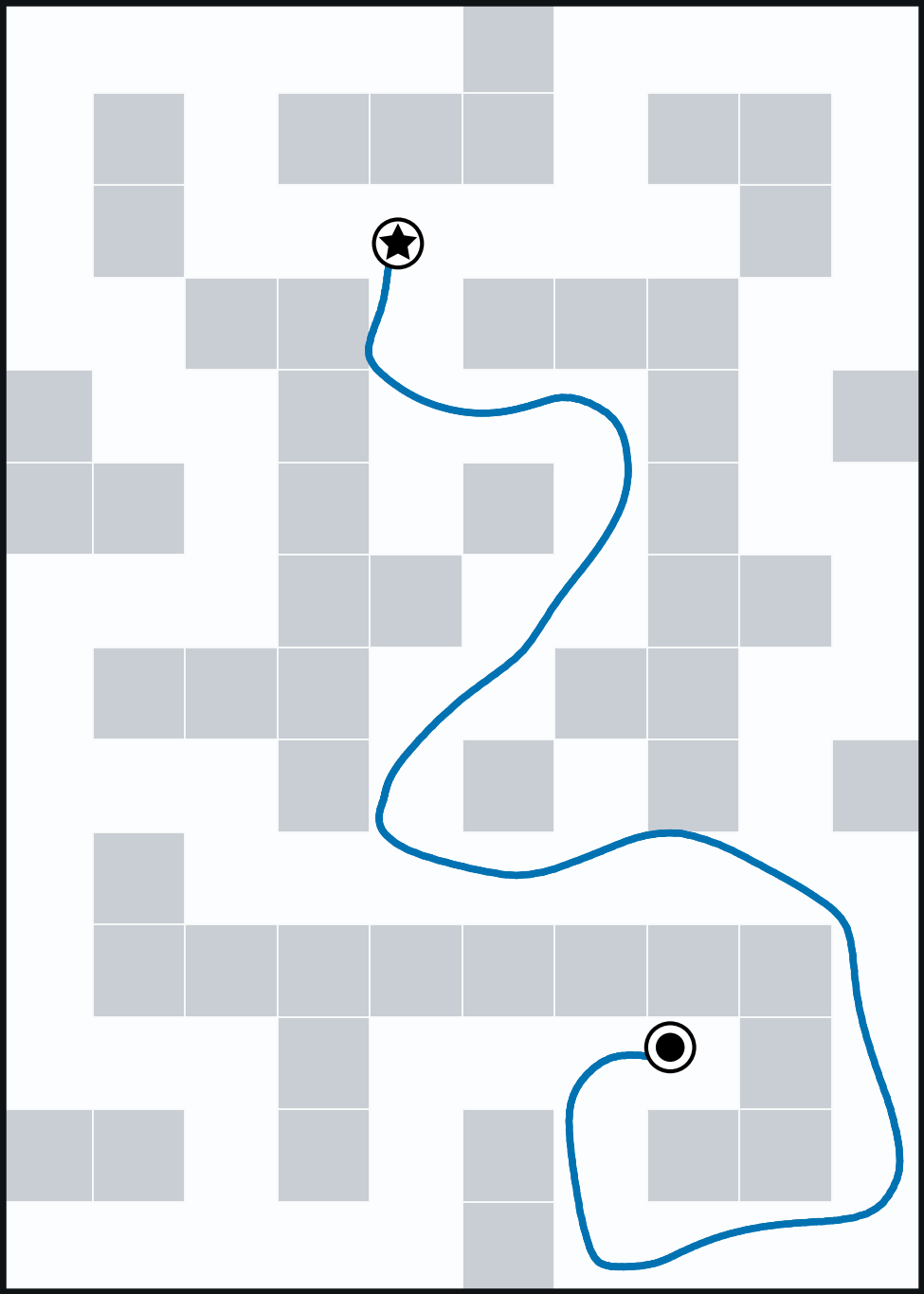}
}
\caption{\textbf{Qualitative results of long-horizon planning in HumanoidMaze-Giant.} SCD successfully finds feasible connections among the chunk candidates. See \cref{fig:more-qualitative-result} for more results.}
\label{fig:qualitative-result}
\end{figure}

\begin{wraptable}{r}{0.45\linewidth}
\vspace{-1em}
\centering
\caption{\textbf{The ratio of geometrically feasible plans.} For each HumanoidMaze-Giant task, we sample 20 plans and report the fraction whose trajectories are geometrically feasible, not passing through walls. The numerical entries in the header row denote the task IDs.}
\label{tab:feasibility-ratio}
\resizebox{\linewidth}{!}{%
\begin{tabular}{l|ccccc|c}
\toprule
\textbf{Method} & \textbf{01} & \textbf{02} & \textbf{03} & \textbf{04} & \textbf{05} & \textbf{Avg.} \\
\midrule
 CD & $30$ & $5$ & $15$ & $10$ & $40$ & $20$\\
 CDGS & $0$ & $30$ & $0$ & $0$ & $45$ & $15$\\
 ECD & $45$ & $45$ & $20$ & $25$ & $50$ & $37$\\
 RCD & $25$ & $30$ & $40$ & $10$ & $60$ & $33$\\
 \midrule
 \textbf{SCD (Ours)} & $\mathbf{95}$ & $\mathbf{80}$ & $\mathbf{55}$ & $\mathbf{75}$ & $\mathbf{75}$ & $\mathbf{76}$\\
\bottomrule
\end{tabular}
}
\vspace{-3em}
\end{wraptable}

First, we examine whether our method produces feasible trajectories. Following \citet{lee2026refiningcompositionaldiffusionreliable}, we sample 20 plans for the five tasks of OGBench HumanoidMaze-Giant-Stitch, and count the plans staying outside the wall without making any shortcuts. This environment requires at least four-times longer horizons than PointMaze and AntMaze, and it is comparatively hard to retain mode consistency. As shown in \cref{tab:feasibility-ratio} and the qualitative result \cref{fig:qualitative-result}, our method consistently improves the feasibility, and this gain holds for all reported tasks.

\subsection{Solving Long-Horizon Tasks}
We examine whether our method remains effective as the required composition horizon increases. We consider OGBench Stitch tasks of PointMaze, AntMaze, and HumanoidMaze across Medium, Large, and Giant maze sizes. Larger mazes require longer paths and more local segments to be composed. This may increase the chance of generating an infeasible connection.

In \cref{tab:ogbench-stitch}, offline GCRL methods show degraded performance as the maze size increases. Diffusion-based planners perform better across most tasks, nearly saturated on PointMaze-Medium and Large. However, their performance declines on longer compositions, particularly in HumanoidMaze-Giant.

Our method remains competitive in the saturated settings while improving certain long-horizon tasks. It achieves success rates of $91\%$, $85\%$, and $83\%$ in Giant PointMaze, AntMaze, and HumanoidMaze tasks, respectively. These results support that our reachability score reliably distinguishes composable segment pairs and that graph-based assembly produces plans with more likely connections.

\subsection{Adaptive Planning Length and Planning Cost}

\begin{figure}[t]
\centering
\includegraphics[width=\linewidth]{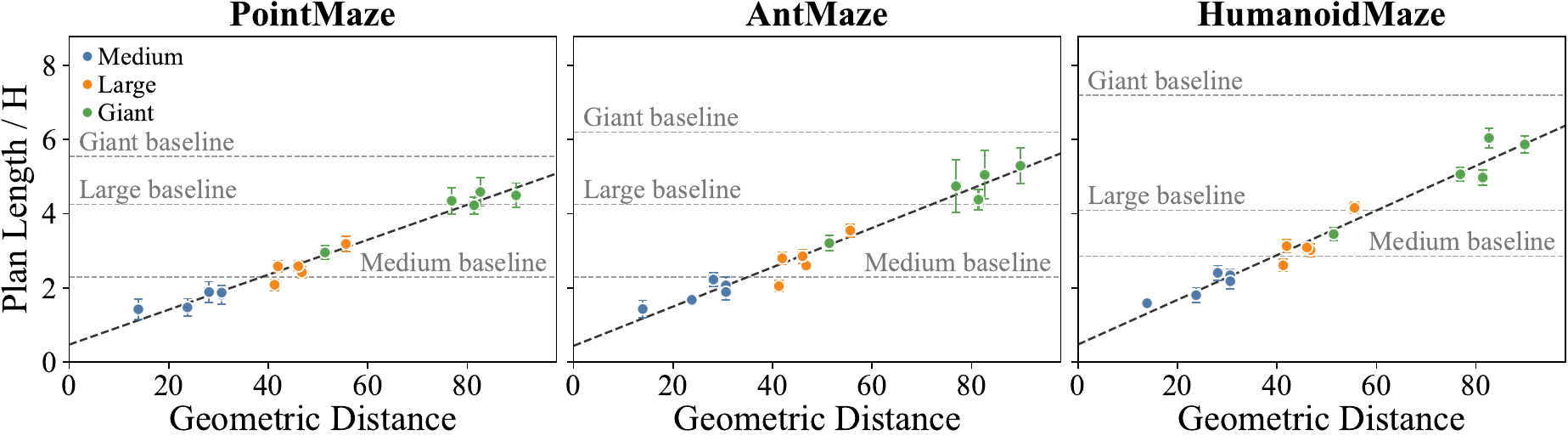}
\caption{\textbf{Adaptive plan length of SCD.} Each point represents one evaluation task, and five tasks are evaluated for each maze size. The x-axis represents a geometric distance of the shortest path and the y-axis represents the length of the generated plan divided by the planner horizon ($H$). Horizontal dashed lines indicate the fixed number of chunks used by compositional planners. Error bars show the standard deviation over 20 plans.}
\label{fig:adaptive-planning}
\end{figure}

The appropriate trajectory length can vary across start and goal state pairs depending on their distances and difficulty. Our method supports adaptive planning by selecting the chunks based on the executability and the trajectory length, which can dynamically alter the number of selected chunks.

As shown in \cref{fig:adaptive-planning}, the resulting trajectory lengths closely follow the distance of the shortest path between the start and goal states, regardless of the size of the environment. The length of plans depends on the distance rather than a predefined number of chunks, and SCD can find an appropriate length for each task to execute.

\begin{wraptable}{r}{0.55\linewidth}
\centering
\vspace{-1.2em}
\captionof{table}{\textbf{Planning cost comparison.} Wall-clock planning times and success rates on AntMaze-Giant and HumanoidMaze-Giant. We report mean time $\pm$ std across 10 plans on a single RTX 4090.}
\label{tab:time-comparison}
\resizebox{\linewidth}{!}{%
\setlength{\tabcolsep}{3.5pt}
\begin{tabular}{c|cr|cr}
    \toprule
    \multirow{2}{*}[-0.53ex]{\textbf{Method}} & \multicolumn{2}{c|}{\textbf{AntMaze}} & \multicolumn{2}{c}{\textbf{HumanoidMaze}} \\
    \cmidrule{2-5} & \textbf{Succ.} & \multicolumn{1}{c|}{\textbf{Time(s)}} & \textbf{Succ.} & \multicolumn{1}{c}{\textbf{Time(s)}} \\
    \midrule
    CD (parallel) & $48{\pm}1$ & $3.09{\pm} 0.02$ & $57{\pm}3$ & $7.58 {\pm} 0.02$ \\
    CD (interleave) & $65{\pm}3$ & $6.31 {\pm} 0.01$ & $67{\pm}4$ & $13.08 {\pm} 0.03$ \\
    \midrule
    CDGS & $85{\pm}3$ & $15.83{\pm}0.03$ & $55{\pm}3$&$20.33 {\pm} 0.03$ \\
    ECD & $82{\pm}6$ & $9.24{\pm}0.01$ & $49{\pm}1$ & $18.71 {\pm} 0.03$ \\
    RCD & $\mathbf{89{\pm}2}$ & $4.52{\pm}0.01$ & $62{\pm}3$ & $11.07 {\pm} 0.04$ \\
    \textbf{SCD (Ours)} & $85{\pm}2$ & $\mathbf{4.24{\pm}0.01}$ & $\mathbf{83{\pm}3}$ & $\mathbf{9.08 {\pm} 0.04}$ \\
    \bottomrule
\end{tabular}
}
\end{wraptable}

Beyond sampling candidate trajectories, our method evaluates candidate connections, searches for the minimum-cost path, and rebuilds the bridges. To quantify the overhead, we measure and compare wall-clock planning time on AntMaze-Giant and HumanoidMaze-Giant with a single RTX 4090. We use the same pretrained compositional model for every method in \cref{tab:time-comparison}. We observe that connection scoring and graph search require a small additional time cost: each candidate is encoded once, and the reachability model evaluates all state pairs through a single matrix multiplication. Only the selected bridges require additional denoiser calls, and they cost much less than generating whole candidates. Since we produce candidates in parallel, SCD runs faster than the listed methods except for CD (parallel).

\subsection{Ablation Study}

Since a single infeasible connection may ruin the entire trajectory, avoiding false-positive edges is important for executable routes. We therefore evaluate whether the classifier retains reachable states and rejects unreachable ones. 

We examine three design choices: using generated states, density reweighting, and the neighborhood size $k$ used by the $k$-NN density estimator. We evaluate each classifier on a controlled test set. Further details on the evaluations are provided in \cref{app:reachability-evaluation}.

\begin{figure}[t]
\centering
\includegraphics[width=\linewidth]{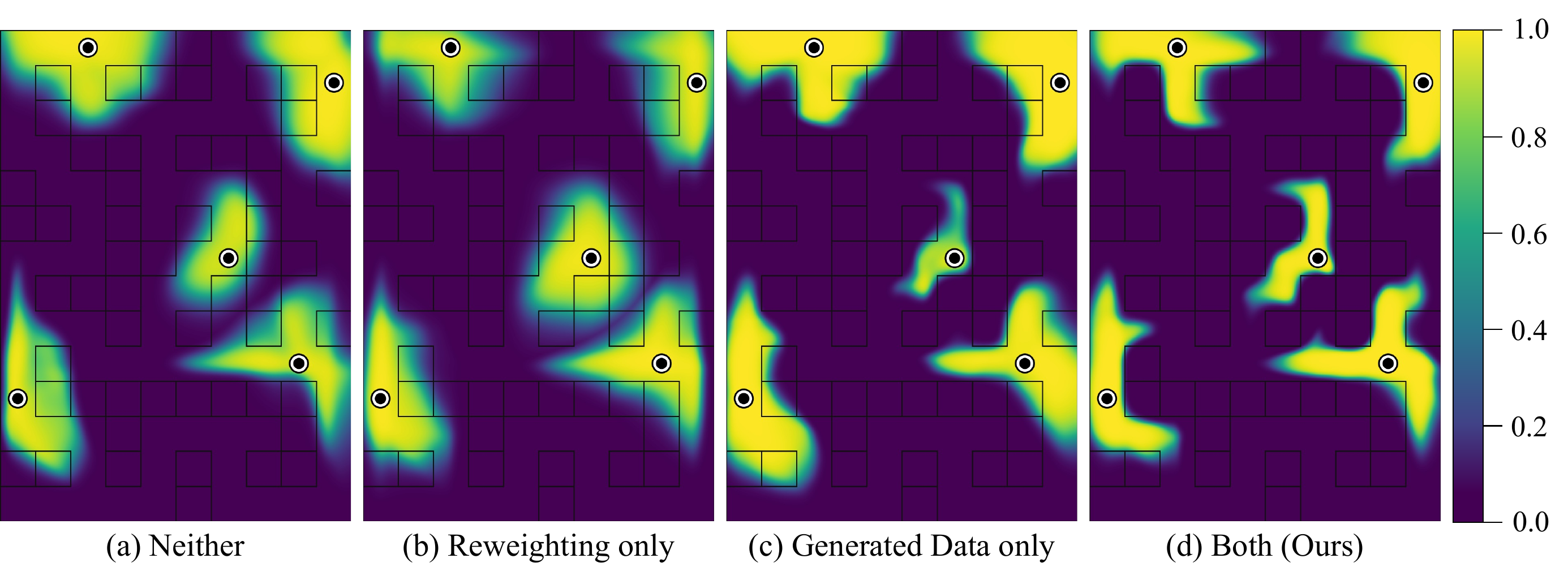}
\caption{\textbf{Reachability heatmap comparison.} $D_\phi(x,y,h_o)$ is evaluated for five anchor states (circles) in AntMaze-Giant, under four combinations of the ablation study in \cref{tab:classifier-ablation}. Including generated states reduces overconfident predictions in regions that are unseen in the offline dataset. Density reweighting removes the start-state bias induced by nonuniform sampling. Together, the two components produce reachability maps that better respect walls and traversable corridors.}
\label{fig:heatmap-comparison}
\end{figure}

\begin{wraptable}{r}{0.55\linewidth}
\centering
\captionof{table}{\textbf{Conditional reachability benchmark.} 
Ranking acceptance rates between feasible (positive) and infeasible (negative) pair scores: positives above the negative 95th percentile and negatives above the positive 95th percentile. We report mean $\pm$ std across 3 seeds.}
\label{tab:classifier-ablation}
\resizebox{\linewidth}{!}{%
\setlength{\tabcolsep}{3.5pt}
\begin{tabular}{ccc|cc}
    \toprule
    \textbf{Generation}&\textbf{Reweight} & $\boldsymbol k$ & \textbf{Feas.\ Accept.}$\mathbf{\ \uparrow}$ & \textbf{Infeas.\ Accept.}$\mathbf{\ \downarrow}$\\
    \midrule
    \ding{55} & \ding{55} & -- & $31 {\pm} 1$ & $0.49{\pm}0.14$ \\
    \ding{55} & \ding{51} & 50 & $40 {\pm} 2$& $0.07{\pm}0.02$  \\
    \ding{51} & \ding{55} & --  & $47{\pm}3$ & $0.05{\pm}0.05$ \\
    \midrule
    \ding{51} & \ding{51} & 10   & $73{\pm}1$ & $\mathbf{0.00{\pm}0.00}$ \\
    \ding{51} & \ding{51} & 50   & $\mathbf{74{\pm}0}$ & $\mathbf{0.00{\pm}0.00}$ \\
    \ding{51} & \ding{51} & 200  & $\mathbf{74{\pm}1}$ &  $\mathbf{0.00{\pm}0.00}$ \\
    \ding{51} & \ding{51} & 1000 & $\mathbf{74{\pm}0}$ & $\mathbf{0.00{\pm}0.00}$ \\
    \bottomrule
\end{tabular}
}
\vspace{-1em}
\end{wraptable}

\cref{tab:classifier-ablation} shows that using generated states and reweighting improve both acceptance of feasible pairs and rejection of infeasible pairs. Without our approach, the score is affected by the sampling density of the generated pool, which is concentrated in the interior of the maze. The classifier then becomes overconfident near the corner and underconfident near the center. \cref{fig:heatmap-comparison} shows this effect, and high-probability regions past walls are also observed without reweighting. Results were insensitive to $k$ over the range we tested.

\section{Conclusion}
We presented a Selective Compositional Diffuser (SCD), which selects composable chunks to form an executable trajectory. The method reformulates the composition problem of compositional diffusion planners as a chunk selection problem over generated chunks. We train a lightweight MLP network in a self-supervised manner to measure reachability. We use generated chunks jointly when training the reachability classifier to reduce the train-test mismatch, and propose the reweighting term for debiased estimation. After generating the candidate plans from diffusion models, we construct a graph based on the generated chunks and add edges weighted with the reachability scores. We find the minimum-cost chunk sequences by traversing the graph, generate bridging trajectories, and refine the full trajectory. We observe that this solution reduces incompatible mode compositions and mode-averaging issues. Across OGBench stitching tasks, SCD improves performance, reaching state-of-the-art success rates on HumanoidMaze-Large and Giant tasks.

\textbf{Limitations and future work.}
Our method depends on the coverage of candidate chunks and the accuracy of the reachability classifier. If the pool contains no start-to-goal route connectable within the bridge budget, selection cannot produce a feasible plan. The reachability classifier is trained with $k$-NN density reweighting, whose errors may introduce false-positive edges that graph search then selects, particularly in high-dimensional state spaces. Pairwise edge evaluation also scales quadratically with the number of candidate vertices, limiting the size of the pool. Future work could improve coverage by generating new chunks around low-score connections, improve the reachability classifier with learned density estimates, and reduce the edge evaluation cost with approximate nearest-neighbor search.

\subsection*{AI use statement}

In this work, we used generative AI tools to polish writing. We have not used generative AI tools for proving mathematical claims, generating synthetic data, drafting sections of the paper, research ideation or execution, or retrieval and discovery. We have reviewed all AI-assisted work. In particular, the authors reviewed and revised all AI-polished writing. We take responsibility for the final content of this work, including text, claims or artifacts produced with the aid of generative AI.

\subsection*{Ethics statement}

The contribution focuses on reformulating the mode-averaging problem as a chunk selection problem, with a modified reachability classification objective. The ethical and social implications are therefore aligned with those of existing generative diffusion planning research. These include concerns about the safety of executing generated plans on physical robots and the potential misuse of autonomous planning systems. All experiments use only publicly available datasets~\citep{park2025ogbench} and codebases~\citep{luo2025generative}, credited by citation and used under their licenses in \cref{app:licenses}. We collect no new data and conduct no human-subject research.

\subsection*{Reproducibility statement}

\cref{sec:method} describes SCD, and \cref{alg:graph} gives the inference procedure in pseudocode. \cref{app:implementation-details,app:experimental-details} provide the implementation details, hyperparameters, evaluation protocols, and computational resources. The sweeps used to select the hyperparameters are included in \cref{app:ablation-analysis}.

\bibliography{iclr2027_conference}
\bibliographystyle{iclr2027_conference}

\clearpage
\appendix

\section{Implementation Details}
\label[appendix]{app:implementation-details}

\subsection{Algorithm}
\label[appendix]{app:algorithm}

SCD constructs a reachability graph over the candidate chunks (Stages 1 and 2), finds a minimum-cost path (Stage 3), and converts the path into the final trajectory (Stages 4 and 5). \textsc{ParallelSample} samples $N$ plans of $K$ chunks in parallel with CompDiffuser, while $q_s$ and $q_g$ are inpainted. \textsc{Bridge} generates the $h_o$ intermediate states while conditioning on $M$ neighboring states of each adjacent chunk by inpainting. \textsc{Noise} noises a trajectory to the noise level $\kappa$. \textsc{Denoise} applies the last $j$-th DDIM step in a windowed manner and blends the window overlaps linearly while keeping $q_s$ and $q_g$ fixed. The procedure is repeated at every replan (\cref{app:main-evaluation}).

\algnewcommand{\Stage}[1]{\Statex\hspace*{-\leftmargin}\textbf{#1}}
\begin{algorithm}[th]
\caption{Selective Compositional Diffuser}
\label{alg:graph}
\begin{algorithmic}[1]
\Require The start and goal states $q_s, q_g$, the classifier $D_\phi$ with the key and query encoders $K_\theta$, $Q_\psi$, the number of plans $N$, the number of chunks in each plan $K$, the boundary stride $S$, the minimum segment length $M$, the discount factor $\gamma$, the maximum lag $L_\mathrm{max}$, the bridge budget $h_o$, the number of bridge candidates $N_\beta$, and the refinement level $\kappa$.
\Stage{Stage 1: Candidate sampling}
\State $\{\tau^{(n)}_{1:K}\}_{n=1}^{N}\gets\textsc{ParallelSample}(q_s,q_g;N,K)$ \Comment{Inpaint $s^{(n,1)}_1=q_s$, $s^{(n,K)}_H=q_g$}
\State $\tau_i\gets\tau^{(n)}_k=s^{(i)}_{1:H}$\quad $\forall i=(n,k)\in\mathcal{I}=[1, N]\times[1, K]$

\Stage{Stage 2: Graph construction}
\State $\mathcal{T}\gets\{1,1+S,1+2S,\ldots\}\cup\{H\}$ \Comment{Boundary stride $S$}
\State $\mathcal{V}\gets\{v_0,v_g\}\cup\{\mathrm{IN}(i,\ell) : i\in\mathcal{I},\ \ell\in\mathcal{T}\}\cup\{\mathrm{OUT}(i,r) : i \in\mathcal I,\ r\in\mathcal T\}$ \Comment{Vertices}
\State $c\gets\infty$ \Comment{Initialize all costs}
\State $c\bigl(v_0,\mathrm{IN}(i,1)\bigr)\gets0$\quad $\forall i=(n,1)$ \Comment{For the root node $v_0$}
\State $c\bigl(\mathrm{OUT}(i,H),v_g\bigr)\gets0$\quad $\forall i=(n,K)$ \Comment{For the goal node $v_g$}
\For{$i\in\mathcal{I}$,\ $\ell,r\in\mathcal{T}$,\ $r-\ell\ge M$} \Comment{For the segment edges}
    \State $\ell=t_0<t_1<\cdots<t_P=r$ \ s.t. \ $t_j-t_{j-1}\le L_\mathrm{max}$
    \State $c\bigl(\mathrm{IN}(i,\ell),\mathrm{OUT}(i,r)\bigr)\gets-\sum_{j=1}^{P}\log D_\phi\bigl(s^{(i)}_{t_{j-1}},s^{(i)}_{t_j},t_j-t_{j-1}\bigr)-(r-\ell)\log\gamma$
\EndFor
\State $\mathbf{Z}_{\rm q}\gets\bigl[Q_\psi\bigl(s^{(i)}_r,\mathrm{PE}(h_o)\bigr)\bigr]_{i\in\mathcal{I},\,r\in\mathcal{T}}$ \Comment{Cached queries}
\State $\mathbf{Z}_{\rm k}\gets\bigl[K_\theta(s^{(i)}_{\ell})/\|K_\theta(s^{(i)}_{\ell})\|_2\bigr]_{i\in\mathcal{I},\,\ell\in\mathcal{T}}$ \Comment{Cached keys}
\State $\mathbf{F}\gets\mathbf{Z}_{\rm q}\mathbf{Z}_{\rm k}^{\top}$ \Comment{Logits of all possible bridges}
\For{$i, i'\in\mathcal{I}$ {\rm \ s.t.} $i\ne i'$,\ $r,\ell'\in\mathcal{T}$} \Comment{Bridge edges}
    \State $c\bigl(\mathrm{OUT}(i,r),\mathrm{IN}(i',\ell')\bigr)\gets-\log\sigma\bigl(\mathbf{F}_{(i,r),(i',\ell')}\bigr)-h_o\log\gamma$
\EndFor
\State $\mathcal{G}\gets(\mathcal{V},c)$ \Comment{Reachability graph with non-negative edges}

\Stage{Stage 3: Path search}
\State $(v_0,v_1,\ldots,v_J)\gets\textsc{Dijkstra}(\mathcal{G};v_0,v_g)$ \Comment{Find the path of $\min\sum_{j=0}^{J-1}c(v_j,v_{j+1})$}
\State $A_t\gets s^{(i_t)}_{\ell_t:r_t}$ \ for \ $(v_{2t-1},v_{2t})=\bigl(\mathrm{IN}(i_t,\ell_t),\mathrm{OUT}(i_t,r_t)\bigr)$ \Comment{$T=(J-1)/2$}

\Stage{Stage 4: Bridge generation}
\For{$t=1,\ldots,T-1$}
    \State $\{\beta_t^{(b)}\}_{b=1}^{N_\beta}\gets\textsc{Bridge}\bigl(s^{(i_t)}_{r_t-M+1:r_t},\,s^{(i_{t+1})}_{\ell_{t+1}:\ell_{t+1}+M-1};\,h_o\bigr)$
    \State $\beta_t\gets\arg\min_{b}\mathcal{E}\bigl(\beta_t^{(b)}\bigr)$ \Comment{\cref{eq:bridge-mismatch}}
\EndFor
\State $\tilde\tau\gets(A_1,\beta_1,A_2,\ldots,\beta_{T-1},A_T)$

\Stage{Stage 5: Refinement}
\State $\tilde\tau\gets\textsc{Noise}(\tilde\tau,\kappa)$ 
\For{$j=\kappa,\ldots,1$}
    \State $\tilde\tau\gets\textsc{Denoise}(\tilde\tau,j;\,q_s,q_g)$ \Comment{Windowed refinement in \cref{sec:bridge-refinement}}
\EndFor
\State \Return $\tilde\tau$
\end{algorithmic}
\end{algorithm}

\subsection{Method Implementation}
\label[appendix]{app:method-implementation}

\textbf{Pretrained Components.} For each environment, we use the pretrained compositional diffusion models and controllers released by CompDiffuser~\citep{luo2025generative}. All diffusion-based methods in our experiments share these checkpoints (\cref{app:baseline-details}). The diffusion model is a 1D U-Net that denoises trajectories over $xy$ states with separate CNN encoders for boundary conditions. These conditions come from either the start and goal conditions or the overlapping parts of neighboring chunks. We sample it with DDIM using 50 steps. For execution, PointMaze uses a PD controller with PD gain $5.0$. AntMaze and HumanoidMaze use MLP inverse-dynamics models that map the full observation and the current $xy$ subgoal to an action. \cref{tab:pretrained-components} summarizes the architectures.

\begin{table}[t]
\centering
\caption{\textbf{Hyperparameters of the released CompDiffuser.}}
\label{tab:pretrained-components}
\resizebox{0.9\linewidth}{!}{%
\begin{tabular}{llll}
\toprule
Component & PointMaze & AntMaze & HumanoidMaze \\
\midrule
\multicolumn{4}{l}{\textit{Diffusion model}} \\
\midrule
Planner dim. & 2D & 2D & 2D \\
U-Net base dim. & 128 & 128 & 128 \\
U-Net dim.\ multipliers & $(1,2,4,8)$ & $(1,2,4,8)$ & $(1,2,4,8)$ \\
Boundary encoder & CNN & CNN & CNN \\
Boundary encoding dim. & 256 & 256 & 256 \\
\midrule
\multicolumn{4}{l}{\textit{Controller}} \\
\midrule
Type & PD controller & MLP inverse dynamics & MLP inverse dynamics \\
PD gain & 5.0 & -- & -- \\
MLP hidden sizes & -- & $(512,512,512)$ & $(512,1024,1024,512,256)$ \\
\bottomrule
\end{tabular}
}
\end{table}

\begin{figure}[h]
\centering
\begin{minipage}{0.025\linewidth}\centering\rotatebox{90}{\small{AntMaze}}\end{minipage}
\begin{minipage}{0.24\linewidth}\centering\includegraphics[width=\linewidth]{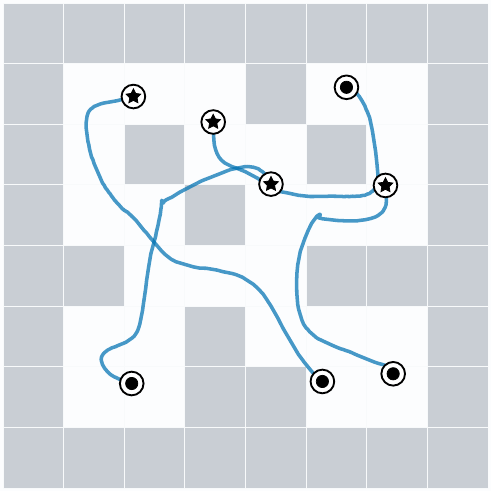}\end{minipage}
\begin{minipage}{0.32\linewidth}\centering\includegraphics[width=\linewidth]{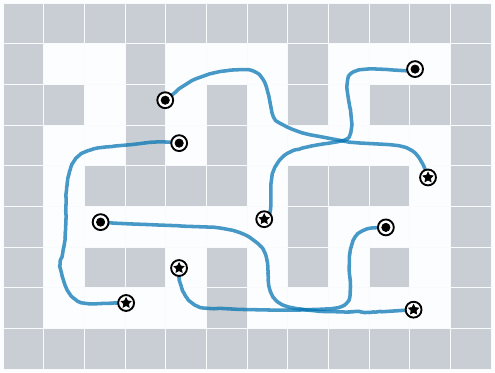}\end{minipage}
\begin{minipage}{0.32\linewidth}\centering\includegraphics[width=\linewidth]{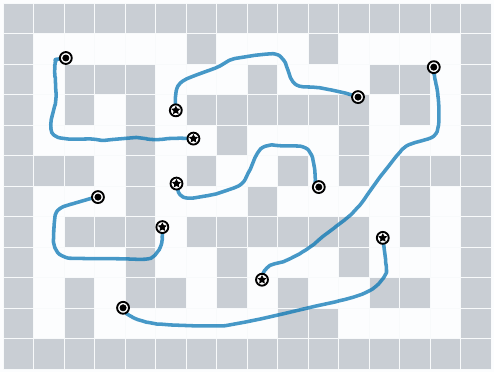}\end{minipage}

\begin{minipage}{0.025\linewidth}\centering\rotatebox{90}{\small{PointMaze}}\end{minipage}
\begin{minipage}{0.24\linewidth}\centering\includegraphics[width=\linewidth]{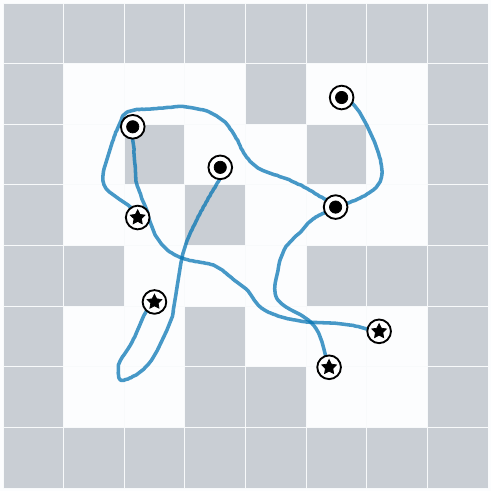}\end{minipage}
\begin{minipage}{0.32\linewidth}\centering\includegraphics[width=\linewidth]{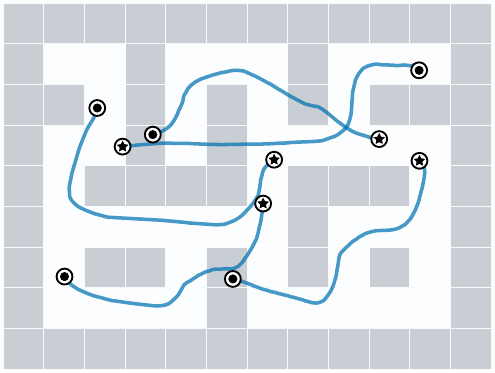}\end{minipage}
\begin{minipage}{0.32\linewidth}\centering\includegraphics[width=\linewidth]{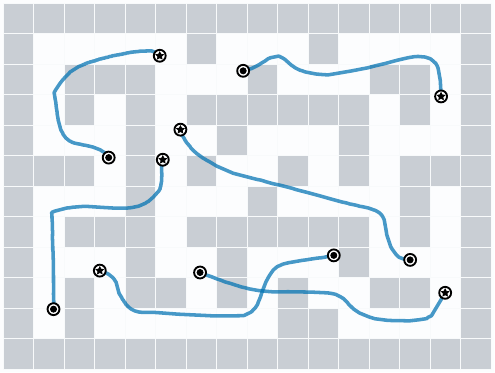}\end{minipage}

\begin{minipage}{0.025\linewidth}\centering\rotatebox{90}{\small{HumanoidMaze}}\end{minipage}
\begin{minipage}{0.24\linewidth}\centering\includegraphics[width=\linewidth]{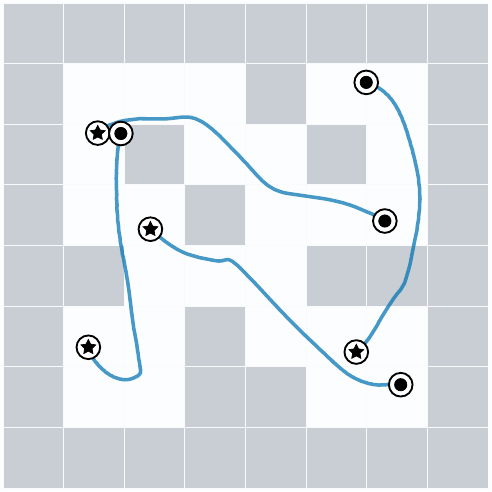}\end{minipage}
\begin{minipage}{0.32\linewidth}\centering\includegraphics[width=\linewidth]{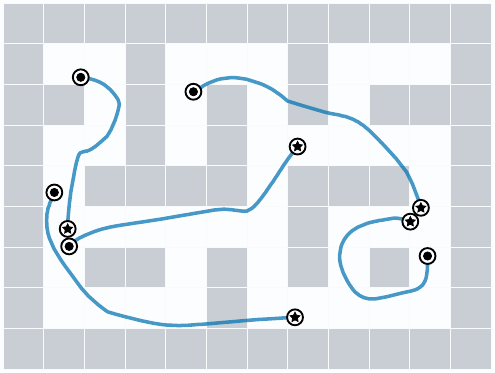}\end{minipage}
\begin{minipage}{0.32\linewidth}\centering\includegraphics[width=\linewidth]{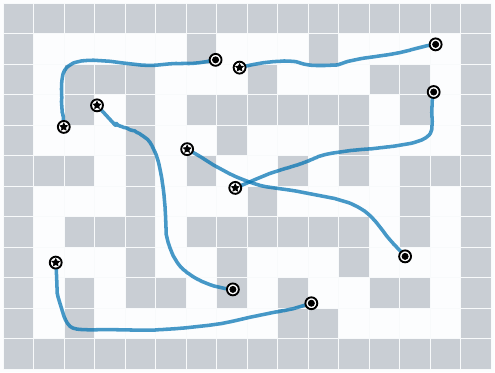}\end{minipage}

\begin{minipage}{0.025\linewidth}\ \end{minipage}
\begin{minipage}{0.24\linewidth}\centering\small{Medium}\end{minipage}
\begin{minipage}{0.32\linewidth}\centering\small{Large}\end{minipage}
\begin{minipage}{0.32\linewidth}\centering\small{Giant}\end{minipage}

\caption{\textbf{Samples of generated trajectory data.}}
\label{fig:generated-data-samples}
\end{figure}

\textbf{Generated State Bank.} For classifier training, we build a generated state bank with states from generated chunks. We use the same pretrained diffusion model for this state bank. Following RCD~\citep{lee2026refiningcompositionaldiffusionreliable}, we compute the self-reconstruction error of each generated sample to keep low-density samples over the offline distribution. We noise each generated sample to a probe level, obtain two independent noised samples, denoise each once, compute the squared error between the original sample and the reconstructions, and use the average of the two error values. We keep generated samples when this self-reconstruction error exceeds the 90th percentile of the errors, computed over the offline validation data. For the probe level, we use 35\% of the denoising steps of the training setting, $179 = \lfloor 512 \times 0.35\rfloor$ for AntMaze and PointMaze, $350 = \lfloor 1000 \times 0.35 \rfloor$ for HumanoidMaze, following the RCD setting. We note that this filtering aims to make chunks more distinguishable over offline and generated chunks for better classifier training, and we do not use this strategy for SCD at inference time. See \cref{fig:generated-data-samples} for samples of generated trajectories.

\begin{table}[th]
\centering
\caption{\textbf{Training settings} for the reachability classifier.}
\label{tab:reachability-defaults}
\resizebox{0.9\linewidth}{!}{%
\begin{tabular}{ll}
\toprule
Setting & Value \\
\midrule
Query encoder $Q_\psi$ & 3-layer MLP, width 512, SiLU \\
Key encoder $K_\theta$ & 3-layer MLP, width 512, SiLU \\
Lag encoding & Sinusoidal positional encoding, 64-dim. \\
Maximum lag $L_{\max}$ & 56 by default / 128 for HumanoidMaze \\
Lag sampling & Uniform $\mathcal U[1, L_{\max})$ for 75\%, $L_{\max}$ for 25\% \\
$k$-NN reweighting & $k=50$, $w_\mathrm{min} = 10^{-8}$, $w_\mathrm{max} = 20$ \\
\midrule
Optimizer & AdamW~\citep{adamw} \\
Learning rate $\eta$ & $3\times 10^{-4}$ \\
Momentum parameters ($\beta_1$, $\beta_2$) & (0.9, 0.999) \\
Weight decay $\lambda$ & 0.01 \\
Epsilon $\epsilon$ & $10^{-8}$ \\
Learning rate scheduler & Cosine decay without warmup ($T_\mathrm{max} = 100\mathrm{K}$) \\
Batch size & 8192 for the positive batch, 8192 for the additional negative batch \\
Training steps & 100K \\
Data split (train / valid / test) & 80 / 10 / 10 \\
\bottomrule
\end{tabular}
}
\end{table}

\textbf{Reachability Classifier.} We train a reachability classifier for each environment. The settings follow \cref{tab:reachability-defaults}, which are shared across environments except $L_{\max}$. The query encoder $Q_\psi$ takes $x$ with a sinusoidal encoding of $L$, and the key encoder $K_\theta$ takes $y$. To build an observed pair, we first sample a lag $L\in [1, L_{\max}]$, where the bridge budget $h_o = L_{\max}$ in \cref{sec:graph construction}. We then pick the target $y_\mathrm{obs}$ uniformly from the next $L$ states of $x_\mathrm{obs}$ in the same offline trajectory. The two states of a negative pair are sampled independently from the bank, and the weights use the $k$-NN distances.

For $k$-NN estimation, we normalize $xy$ into the range $[-1, 1]$ with min-max normalization. We use FAISS's \texttt{IndexIVFFlat} with \texttt{IndexFlatL2} for efficient approximate nearest-neighbor search.

\begin{table}[th]
\centering
\caption{\textbf{Default settings} for candidate generation, graph search, and trajectory refinement. We present environmental settings such as the chunk length $H$ and the bridge budget $h_o$ in \cref{tab:eval-settings}.}
\label{tab:scd-defaults}
{\small
\begin{tabular}{ll}
\toprule
Setting & Value \\
\midrule
Baseline & CompDiffuser, sample chunks in parallel (without interleaving) \\
Scheduler & 50-step DDIM~\citep{song2021denoising} \\
Minimum segment length $M$ & 48 by default / 80 for HumanoidMaze (i.e., $M=H-2h_o$) \\
Boundary stride $S$& 24 by default / 40 for HumanoidMaze \\
Discount factor $\gamma$ & 0.999 \\
Shortest-path solver & Dijkstra's algorithm (SciPy, CPU) \\
Bridge candidates $N_\beta$ & 5 \\
Refinement noise level $\kappa$ & Last five DDIM steps \\
\bottomrule
\end{tabular}
}
\end{table}

\textbf{Planning.} We sample $N$ candidate plans of $K$ chunks with the parallel sampler of CompDiffuser, which denoises all chunks of each plan together. The first state of the first chunk and the last state of the last chunk are inpainted with $q_s$ and $q_g$. We set candidate boundaries at a fixed stride of $S$, such as $\ell \in \{1,25, 49, 73, \ldots\}$ (see \cref{fig:graph-construction} in \cref{sec:graph construction}). If a segment is longer than $L_\mathrm{max}$, we partition the segment into subparts, equal to or shorter than $L_\mathrm{max}$, and accumulate costs of each subpart. We set a root vertex as the starting point $q_s$, connecting to each first chunk with zero cost, and run Dijkstra's algorithm to find the minimum-cost path from the root vertex to the goal state vertex.

We generate each bridge with DDIM~\citep{song2021denoising} for 50 sampling steps. We inpaint $M$ states on each side with the neighboring states from the adjacent chunks, and denoise the $h_o$ bridging states between them. If there are remaining states ($H > h_o + 2M$), we leave both ends untouched until denoising:
\begin{align*}
\underbrace{
\big|\ \text{untouched}\ 
\big|\ M\ \text{inpainted}\ 
\big|\ h_o\ \text{mainly denoised}\ 
\big|\ M\ \text{inpainted}\ 
\big|\ \text{untouched}\ 
\big|
}_H.
\end{align*}
We draw $N_\beta$ candidate samples per bridge and measure the mismatch of each bridge candidate $\beta$ as
\begin{align}\label{eq:bridge-mismatch}
\mathcal{E}(\beta) = \sum_{i=0}^{M-1} \frac{i}{M-1}\,\bigl\|\hat{s}^{L}_i - s^{L}_i\bigr\|^2 + \sum_{i=0}^{M-1} \frac{M-1-i}{M-1}\,\bigl\|\hat{s}^{R}_i - s^{R}_i\bigr\|^2,
\end{align}
where $s^{L}_i$ and $s^{R}_i$ are the $i$-th inpainted states, and $\hat{s}^{L}_i$ and $\hat{s}^{R}_i$ are their clean predictions at the last DDIM step. We select the candidate with the smallest $\mathcal{E}(\beta)$. Finally, we noise the assembled trajectory to the last five DDIM steps and denoise it with the same windows and overlaps as the initial plan generation. We blend the overlaps linearly while keeping $q_s$ and $q_g$ fixed, as introduced in \cref{eq:interpolation}. This planner outputs a sequence of waypoints, one per environment step, which the controllers execute as described in \cref{app:main-evaluation}. Detailed parameters follow \cref{tab:scd-defaults}.

\section{Experimental Details}
\label[appendix]{app:experimental-details}

\subsection{Environment Details}
\begin{figure}[ht]
\centering
\begin{minipage}{0.025\linewidth}\centering\rotatebox{90}{\small{AntMaze}}\end{minipage}
\begin{minipage}{0.24\linewidth}\centering\includegraphics[width=\linewidth]{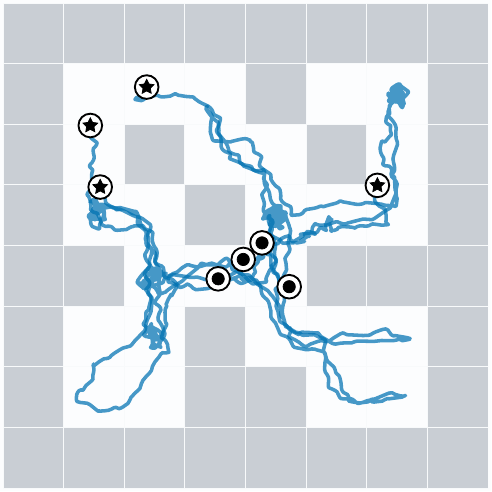}\end{minipage}
\begin{minipage}{0.32\linewidth}\centering\includegraphics[width=\linewidth]{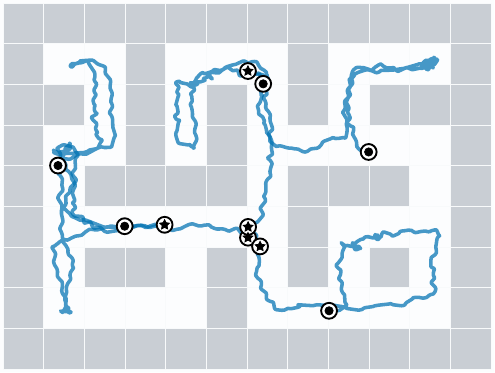}\end{minipage}
\begin{minipage}{0.32\linewidth}\centering\includegraphics[width=\linewidth]{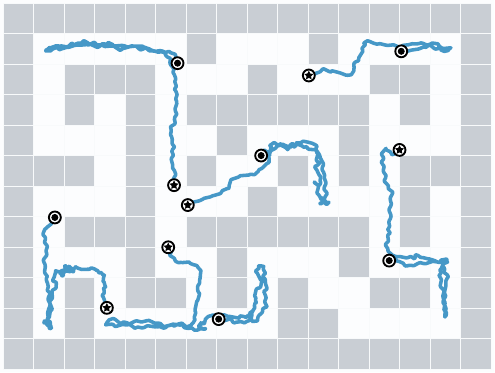}\end{minipage}

\begin{minipage}{0.025\linewidth}\centering\rotatebox{90}{\small{PointMaze}}\end{minipage}
\begin{minipage}{0.24\linewidth}\centering\includegraphics[width=\linewidth]{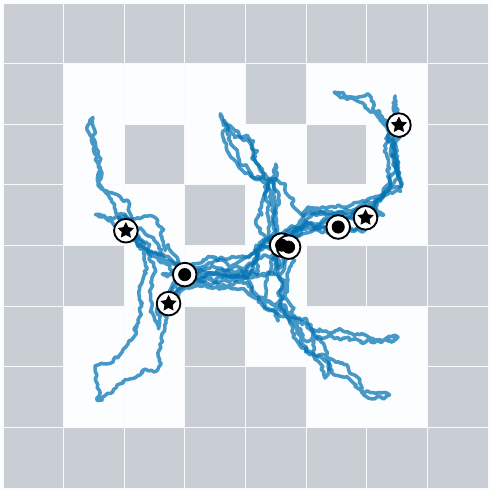}\end{minipage}
\begin{minipage}{0.32\linewidth}\centering\includegraphics[width=\linewidth]{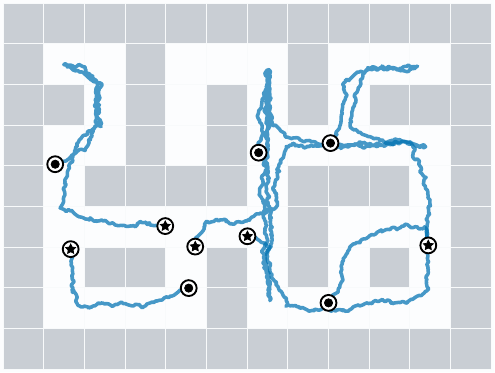}\end{minipage}
\begin{minipage}{0.32\linewidth}\centering\includegraphics[width=\linewidth]{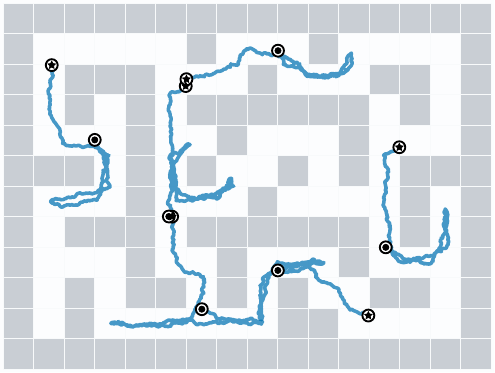}\end{minipage}

\begin{minipage}{0.025\linewidth}\centering\rotatebox{90}{\small{HumanoidMaze}}\end{minipage}
\begin{minipage}{0.24\linewidth}\centering\includegraphics[width=\linewidth]{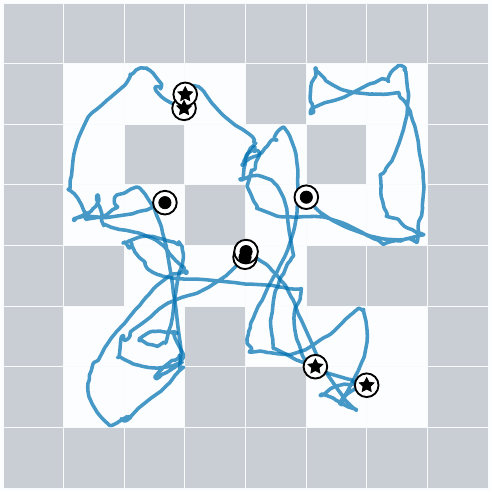}\end{minipage}
\begin{minipage}{0.32\linewidth}\centering\includegraphics[width=\linewidth]{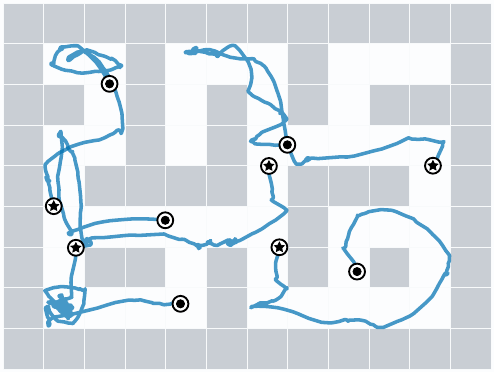}\end{minipage}
\begin{minipage}{0.32\linewidth}\centering\includegraphics[width=\linewidth]{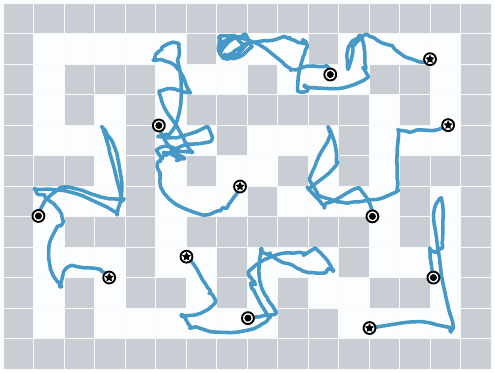}\end{minipage}

\begin{minipage}{0.025\linewidth}\ \end{minipage}
\begin{minipage}{0.24\linewidth}\centering\small{Medium}\end{minipage}
\begin{minipage}{0.32\linewidth}\centering\small{Large}\end{minipage}
\begin{minipage}{0.32\linewidth}\centering\small{Giant}\end{minipage}

\caption{\textbf{Samples of offline trajectory data}, with each source (circle) connected to a target (star).}
\label{fig:offline-data-samples}
\end{figure}

\label[appendix]{app:environment-details}
We evaluate on the OGBench Stitch datasets for PointMaze, AntMaze, and HumanoidMaze, each at Medium, Large, and Giant maze sizes. Each trajectory covers only local transitions of the maze, whereas evaluation requires reaching start--goal pairs across longer distances. The planner must compose behaviors beyond the horizon of individual training trajectories. The dataset identifiers follow the pattern \texttt{<env>-<size>-stitch-v0} (e.g., \texttt{antmaze-giant-stitch-v0}), and all planners operate in the 2D $xy$ space. \Cref{tab:eval-settings} lists the episode limit of each environment. See \cref{fig:offline-data-samples} for samples of offline stitch data.

\subsection{Baselines}
\label[appendix]{app:baseline-details}

The baseline scores in \cref{tab:ogbench-stitch} are drawn from published results. Scores for GCBC, GCIVL, GCIQL, QRL, CRL, HIQL, GSC, and CD are taken from CompDiffuser~\citep{luo2025generative}, and scores for CDGS, ECD, and RCD are taken from their respective papers~\citep{ICLR2026_6fbe71b5,sun2026energybased,lee2026refiningcompositionaldiffusionreliable}. For the feasibility and planning cost analyses in \cref{tab:feasibility-ratio,tab:time-comparison}, we run the publicly released implementations of CDGS, ECD, and RCD with their default settings and the same pretrained local diffusion model, allowing us to compare their plans and runtimes under a shared model.

In \cref{tab:time-comparison}, \textit{CD (parallel)} denoises all chunks simultaneously, whereas \textit{CD (interleave)} sequentially denoises chunks, allowing later chunks to condition on previously updated neighbors.

\subsection{Evaluation Details}
\label[appendix]{app:evaluation-details}

All experiments, including the planning-time measurements in \cref{tab:time-comparison}, are conducted on a single NVIDIA RTX 4090 GPU. SCD uses the pretrained diffusion model of CompDiffuser~\citep{luo2025generative} without further training, and only the reachability classifier is trained for our method.

\subsubsection{Feasibility Evaluation}
\label[appendix]{app:feasibility-evaluation}

We measure the geometric feasibility of planned paths. For each of the five HumanoidMaze-Giant-Stitch tasks, we generate 20 plans per method, yielding 100 plans per method. \cref{tab:feasibility-ratio} reports the percentage of feasible plans for each task and the mean across tasks.

\begin{wrapfigure}{r}{0.5\linewidth}
\vspace{-1em}
\centering
\includegraphics[width=\linewidth]{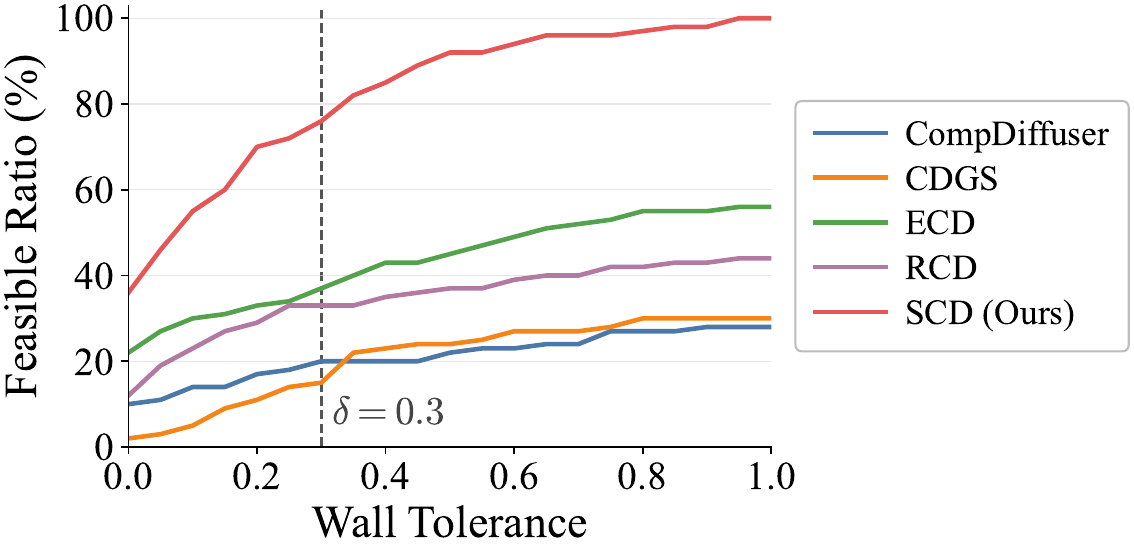}
\caption{\textbf{Feasible ratio over wall tolerance.} Feasible ratio of the plans in \cref{tab:feasibility-ratio} as a function of the tolerance $\delta$. The dashed line marks $\delta=0.3$ used in \cref{tab:feasibility-ratio}.}
\label{fig:feasibility-tolerance}
\vspace{-1em}
\end{wrapfigure}

We evaluate geometric feasibility by checking whether any line segment between consecutive planned positions crosses a maze wall. Each wall is represented by a $4\times4$ square, contracted inward by $\delta$ on each side to tolerate small boundary incursions. We use $\delta=0.3$ for \cref{tab:feasibility-ratio}. To detect paths passing between adjacent walls, we also check whether they intersect the line segments connecting the walls' centers. We count a plan as feasible only if no segment between consecutive $xy$ positions intersects a contracted wall or one of these lines. \cref{fig:feasibility-tolerance} reports the feasible-plan ratio as $\delta$ varies; SCD remains highest throughout the tested range.

\subsubsection{Main Evaluation}
\label[appendix]{app:main-evaluation}

\begin{table}[t]
\centering
\caption{\textbf{Planning and evaluation settings per environment.} \textit{Max steps} denotes the episode limit. $H$ denotes the chunk horizon, $h_o$ the overlap length of the pretrained planner, $K$ the number of chunks per candidate plan, and $N$ the number of candidate plans. \textit{Replan thr.} and \textit{Max replans} denote the tracking threshold and the maximum number of replans. All values follow the released configuration of CompDiffuser~\citep{luo2025generative}.}
\label{tab:eval-settings}
\resizebox{0.95\linewidth}{!}{%
\begin{tabular}{llccccccc}
\toprule
\textbf{Environment} & \textbf{Size} & \textbf{Max steps} & $\boldsymbol{H}$ & $\boldsymbol{h_o}$ & $\boldsymbol{K}$ & $\boldsymbol{N}$ & \textbf{Replan thr.} & \textbf{Max replans} \\
\midrule
\multirow{3}{*}{PointMaze}
    & Medium & 1000 & 160 & 56 & 3 & 40 & 1 & 0 \\
    & Large  & 1000 & 160 & 56 & 6 & 40 & 1 & 0 \\
    & Giant  & 1000 & 160 & 56 & 8 & 40 & 1 & 10 \\
\midrule
\multirow{3}{*}{AntMaze}
    & Medium & 1000 & 160 & 56 & 3 & 40 & 4 & 10 \\
    & Large  & 2000 & 160 & 56 & 6 & 40 & 4 & 10 \\
    & Giant  & 2000 & 160 & 56 & 9 & 40 & 4 & 10 \\
\midrule
\multirow{3}{*}{HumanoidMaze}
    & Medium & 5000 & 336 & 128 & 4 & 40 & 10 & 10 \\
    & Large  & 5000 & 336 & 128 & 6 & 40 & 10 & 10 \\
    & Giant  & 8000 & 336 & 128 & 11 & 40 & 10 & 10 \\
\bottomrule
\end{tabular}
}
\end{table}

We evaluate task success on nine OGBench Stitch environments in \cref{tab:eval-settings}. Each environment provides five evaluation tasks. Each task consists of 20 fixed start--goal pairs, and we run one episode per pair with each of three random seeds, yielding 300 rollouts per environment in total. The seeds control only the stochastic sampling during planning, and the start--goal pairs are shared across seeds. An episode is successful if the environment reports that the $xy$ distance between the agent and the goal is at most $\epsilon$ within the episode limit. OGBench sets $\epsilon=1.0$ for PointMaze and $\epsilon=0.5$ for AntMaze and HumanoidMaze. We average the success rates of the five tasks within each seed and report the mean and the standard deviation across seeds in \cref{tab:ogbench-stitch}. Baseline results in that table are borrowed from the sources listed in \cref{app:baseline-details}.

We execute planned trajectories with the controllers in \cref{app:method-implementation}, which are shared by all diffusion-based methods. At each step, the controller takes the waypoint of the plan at the current time step as a subgoal and predicts an action from the current state and the subgoal. Following CompDiffuser, we replan from the current state to the original goal when the Euclidean distance to the subgoal exceeds a threshold, which differs by environment, or when the agent runs out of remaining subgoals. We also follow its schedule that reduces the number of chunks $K$ at each replan. \cref{tab:eval-settings} lists the threshold and the maximum number of replans for each environment, together with other planner settings. Once the replanning budget is exhausted, the agent tracks the last waypoint until the episode limit. For PointMaze-Giant, increasing the replanning threshold from $1$ to $4$ raises SCD's success rate from $91{\pm}1$ to $99{\pm}1$. However, we retain the threshold of $1$ in \cref{tab:ogbench-stitch} for fair comparison.

\subsubsection{Reachability Evaluation}
\begin{figure}[t]
\centering
\includegraphics[width=\linewidth]{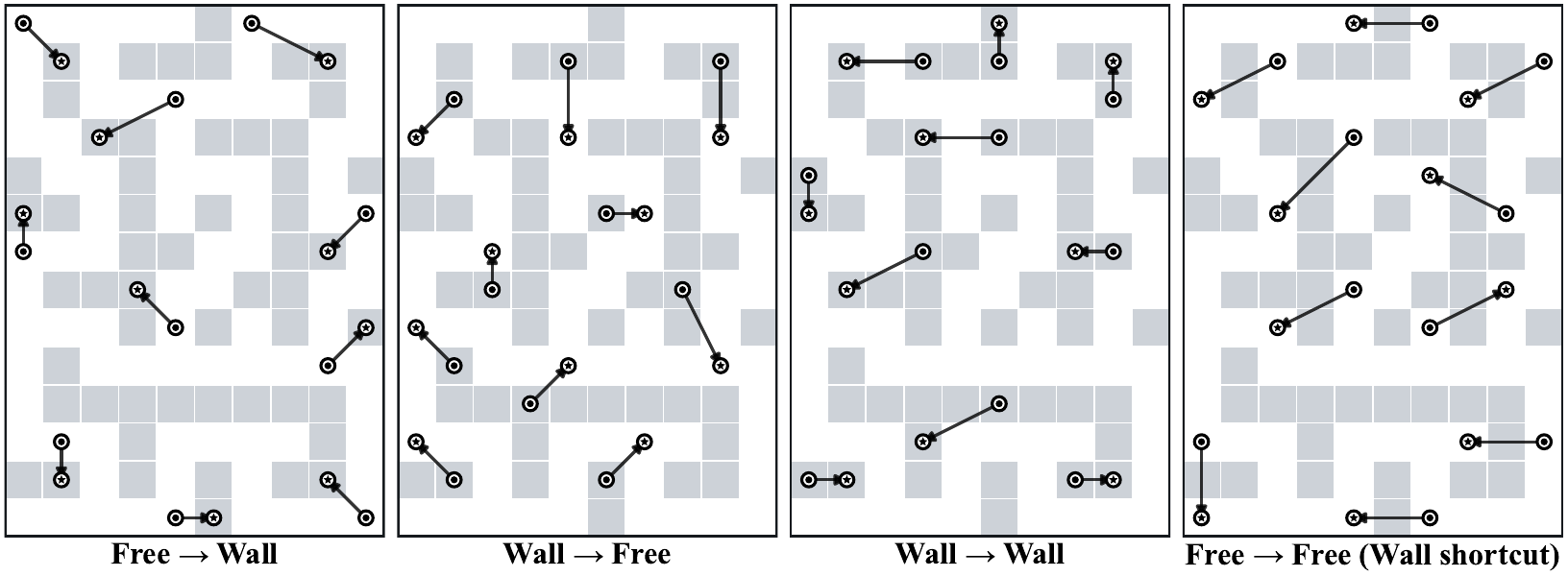}
\caption{\textbf{Examples of the four infeasible pair types} in the reachability test set (AntMaze-Giant). Each arrow points from a source (circle) to a target (star), and gray cells denote walls.}
\label{fig:negative-pair-types}
\end{figure}

\label[appendix]{app:reachability-evaluation}
We evaluate the reachability classifier on a fixed test set in AntMaze-Giant with lags $L\le 56$. The test set consists of 400 feasible and 400 infeasible base start-to-goal state pairs on the maze grid. Each feasible base pair takes the cells of an offline state pair $(s_t, s_{t+L})$ that is not used for training. The infeasible pairs consist of four types with 100 pairs each: free-to-wall, wall-to-free, wall-to-wall, and free-to-free pairs. In particular, free-to-free pairs are close in Euclidean distance but far in geodesic distance. For each base pair, we sample the source and target 100 times uniformly, yielding 40,000 feasible and 40,000 infeasible pairs. \Cref{fig:negative-pair-types} shows examples of the four infeasible types.

We compute the classifier logit $f_\phi(x,y,L)$ for all pairs and set two thresholds: $t_{\mathrm{neg}}$, the 95th percentile of the infeasible scores, and $t_{\mathrm{pos}}$, the 95th percentile of the feasible scores. \textit{Feas. Accept.} in \cref{tab:classifier-ablation} is the percentage of feasible pairs scoring above $t_{\mathrm{neg}}$, and \textit{Infeas. Accept.} is the percentage of infeasible pairs scoring above $t_{\mathrm{pos}}$.

\textit{Generation} \ding{55} samples the negative endpoints only from the offline dataset states, whereas \ding{51} samples them from both the offline dataset states and the generated states. \textit{Reweight} \ding{55} sets all reweighting terms to $w(x)=1$, and \ding{51} uses the clipped inverse-density weights in \cref{sec:discriminator} estimated with $k$-NN distances. All variants share the same architecture and training setup in \cref{app:implementation-details}. We train each variant with three random seeds and report the mean and the standard deviation across seeds. For \cref{fig:heatmap-comparison}, we choose five start states $s_1,\ldots,s_5$ outside the walls that are far enough apart that no state is reachable from more than one of them. We then compute $D_\phi(s_i,y,L)$ with $L=56$ for every state $y$ in the maze and show $\max_i D_\phi(s_i,y,L)$.

\section{Ablation Analysis}
\label[appendix]{app:ablation-analysis}

We conduct an ablation study of hyperparameters on AntMaze-Giant in \cref{tab:planning-ablation}. We follow the evaluation protocols in \cref{app:main-evaluation} and use default settings for all other hyperparameters.

\begin{table}[ht]
\centering
\caption{\textbf{Ablation of hyperparameters on AntMaze-Giant.} Each subtable varies one hyperparameter. We report success rates (\%) as mean $\pm$ std across three seeds.}
\label{tab:planning-ablation}
{
\subfloat[Bridge samples]{%
\begin{tabular}{c|c}
    \toprule
    $\boldsymbol{N_\beta}$ & \textbf{Succ.} \\
    \midrule
    1 & $83.0{\pm}2.2$ \\
    3 & $84.0{\pm}2.2$ \\
    \cellcolor[HTML]{efefef}5 & \cellcolor[HTML]{efefef}$\mathbf{85.3{\pm}1.7}$ \\
    \bottomrule
\end{tabular}}
\qquad
\subfloat[Boundary stride]{%
\begin{tabular}{c|c}
    \toprule
    $\boldsymbol{S}$ & \textbf{Succ.} \\
    \midrule
    12 & $82.0{\pm}3.6$ \\
    \cellcolor[HTML]{efefef}24 & \cellcolor[HTML]{efefef}$\mathbf{85.3{\pm}1.7}$ \\
    Endpoints & $85.0{\pm}2.8$ \\
    \bottomrule
\end{tabular}}
\qquad
\subfloat[Refinement steps]{%
\begin{tabular}{c|c}
    \toprule
    $\boldsymbol{\kappa}$ & \textbf{Succ.} \\
    \midrule
    0 & $84.3{\pm}2.1$ \\
    \cellcolor[HTML]{efefef}5 & \cellcolor[HTML]{efefef}$\mathbf{85.3{\pm}1.7}$ \\
    10 & $81.3{\pm}1.3$ \\
    \bottomrule
\end{tabular}}

\subfloat[Candidates]{%
\begin{tabular}{c|c}
    \toprule
    $\boldsymbol{N}$ & \textbf{Succ.} \\
    \midrule
    10 & $79.3{\pm}5.6$ \\
    20 & $81.7{\pm}5.0$ \\
    \cellcolor[HTML]{efefef}40 & \cellcolor[HTML]{efefef}$\mathbf{85.3{\pm}1.7}$ \\
    80 & $84.7{\pm}0.5$ \\
    \bottomrule
\end{tabular}}
\qquad
\subfloat[Discount factor]{%
\begin{tabular}{c|c}
    \toprule
    $\boldsymbol{\gamma}$ & \textbf{Succ.} \\
    \midrule
    0.99 & $69.0{\pm}2.9$ \\
    \cellcolor[HTML]{efefef}0.999 & \cellcolor[HTML]{efefef}$\mathbf{85.3{\pm}1.7}$ \\
    0.9999 & $83.7{\pm}1.3$ \\
    1 & $75.7{\pm}1.3$ \\
    \bottomrule
\end{tabular}}
}
\end{table}

Changing the number of bridge samples, the boundary stride, the refinement steps, or the number of candidate plans ($N\ge20$) shifts the success rate by at most 4\%. $N=10$ produces a larger drop, consistent with the coverage limitation. The discount factor $\gamma$ shows a stronger effect as it trades the reachability for the plan length. A $\Delta$-step-shorter plan is preferred when its reachability score is larger than $\gamma^{\Delta}$ times the original score. This means that when $\gamma = 0.999$, a 200-step-shorter trajectory would be selected when its reachability score is at least 82\% ($\approx 0.999^{200}$) of that of the longer plan. This threshold falls to $0.99^{200}\approx13\%$ at $\gamma=0.99$, which may disrupt the reachability assessment. At $\gamma=1$, there is no length penalty to discourage unnecessarily long paths, which can hit the time limit of each task.

\section{Qualitative Results}
\label[appendix]{app:qualitative-results}

\begin{figure}[h!]
    \centering
    \includegraphics[width=0.65\linewidth]{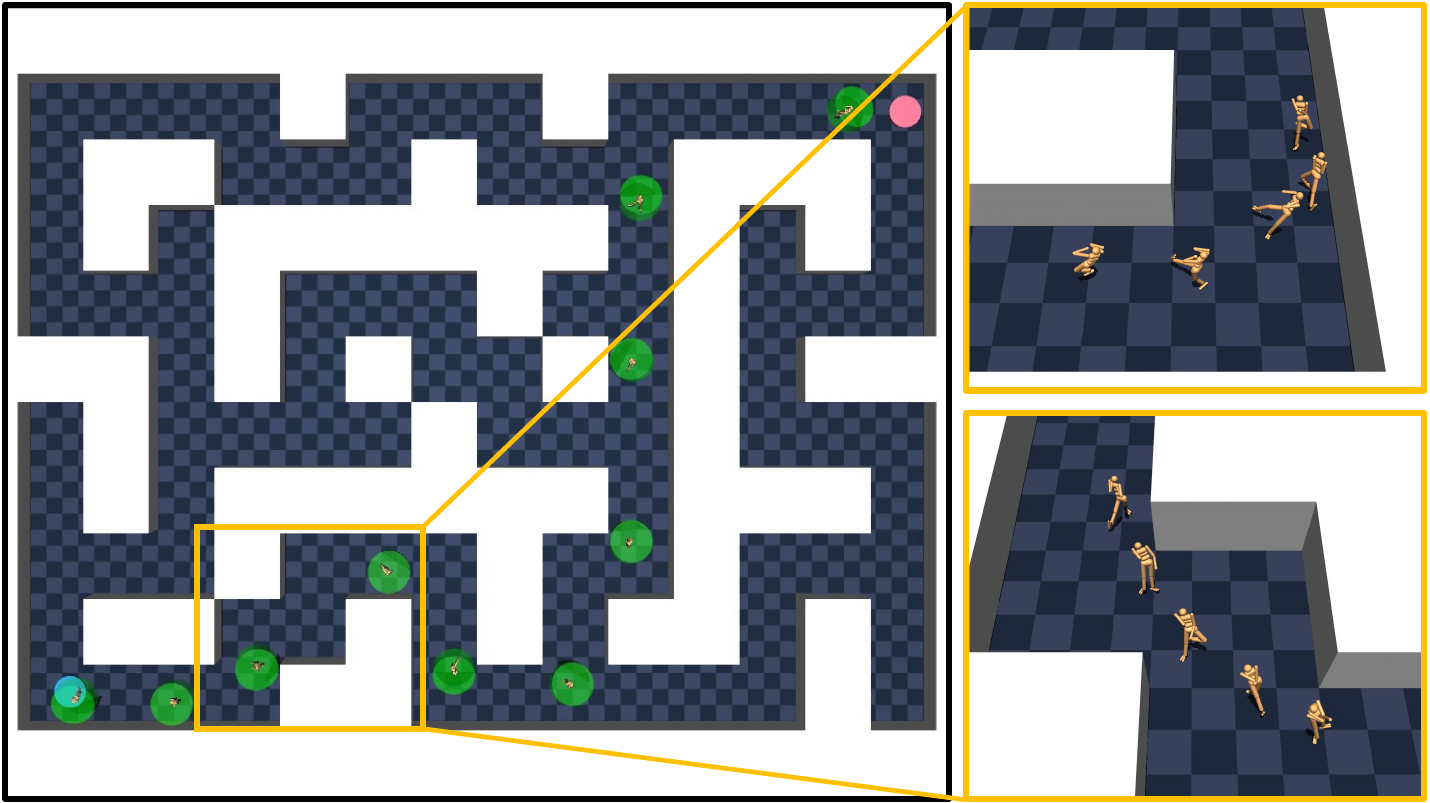}
    \caption{\textbf{Top-down plan overview.} SCD rollout in HumanoidMaze-Giant, where green circles mark the humanoid at intervals along its path from the start state (cyan) to the goal state (pink). The insets magnify the boxed region, showing the humanoid following through the plan.}
    \label{fig:topdown-render}
\end{figure}

\begin{figure}[p]
\centering
\begin{minipage}{0.025\linewidth}\centering\rotatebox{90}{\small{CompDiffuser}}\end{minipage}
\begin{minipage}{0.18\linewidth}\centering\includegraphics[width=\linewidth]{fig/C_failures/C_CD_1.pdf}\end{minipage}
\begin{minipage}{0.18\linewidth}\centering\includegraphics[width=\linewidth]{fig/C_failures/C_CD_2.pdf}\end{minipage}
\begin{minipage}{0.18\linewidth}\centering\includegraphics[width=\linewidth]{fig/C_failures/C_CD_3.pdf}\end{minipage}
\begin{minipage}{0.18\linewidth}\centering\includegraphics[width=\linewidth]{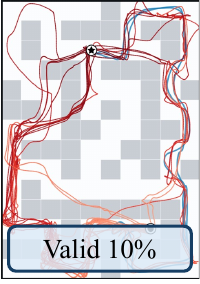}\end{minipage}
\begin{minipage}{0.18\linewidth}\centering\includegraphics[width=\linewidth]{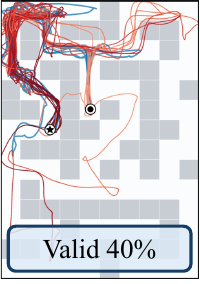}\end{minipage}

\begin{minipage}{0.025\linewidth}\centering\rotatebox{90}{\small{CDGS}}\end{minipage}
\begin{minipage}{0.18\linewidth}\centering\includegraphics[width=\linewidth]{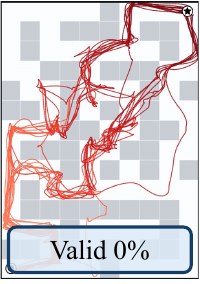}\end{minipage}
\begin{minipage}{0.18\linewidth}\centering\includegraphics[width=\linewidth]{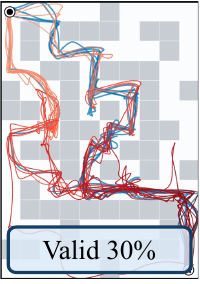}\end{minipage}
\begin{minipage}{0.18\linewidth}\centering\includegraphics[width=\linewidth]{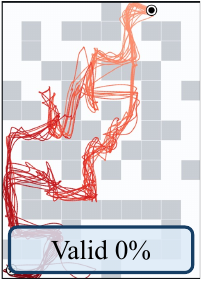}\end{minipage}
\begin{minipage}{0.18\linewidth}\centering\includegraphics[width=\linewidth]{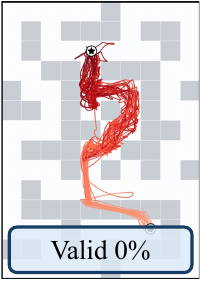}\end{minipage}
\begin{minipage}{0.18\linewidth}\centering\includegraphics[width=\linewidth]{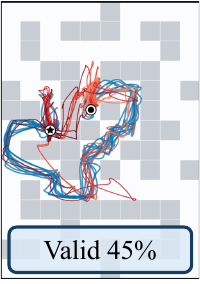}\end{minipage}

\begin{minipage}{0.025\linewidth}\centering\rotatebox{90}{\small{ECD}}\end{minipage}
\begin{minipage}{0.18\linewidth}\centering\includegraphics[width=\linewidth]{fig/C_failures/C_ECD_1.pdf}\end{minipage}
\begin{minipage}{0.18\linewidth}\centering\includegraphics[width=\linewidth]{fig/C_failures/C_ECD_2.pdf}\end{minipage}
\begin{minipage}{0.18\linewidth}\centering\includegraphics[width=\linewidth]{fig/C_failures/C_ECD_3.pdf}\end{minipage}
\begin{minipage}{0.18\linewidth}\centering\includegraphics[width=\linewidth]{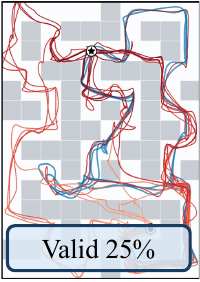}\end{minipage}
\begin{minipage}{0.18\linewidth}\centering\includegraphics[width=\linewidth]{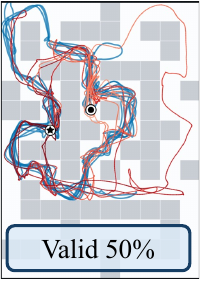}\end{minipage}

\begin{minipage}{0.025\linewidth}\centering\rotatebox{90}{\small{RCD}}\end{minipage}
\begin{minipage}{0.18\linewidth}\centering\includegraphics[width=\linewidth]{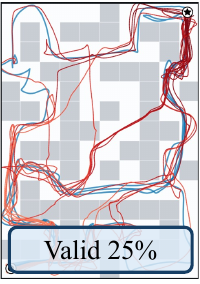}\end{minipage}
\begin{minipage}{0.18\linewidth}\centering\includegraphics[width=\linewidth]{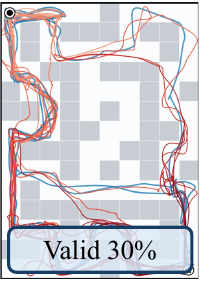}\end{minipage}
\begin{minipage}{0.18\linewidth}\centering\includegraphics[width=\linewidth]{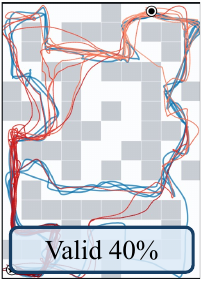}\end{minipage}
\begin{minipage}{0.18\linewidth}\centering\includegraphics[width=\linewidth]{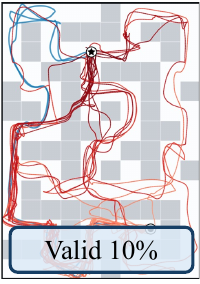}\end{minipage}
\begin{minipage}{0.18\linewidth}\centering\includegraphics[width=\linewidth]{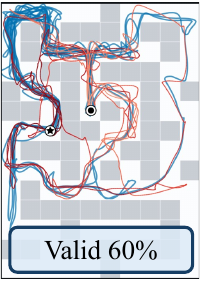}\end{minipage}

\begin{minipage}{0.025\linewidth}\centering\rotatebox{90}{\small\textbf{SCD (Ours)}}\end{minipage}
\begin{minipage}{0.18\linewidth}\centering\includegraphics[width=\linewidth]{fig/C_failures/C_SCD_1.pdf}\end{minipage}
\begin{minipage}{0.18\linewidth}\centering\includegraphics[width=\linewidth]{fig/C_failures/C_SCD_2.pdf}\end{minipage}
\begin{minipage}{0.18\linewidth}\centering\includegraphics[width=\linewidth]{fig/C_failures/C_SCD_3.pdf}\end{minipage}
\begin{minipage}{0.18\linewidth}\centering\includegraphics[width=\linewidth]{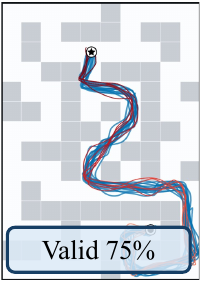}\end{minipage}
\begin{minipage}{0.18\linewidth}\centering\includegraphics[width=\linewidth]{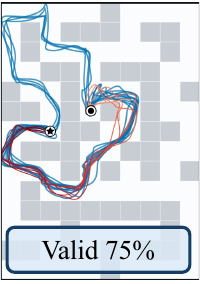}\end{minipage}

\begin{minipage}{0.025\linewidth}\ \end{minipage}
\begin{minipage}{0.18\linewidth}\centering\small{(a) Task 01}\end{minipage}
\begin{minipage}{0.18\linewidth}\centering\small{(b) Task 02}\end{minipage}
\begin{minipage}{0.18\linewidth}\centering\small{(c) Task 03}\end{minipage}
\begin{minipage}{0.18\linewidth}\centering\small{(d) Task 04}\end{minipage}
\begin{minipage}{0.18\linewidth}\centering\small{(e) Task 05}\end{minipage}

\caption{\textbf{Failure cases in HumanoidMaze-Giant.} The red color denotes infeasible trajectories and the blue color denotes feasible ones. \textit{Valid} refers to the percentage of feasible trajectories.}
\label{fig:more-failure-cases}
\end{figure}

\begin{figure}[p]
\centering
\begin{minipage}{0.025\linewidth}\centering\rotatebox{90}{\small{CompDiffuser}}\end{minipage}
\begin{minipage}{0.18\linewidth}\centering\includegraphics[width=\linewidth]{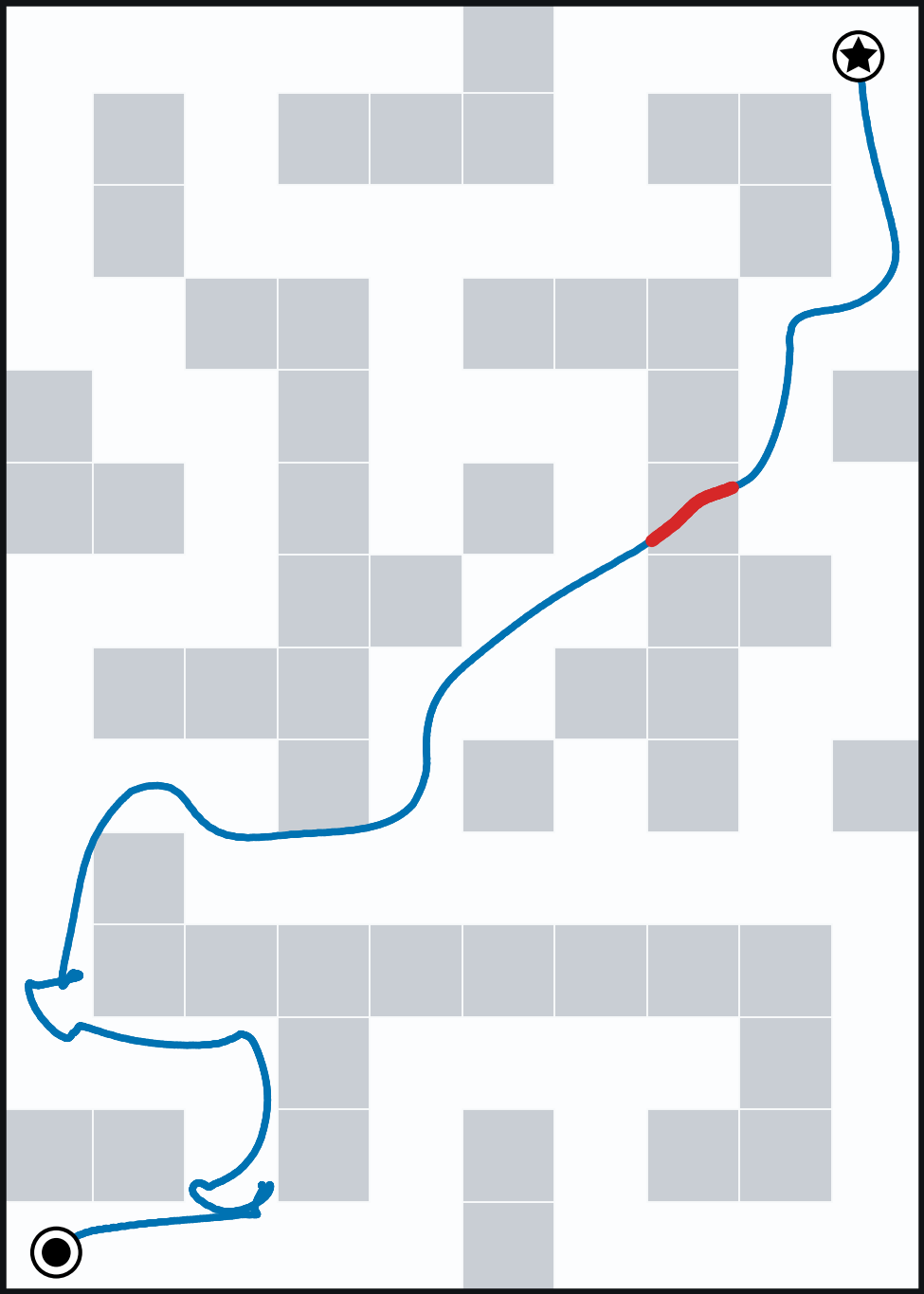}\end{minipage}
\begin{minipage}{0.18\linewidth}\centering\includegraphics[width=\linewidth]{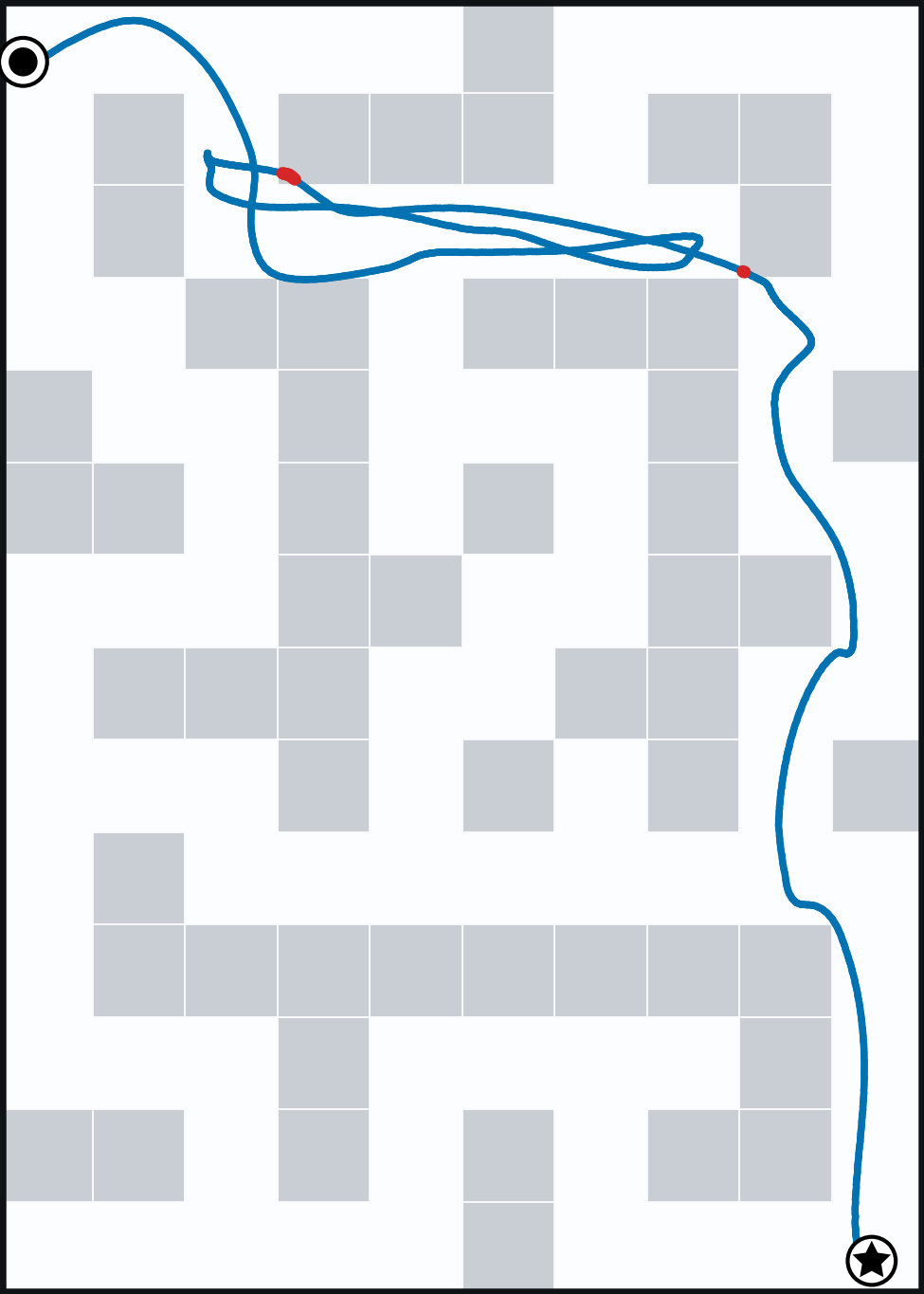}\end{minipage}
\begin{minipage}{0.18\linewidth}\centering\includegraphics[width=\linewidth]{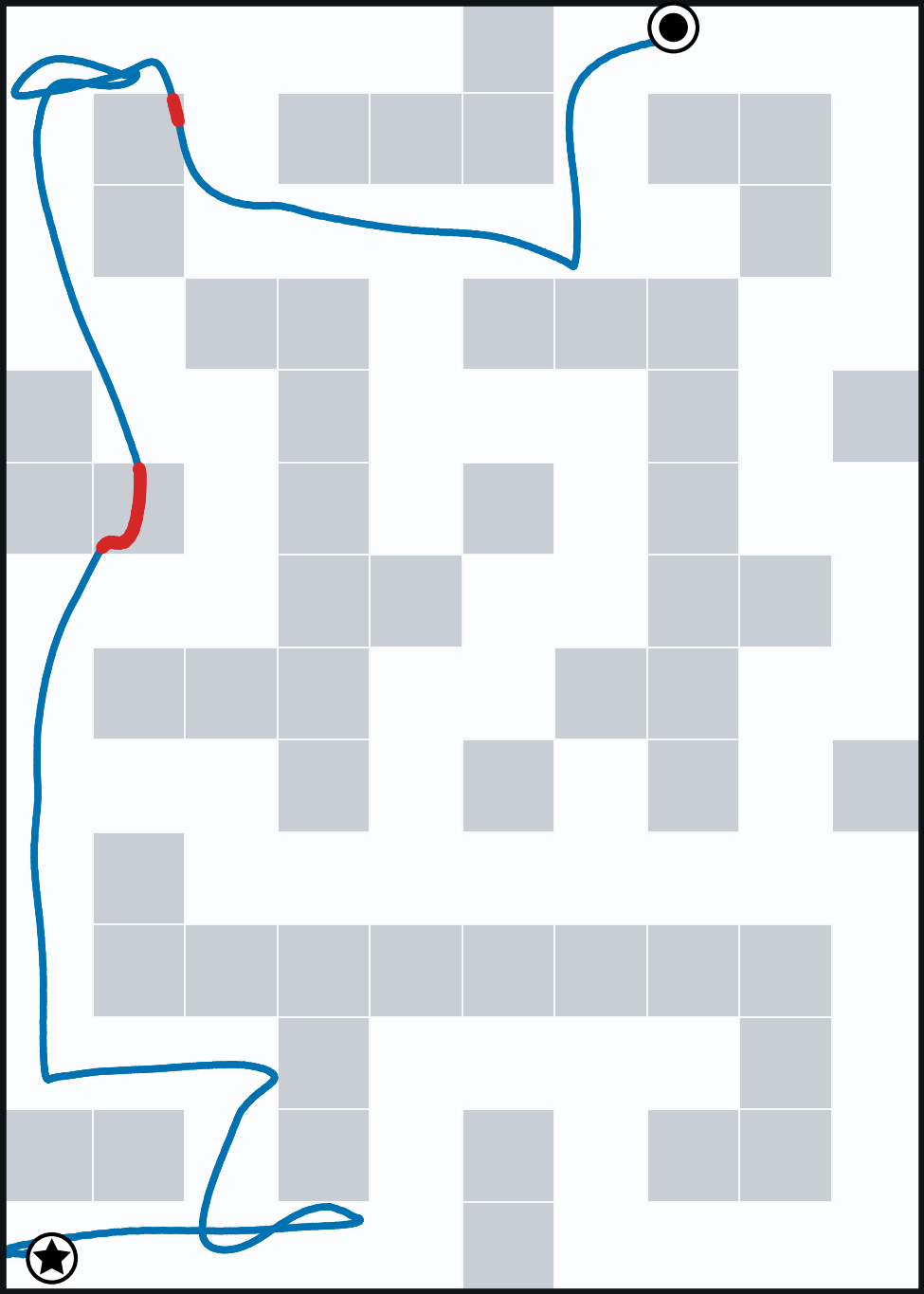}\end{minipage}
\begin{minipage}{0.18\linewidth}\centering\includegraphics[width=\linewidth]{fig/F_qualitative/task04/compdiffuser.png}\end{minipage}
\begin{minipage}{0.18\linewidth}\centering\includegraphics[width=\linewidth]{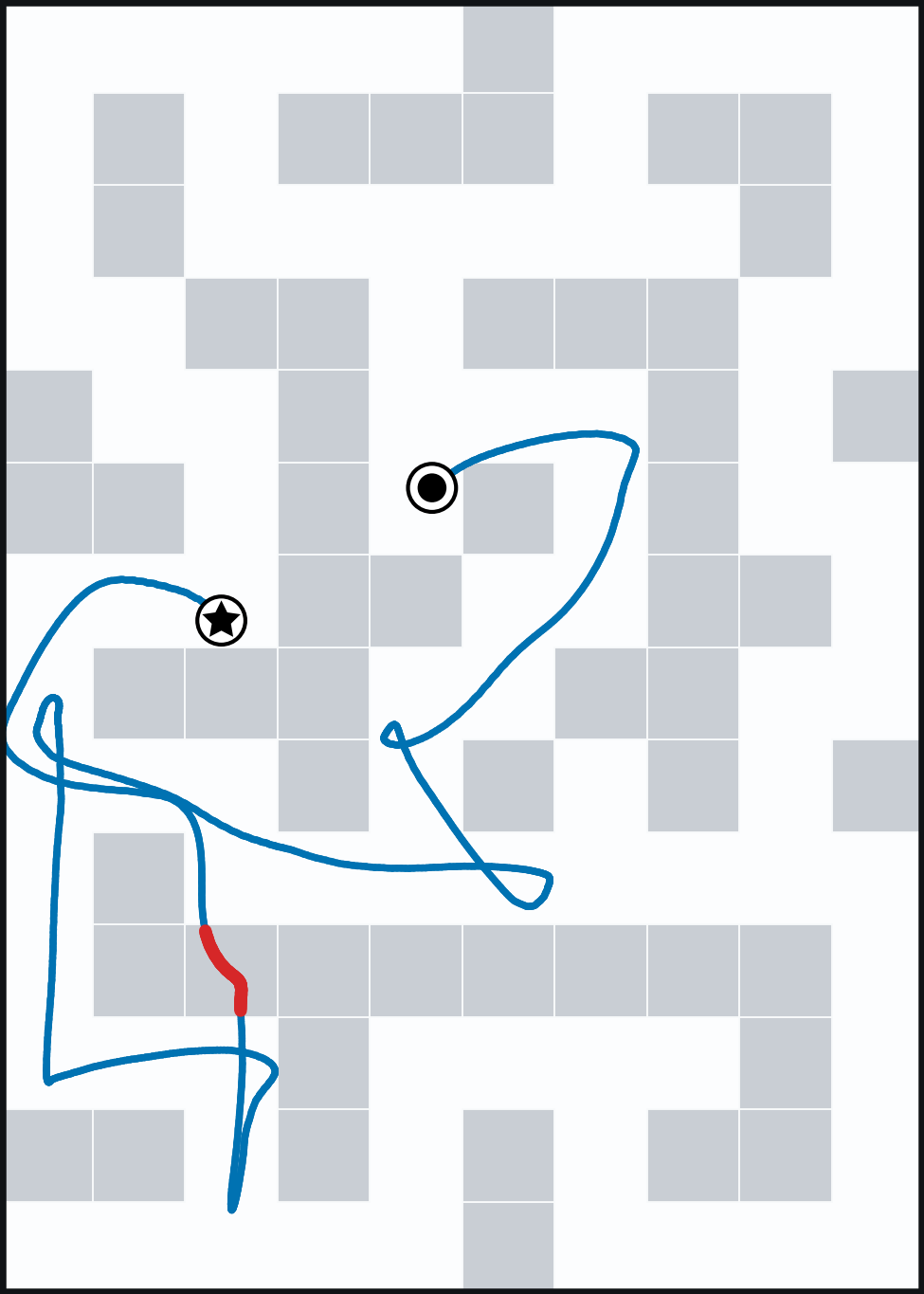}\end{minipage}

\begin{minipage}{0.025\linewidth}\centering\rotatebox{90}{\small{CDGS}}\end{minipage}
\begin{minipage}{0.18\linewidth}\centering\includegraphics[width=\linewidth]{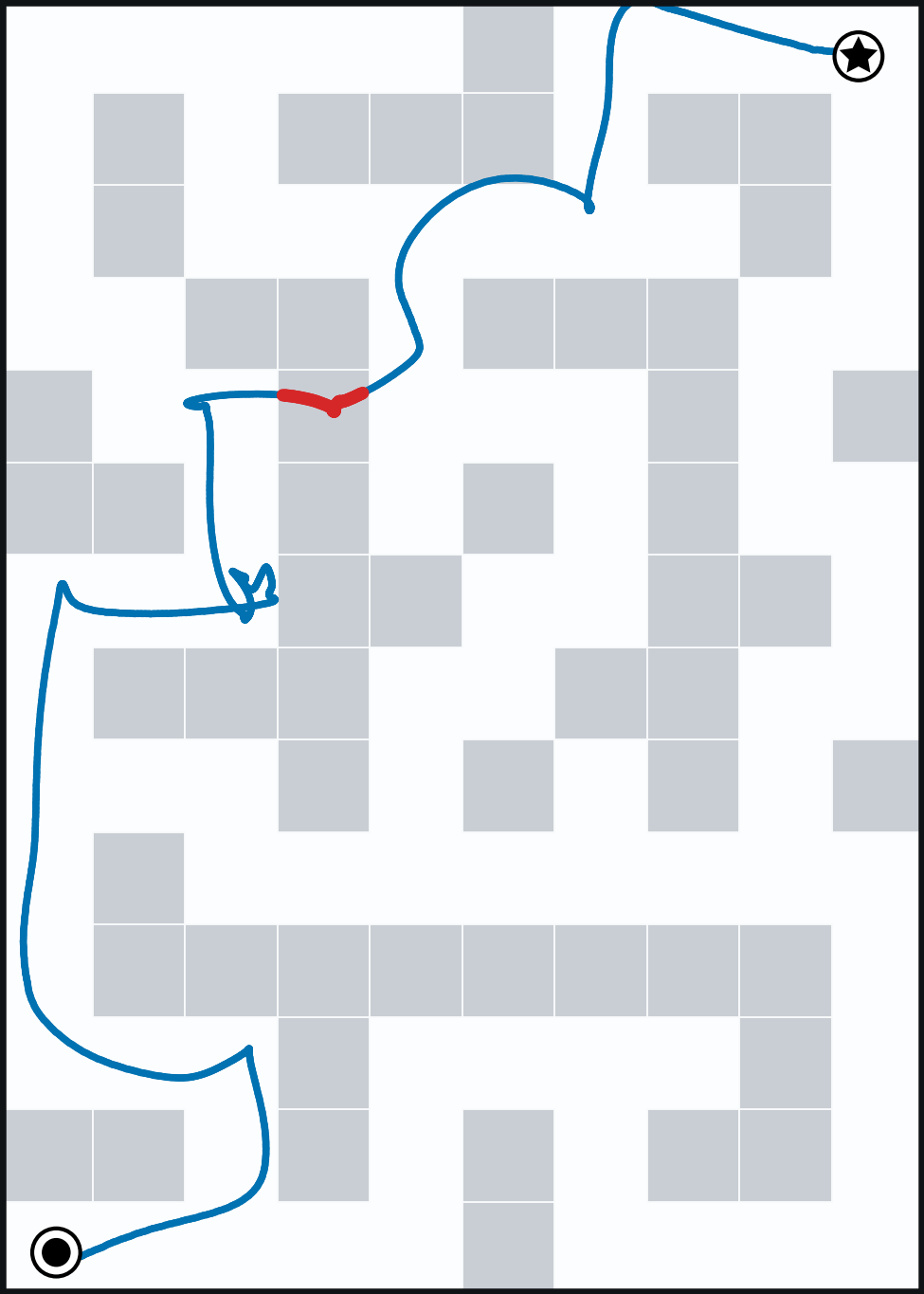}\end{minipage}
\begin{minipage}{0.18\linewidth}\centering\includegraphics[width=\linewidth]{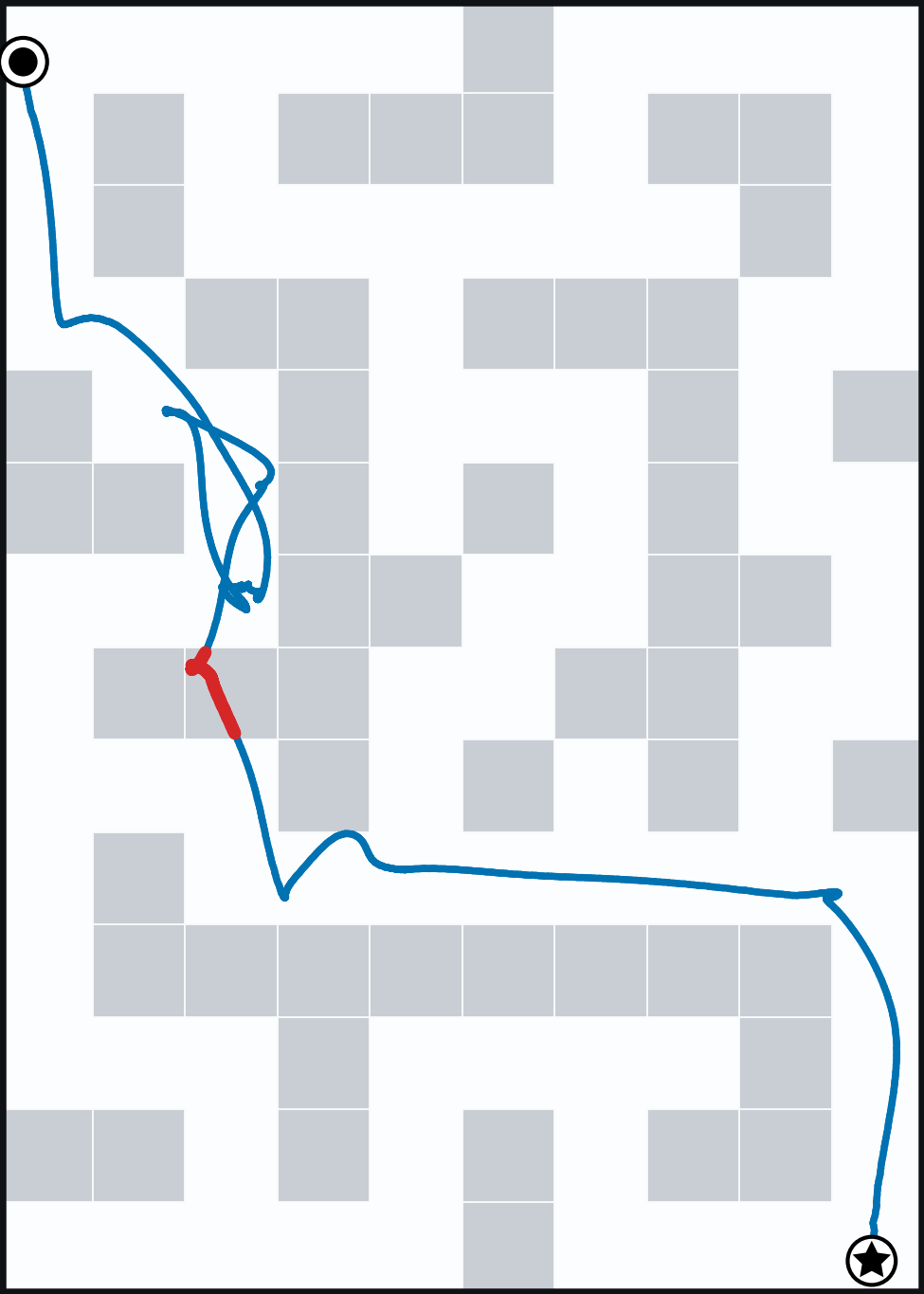}\end{minipage}
\begin{minipage}{0.18\linewidth}\centering\includegraphics[width=\linewidth]{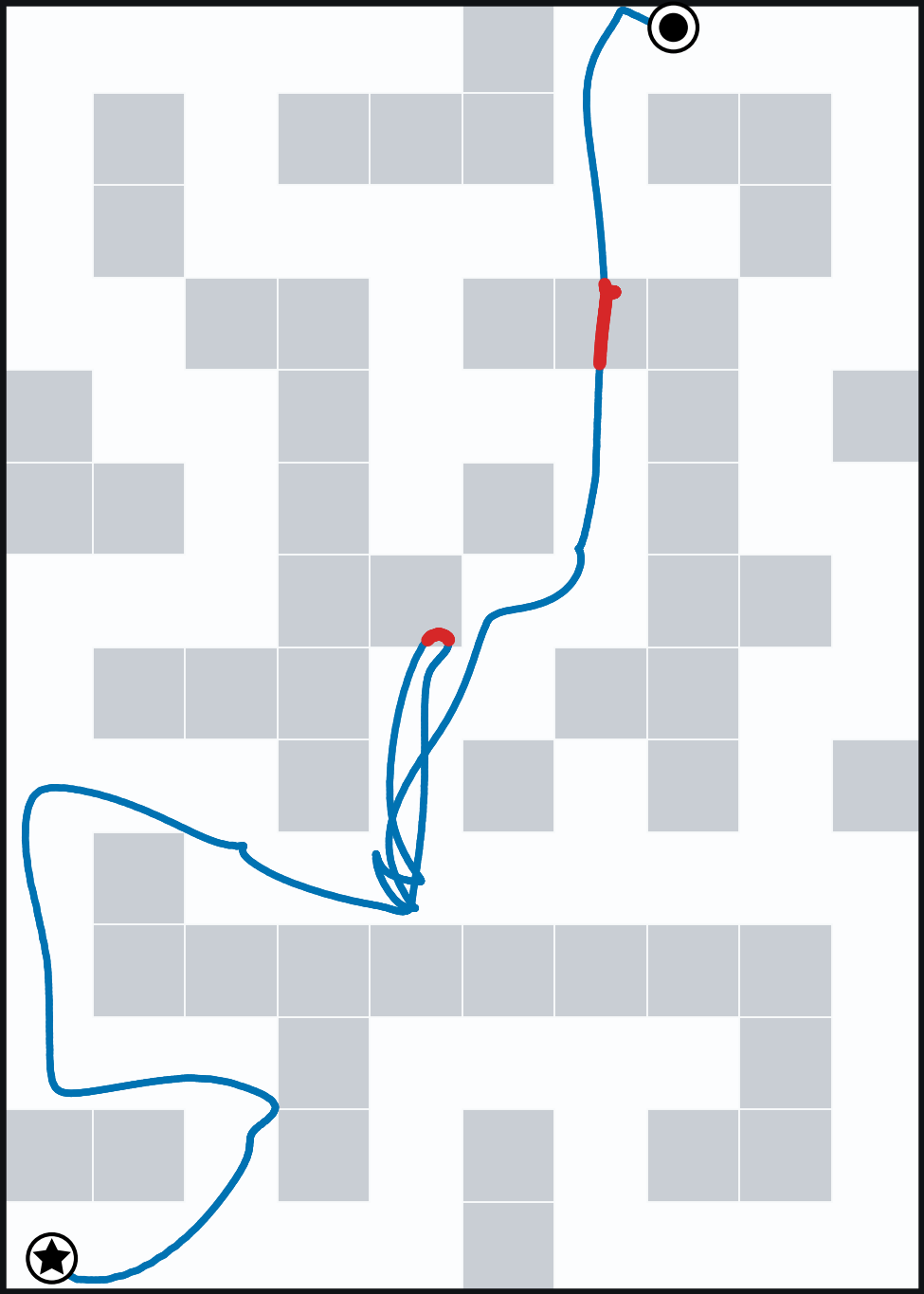}\end{minipage}
\begin{minipage}{0.18\linewidth}\centering\includegraphics[width=\linewidth]{fig/F_qualitative/task04/cdgs.png}\end{minipage}
\begin{minipage}{0.18\linewidth}\centering\includegraphics[width=\linewidth]{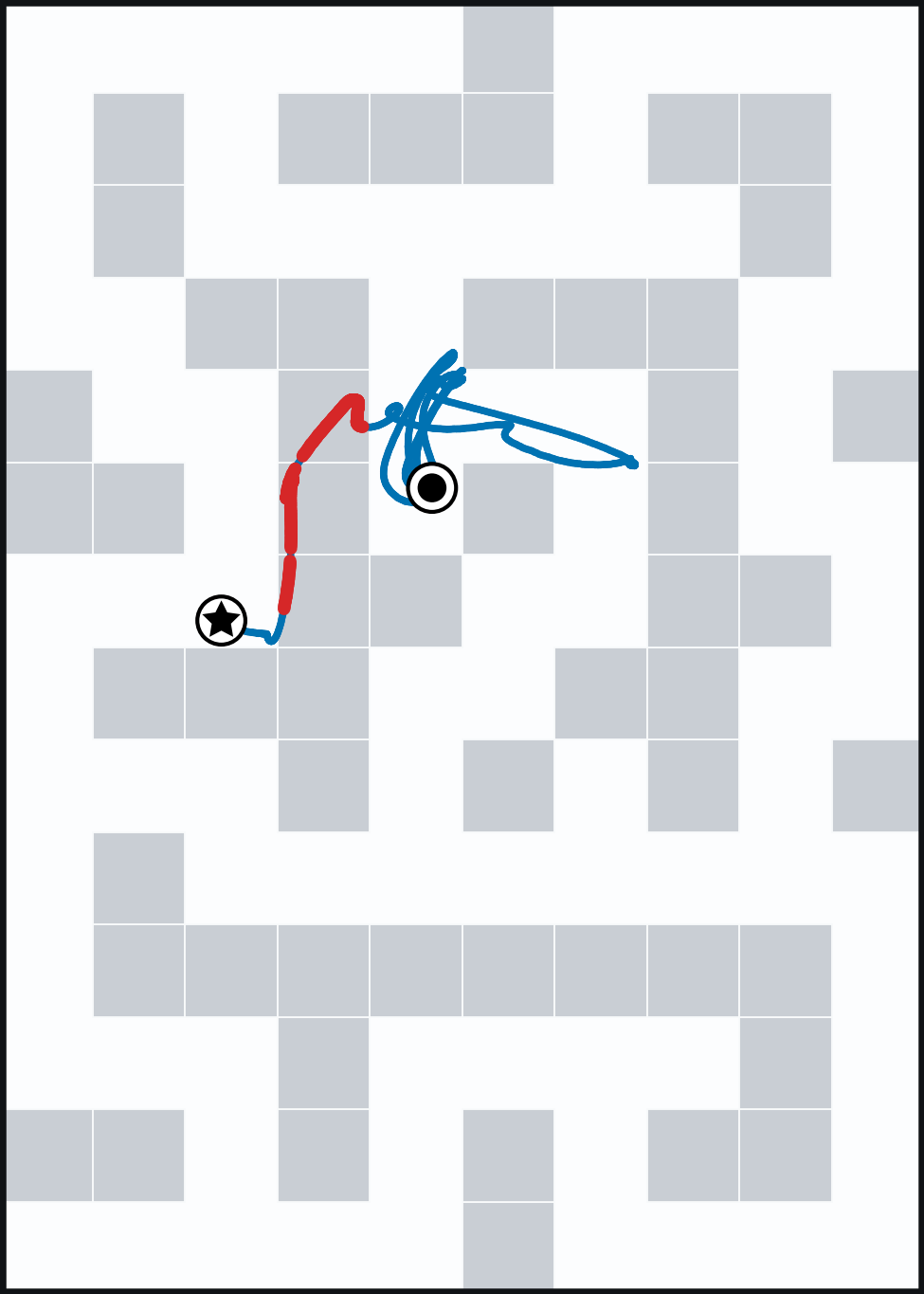}\end{minipage}

\begin{minipage}{0.025\linewidth}\centering\rotatebox{90}{\small{ECD}}\end{minipage}
\begin{minipage}{0.18\linewidth}\centering\includegraphics[width=\linewidth]{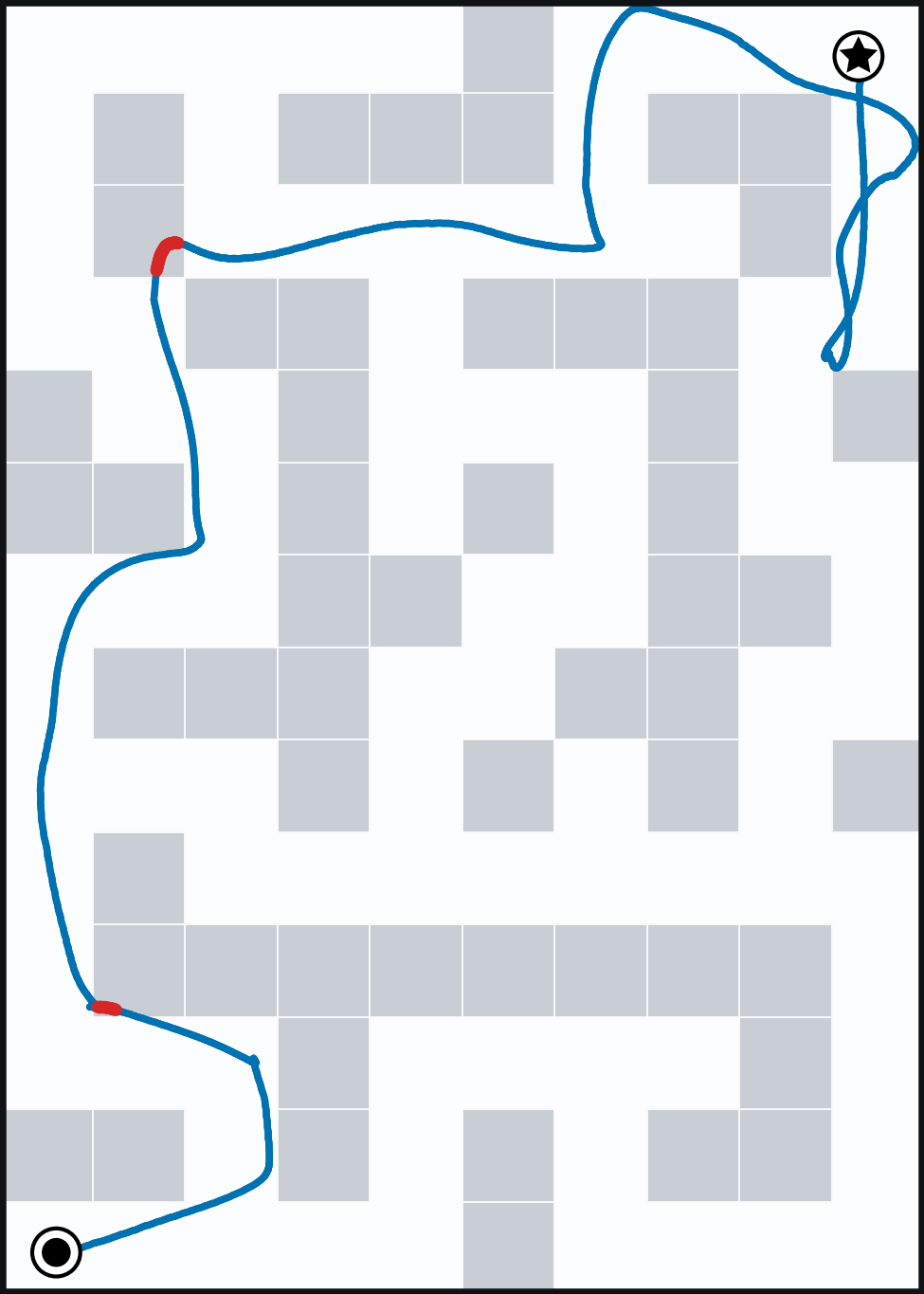}\end{minipage}
\begin{minipage}{0.18\linewidth}\centering\includegraphics[width=\linewidth]{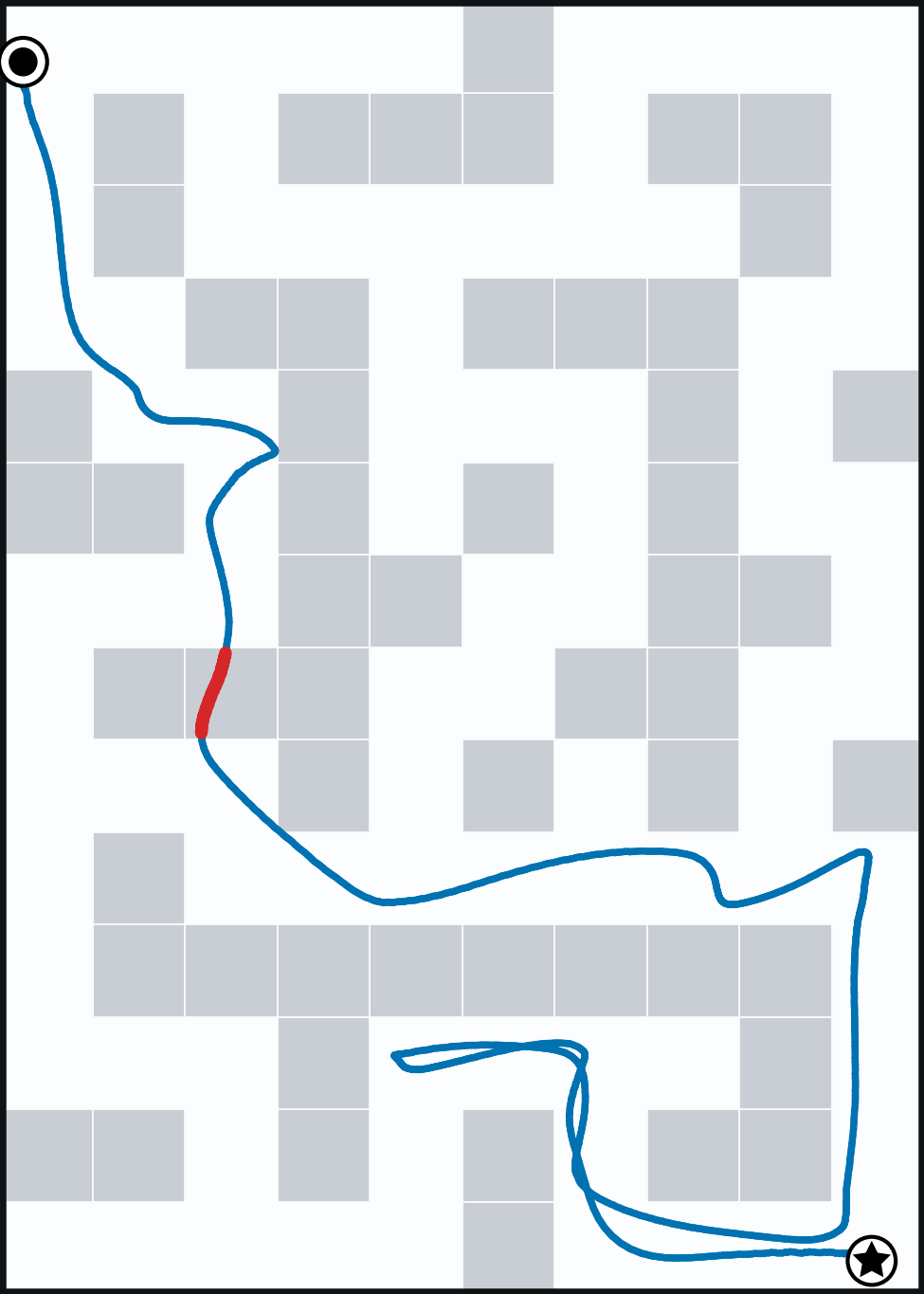}\end{minipage}
\begin{minipage}{0.18\linewidth}\centering\includegraphics[width=\linewidth]{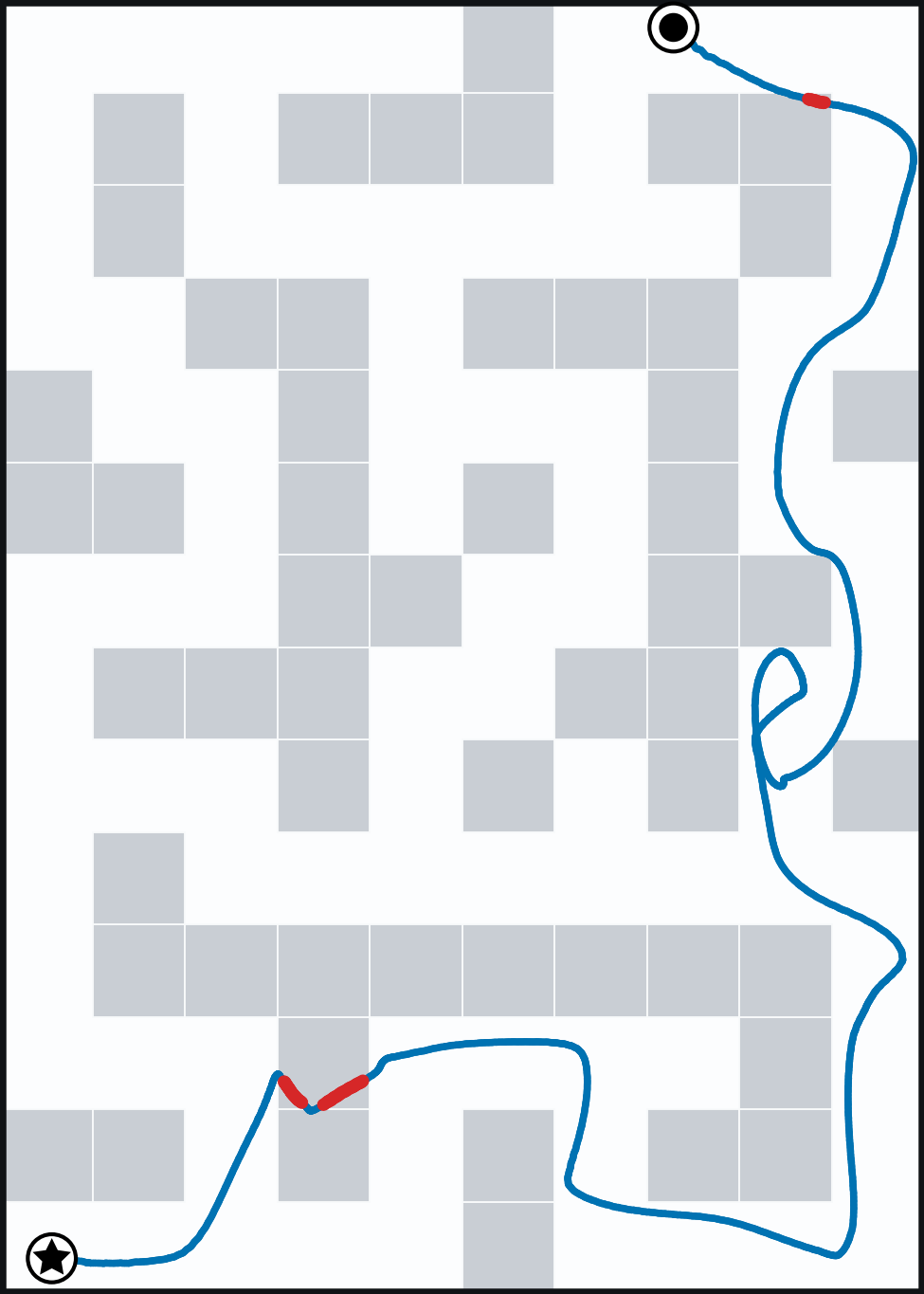}\end{minipage}
\begin{minipage}{0.18\linewidth}\centering\includegraphics[width=\linewidth]{fig/F_qualitative/task04/ecd.png}\end{minipage}
\begin{minipage}{0.18\linewidth}\centering\includegraphics[width=\linewidth]{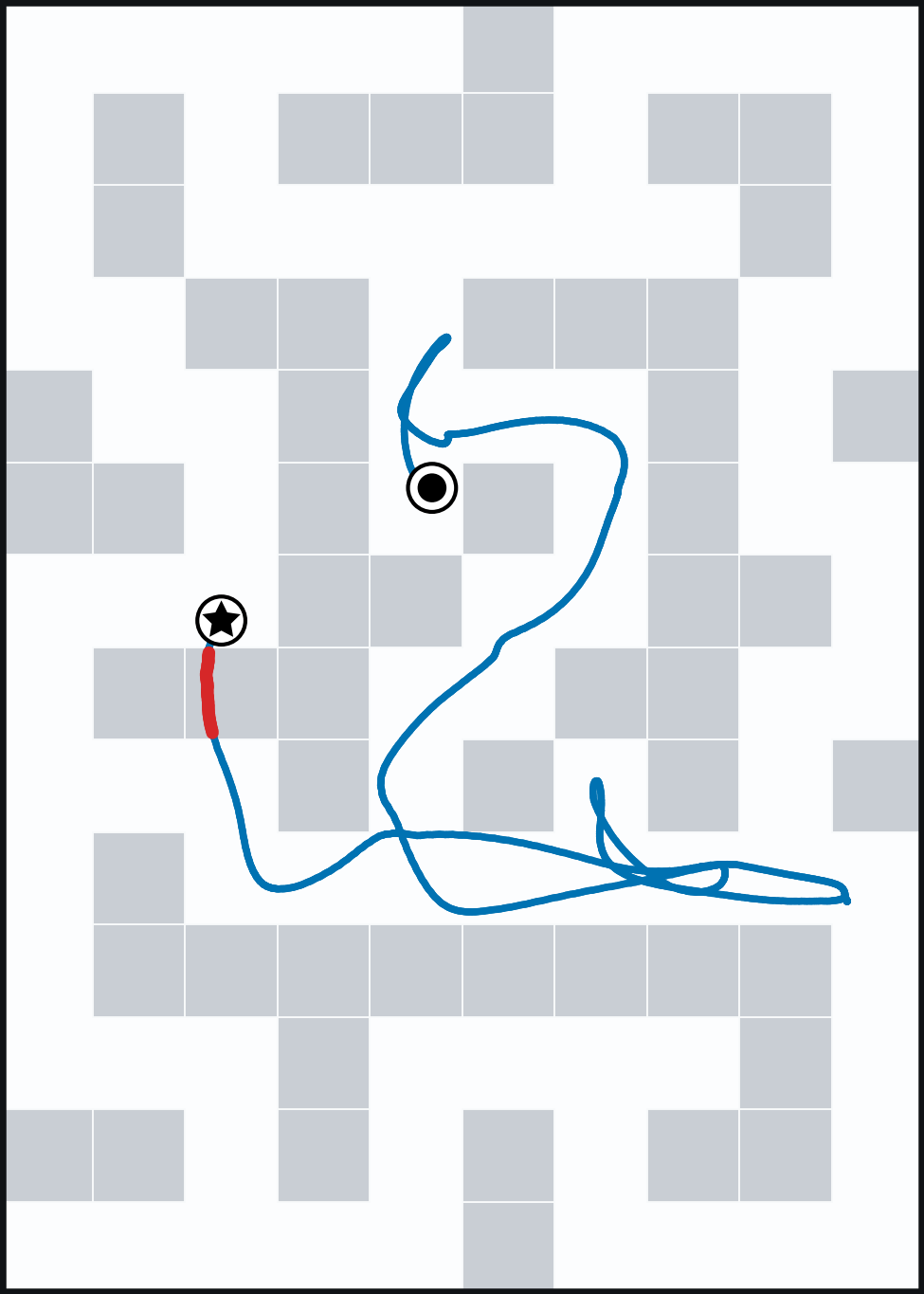}\end{minipage}

\begin{minipage}{0.025\linewidth}\centering\rotatebox{90}{\small{RCD}}\end{minipage}
\begin{minipage}{0.18\linewidth}\centering\includegraphics[width=\linewidth]{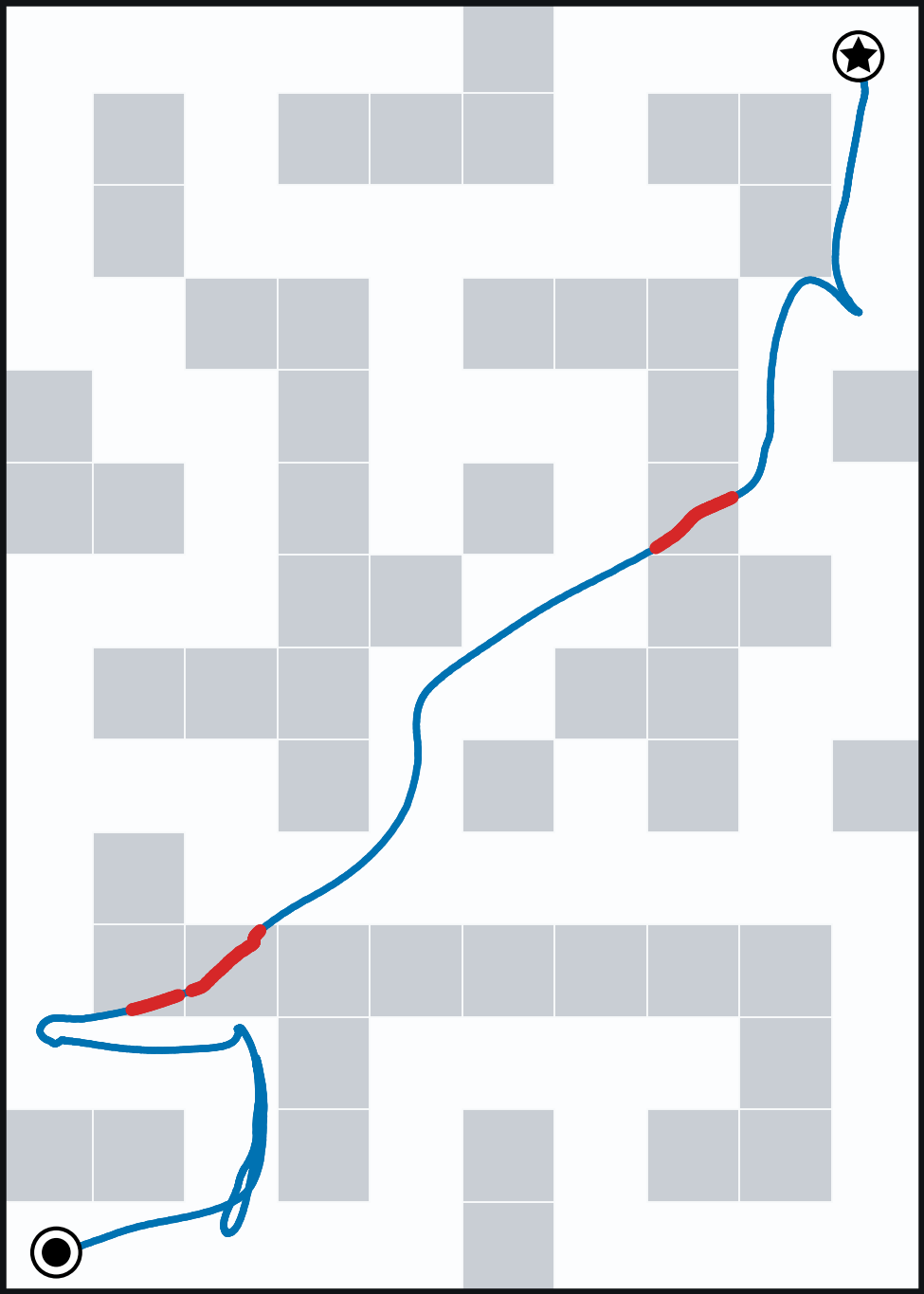}\end{minipage}
\begin{minipage}{0.18\linewidth}\centering\includegraphics[width=\linewidth]{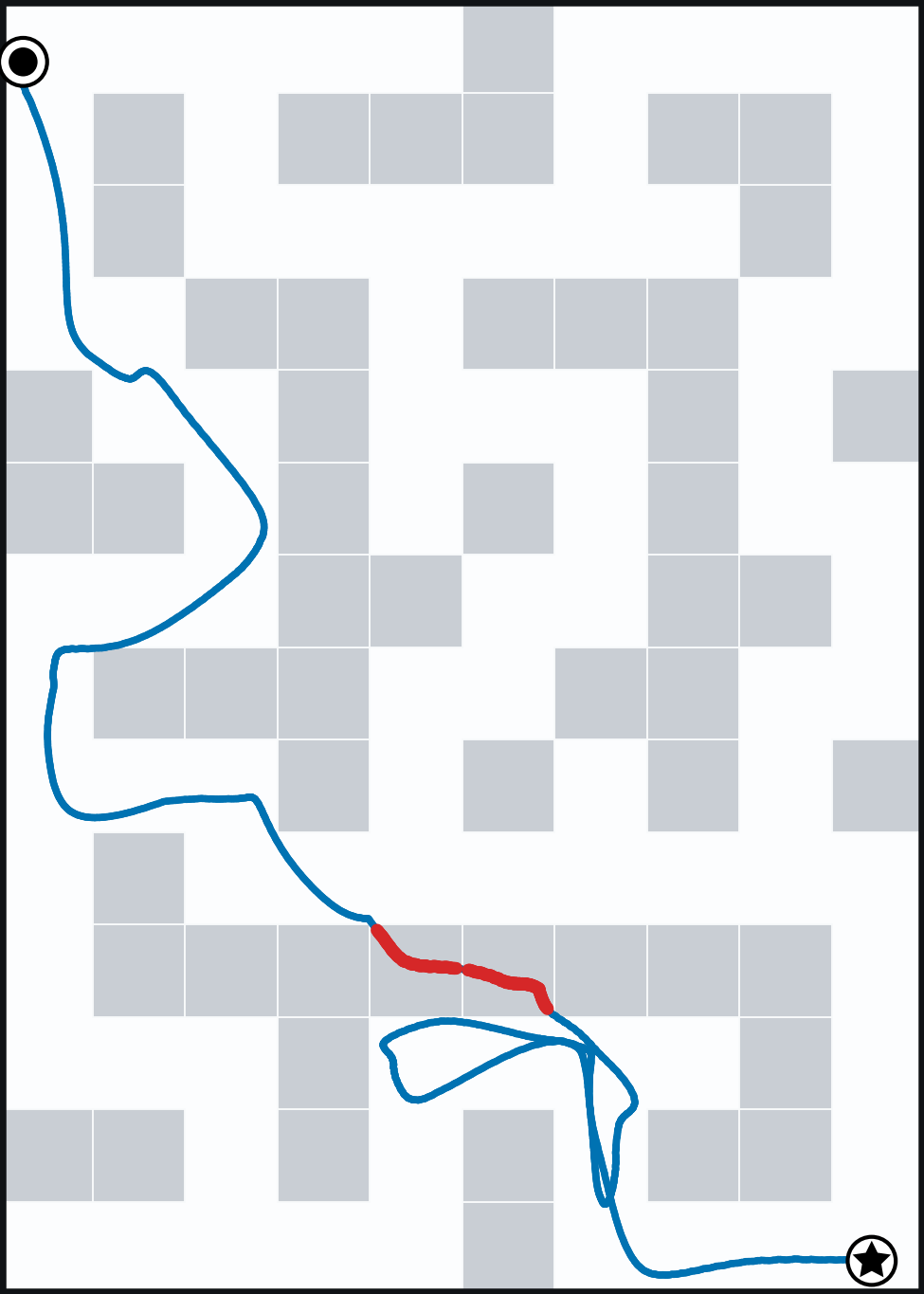}\end{minipage}
\begin{minipage}{0.18\linewidth}\centering\includegraphics[width=\linewidth]{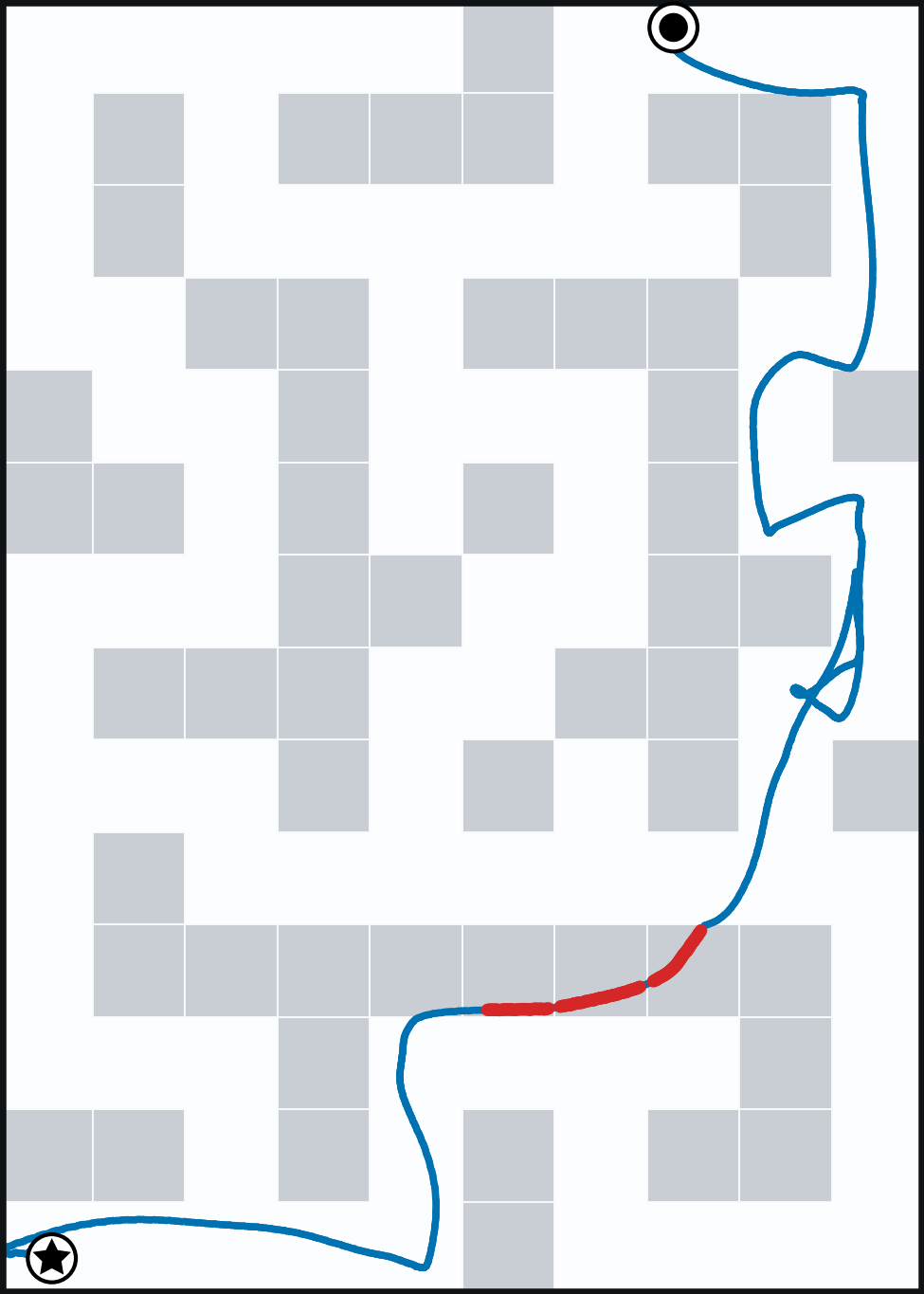}\end{minipage}
\begin{minipage}{0.18\linewidth}\centering\includegraphics[width=\linewidth]{fig/F_qualitative/task04/rcd.png}\end{minipage}
\begin{minipage}{0.18\linewidth}\centering\includegraphics[width=\linewidth]{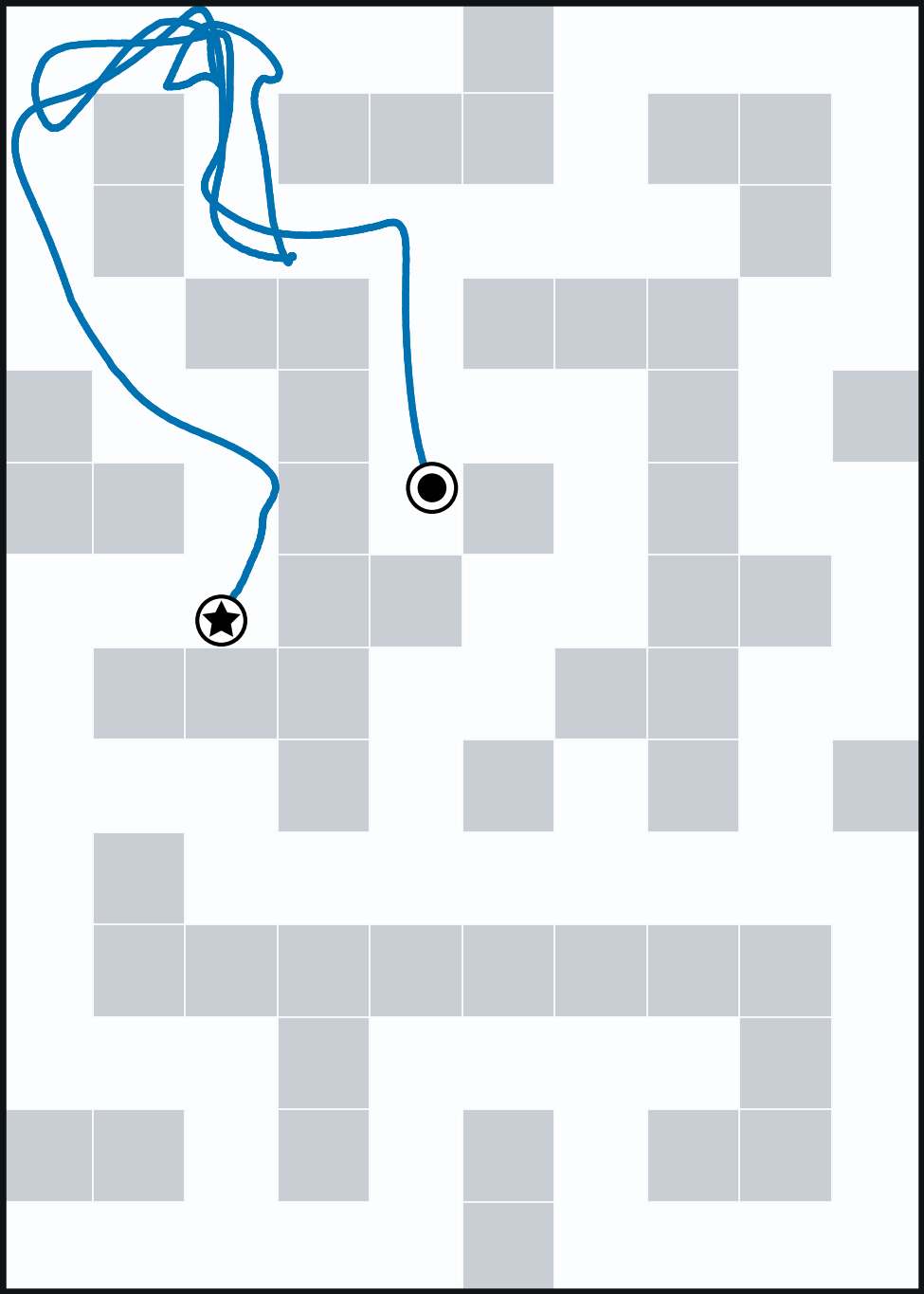}\end{minipage}

\begin{minipage}{0.025\linewidth}\centering\rotatebox{90}{\small\textbf{SCD (Ours)}}\end{minipage}
\begin{minipage}{0.18\linewidth}\centering\includegraphics[width=\linewidth]{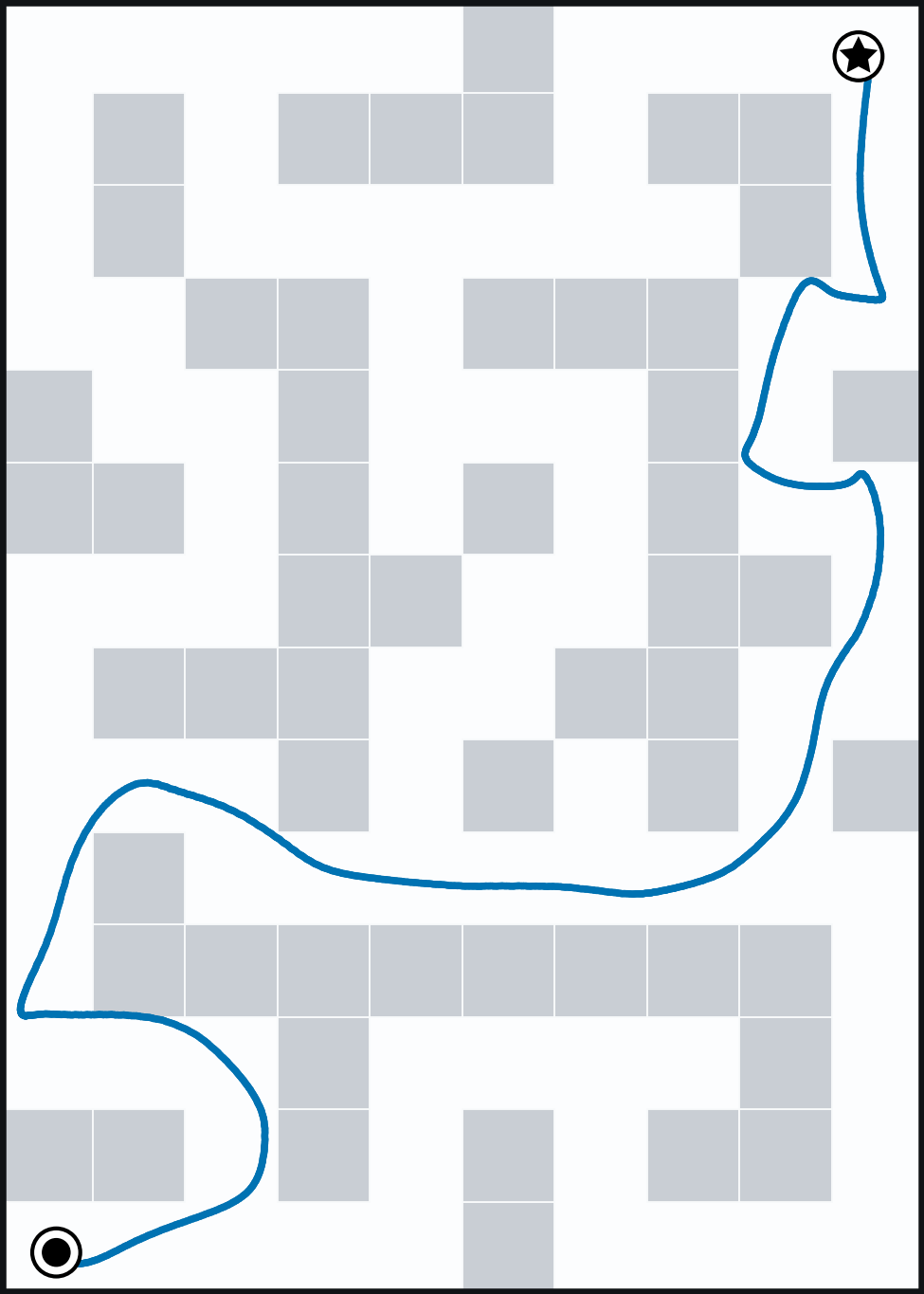}\end{minipage}
\begin{minipage}{0.18\linewidth}\centering\includegraphics[width=\linewidth]{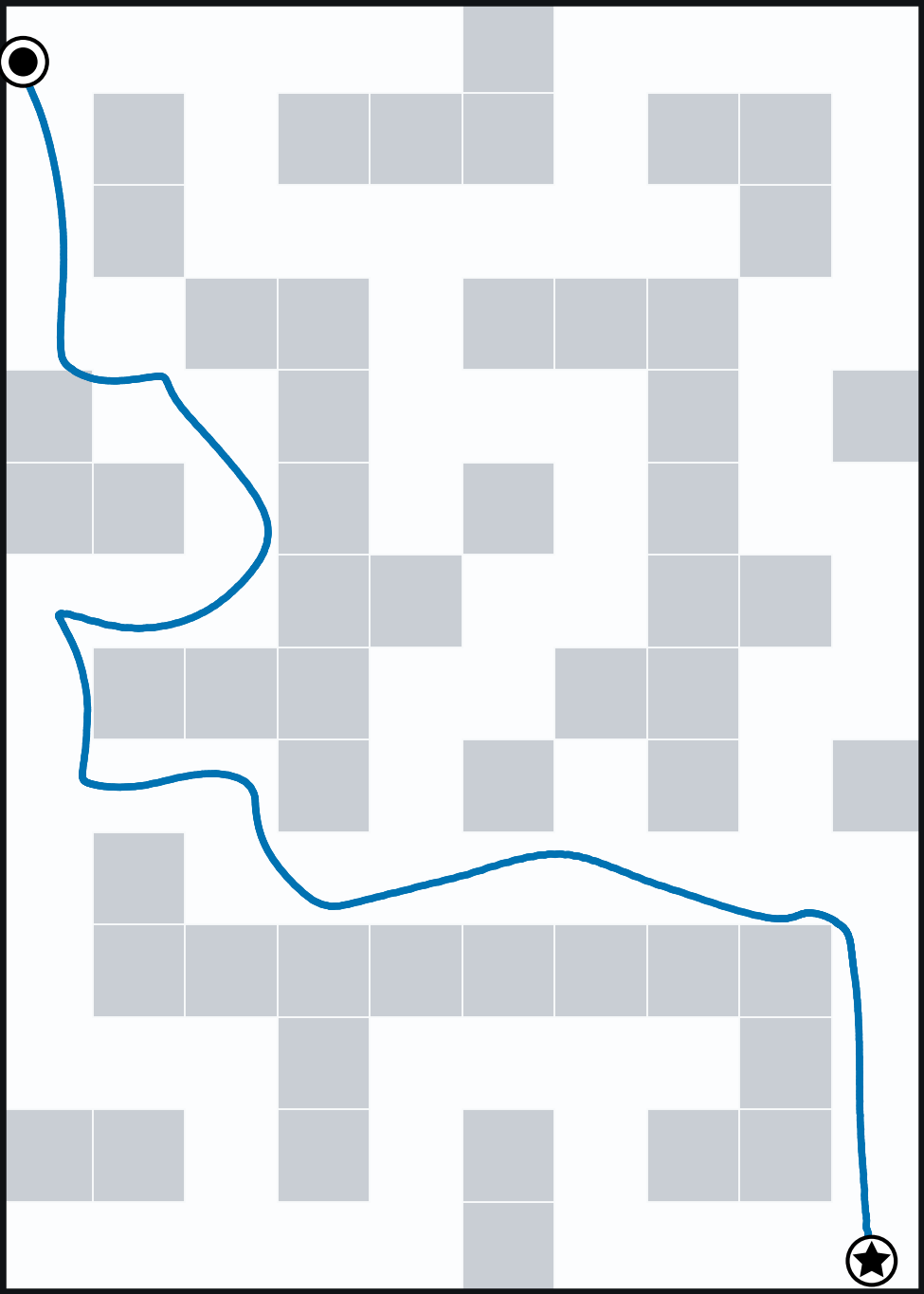}\end{minipage}
\begin{minipage}{0.18\linewidth}\centering\includegraphics[width=\linewidth]{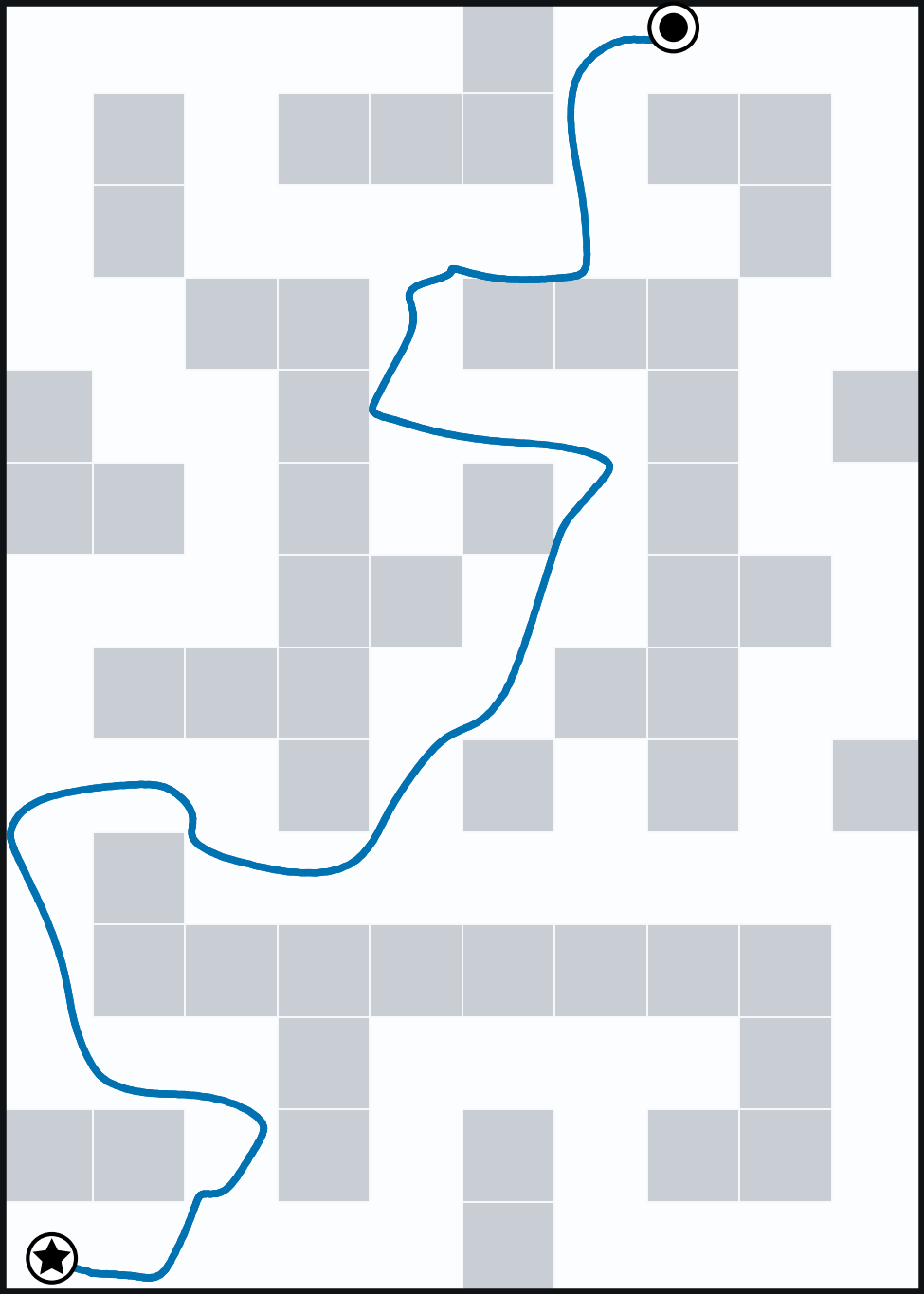}\end{minipage}
\begin{minipage}{0.18\linewidth}\centering\includegraphics[width=\linewidth]{fig/F_qualitative/task04/scd.png}\end{minipage}
\begin{minipage}{0.18\linewidth}\centering\includegraphics[width=\linewidth]{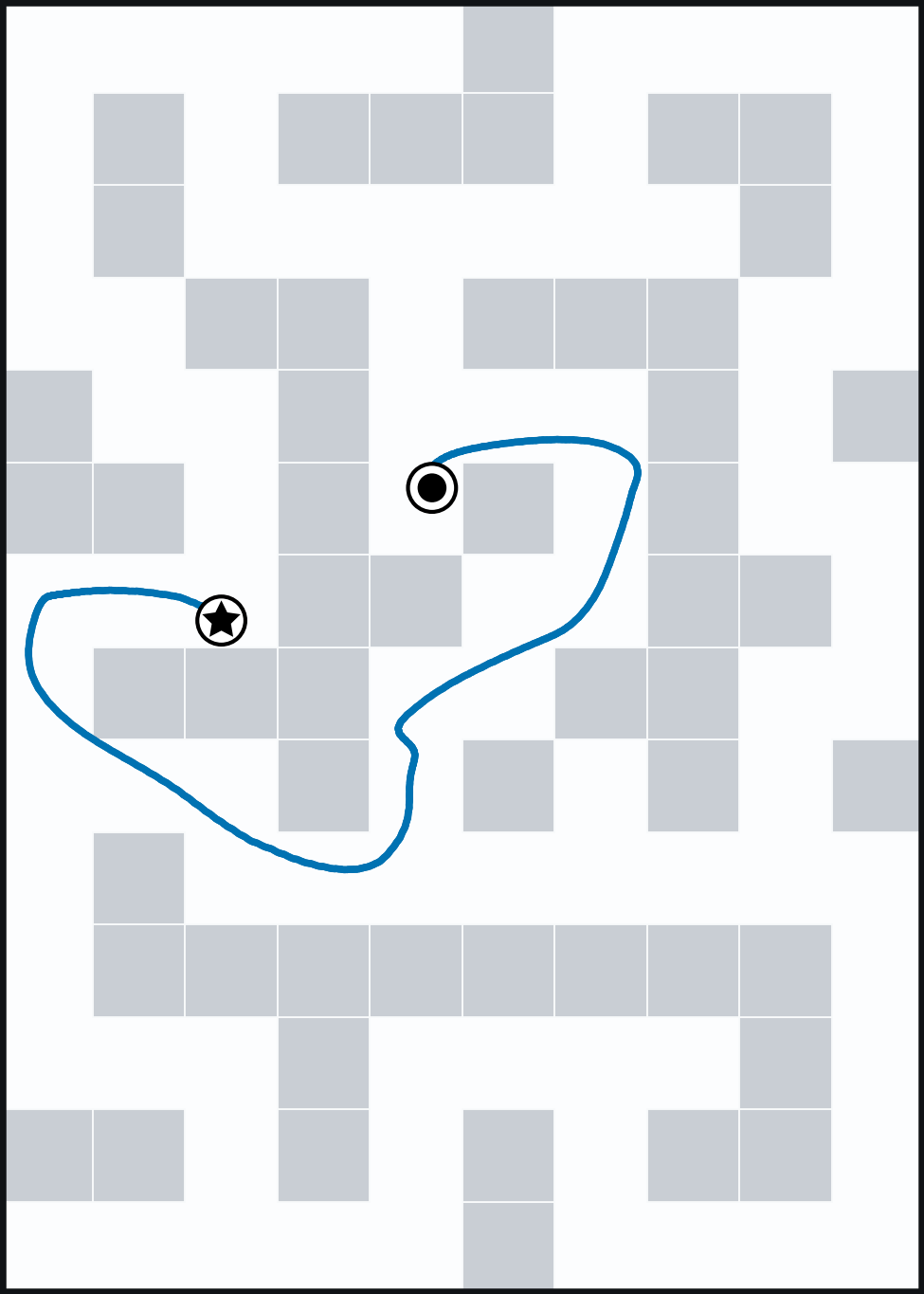}\end{minipage}

\begin{minipage}{0.025\linewidth}\ \end{minipage}
\begin{minipage}{0.18\linewidth}\centering\small{(a) Task 01}\end{minipage}
\begin{minipage}{0.18\linewidth}\centering\small{(b) Task 02}\end{minipage}
\begin{minipage}{0.18\linewidth}\centering\small{(c) Task 03}\end{minipage}
\begin{minipage}{0.18\linewidth}\centering\small{(d) Task 04}\end{minipage}
\begin{minipage}{0.18\linewidth}\centering\small{(e) Task 05}\end{minipage}
\caption{\textbf{Qualitative results of long-horizon planning in HumanoidMaze-Giant.} The red color denotes infeasible parts and the blue color denotes feasible parts.}
\label{fig:more-qualitative-result}
\end{figure}

\clearpage
\section{Licenses}
\label[appendix]{app:licenses}

This work uses the following existing codebases, models, and datasets:
\begin{itemize}
\item OGBench~\citep[][\url{https://github.com/seohongpark/ogbench}]{park2025ogbench}: Used for the datasets and evaluation environments. Licensed under the MIT License.
\item MuJoCo (\url{https://github.com/google-deepmind/mujoco}): Used as the physics simulator of OGBench. Licensed under the Apache-2.0 License.
\item CompDiffuser~\citep[][\url{https://github.com/devinluo27/comp_diffuser_release}]{luo2025generative}: Used for the pretrained diffusion models and controllers, and as the base implementation of the planner. Licensed under the MIT License.
\item CDGS~\citep[][\url{https://github.com/UtkarshMishra04/CDGS_ogbench}]{ICLR2026_6fbe71b5}: Used as a reference implementation for the CDGS baseline. Licensed under the MIT License.
\item ECD~\citep[][\url{https://github.com/GradientSpaces/ECD}]{sun2026energybased}: Used as a reference implementation for the ECD baseline. Licensed under the Apache-2.0 License.
\item RCD~\citep[][\url{https://github.com/leekwoon/rcd}]{lee2026refiningcompositionaldiffusionreliable}: Used as a reference implementation for the RCD baseline. Licensed under the MIT License.
\item FAISS (\url{https://github.com/facebookresearch/faiss}): Used for the $k$-NN search in density reweighting. Licensed under the MIT License.
\end{itemize}

\end{document}